\pdfoutput=1
\documentclass[11pt]{article}

\usepackage{acl}

\usepackage{times}
\usepackage{latexsym}

\usepackage[T1]{fontenc}
\usepackage[utf8]{inputenc}

\usepackage{microtype}

\usepackage{amsmath,amsfonts,bm}

\def\eqref#1{equation~\ref{#1}}
\def\1{\bm{1}}

\def\va{{\bm{a}}}

\def\ve{{\bm{e}}}

\def\vh{{\bm{h}}}

\def\vr{{\bm{r}}}

\def\vv{{\bm{v}}}

\def\vx{{\bm{x}}}
\def\vy{{\bm{y}}}

\def\mB{{\bm{B}}}

\def\mE{{\bm{E}}}

\def\mH{{\bm{H}}}

\def\mW{{\bm{W}}}

\DeclareMathAlphabet{\mathsfit}{\encodingdefault}{\sfdefault}{m}{sl}
\SetMathAlphabet{\mathsfit}{bold}{\encodingdefault}{\sfdefault}{bx}{n}

\newcommand{\softmax}{\mathrm{softmax}}

\usepackage{hyperref}       % hyperlinks
\usepackage{url}            % simple URL typesetting
\usepackage{booktabs}       % professional-quality tables
\usepackage{amsfonts}       % blackboard math symbols
\usepackage{nicefrac}       % compact symbols for 1/2, etc.
\usepackage{microtype}      % microtypography
\usepackage{xcolor}         % colors
\usepackage{graphicx}
\usepackage{enumitem}
\usepackage{arydshln}
\usepackage{mathtools}

\usepackage{makecell}
\usepackage{multirow}
\usepackage{subcaption}
\usepackage{wrapfig}
\usepackage{stfloats}       % Place the figure at the bottom

\definecolor{red}{RGB}{205,33,42}

\definecolor{blue}{RGB}{33,105,225}

\definecolor{realred}{RGB}{255,0,0}
\newcommand{\RED}[1]{\textbf{\textcolor{red}{#1}}}
\newcommand{\BLUE}[1]{\textbf{\textcolor{blue}{#1}}}

\title{KALA: Knowledge-Augmented Language Model Adaptation} 

\author{Minki Kang$^1$$^,$$^2$\thanks{* Equal contribution} \: Jinheon Baek$^{1}$$^*$ \: Sung Ju Hwang$^1$$^,$$^2$ \\  
KAIST$^1$,  AITRICS$^2$\\
\texttt{\{zzxc1133, jinheon.baek, sjhwang82\}@kaist.ac.kr}}

\begin{document}
\maketitle

\begin{abstract}
Pre-trained language models (PLMs) have achieved remarkable success on various natural language understanding tasks.
Simple fine-tuning of PLMs, on the other hand, might be suboptimal for domain-specific tasks because they cannot possibly cover knowledge from all domains. 
While adaptive pre-training of PLMs can help them obtain domain-specific knowledge, it requires a large training cost.
Moreover, adaptive pre-training can harm the PLM's performance on the downstream task by causing catastrophic forgetting of its general knowledge.
To overcome such limitations of adaptive pre-training for PLM adaption, we propose a novel domain adaption framework for PLMs coined as \textbf{K}nowledge-\textbf{A}ugmented \textbf{L}anguage model \textbf{A}daptation (\textbf{KALA}), which modulates the intermediate hidden representations of PLMs with domain knowledge, consisting of entities and their relational facts. 
We validate the performance of our KALA on question answering and named entity recognition tasks on multiple datasets across various domains. 
The results show that, despite being computationally efficient, our KALA largely outperforms adaptive pre-training.
Code is available at: \href{https://github.com/Nardien/KALA/}{https://github.com/Nardien/KALA}.
\end{abstract}
\section{Introduction}
\label{sec:intro}
Pre-trained Language Models (PLMs)~\citep{BERT, GPT3} have shown to be effective on various Natural Language Understanding (NLU) tasks. Although PLMs aim to address diverse downstream tasks from various data sources, there have been considerable efforts to adapt the PLMs to specific \textit{domains} —distributions over the language characterizing a given topic or genre~\citep{dontstoppt}— for which the acquisition of domain knowledge is required to accurately solve the downstream tasks 
(e.g., Biomedical Named Entity Recognition~\citep{NCBI}).

\begin{figure}[t]
    \centering
    \includegraphics[width=1.0\linewidth]{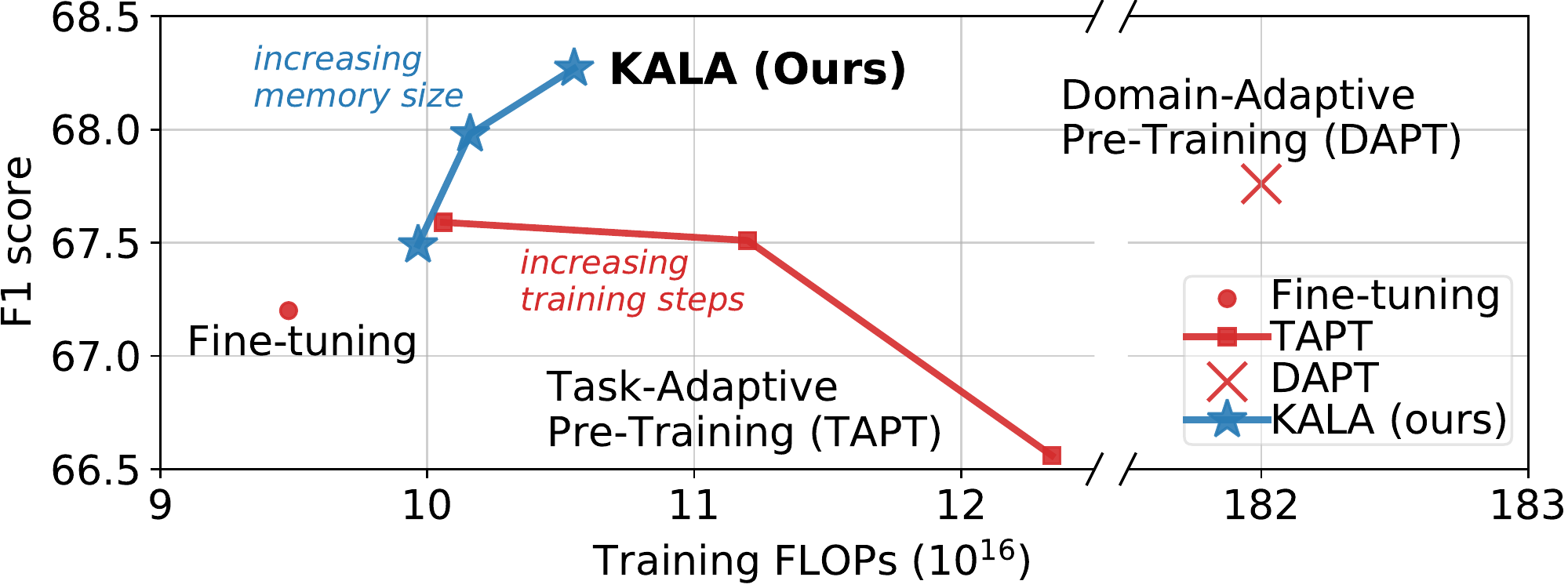}
    \vskip -0.1in
    \caption{\small F1 Score and Training FLOPs for different methods on Question Answering (NewsQA). Note that DAPT uses about 112 times larger data for adaptation. Details are in $\S$\ref{sec:efficiency}}
    \vskip -0.15in 
    \label{fig:main}
\end{figure}

\begin{figure*}[ht]
    \centering
    \begin{minipage}{0.80\linewidth}
        \centerline{\includegraphics[width=0.975\columnwidth]{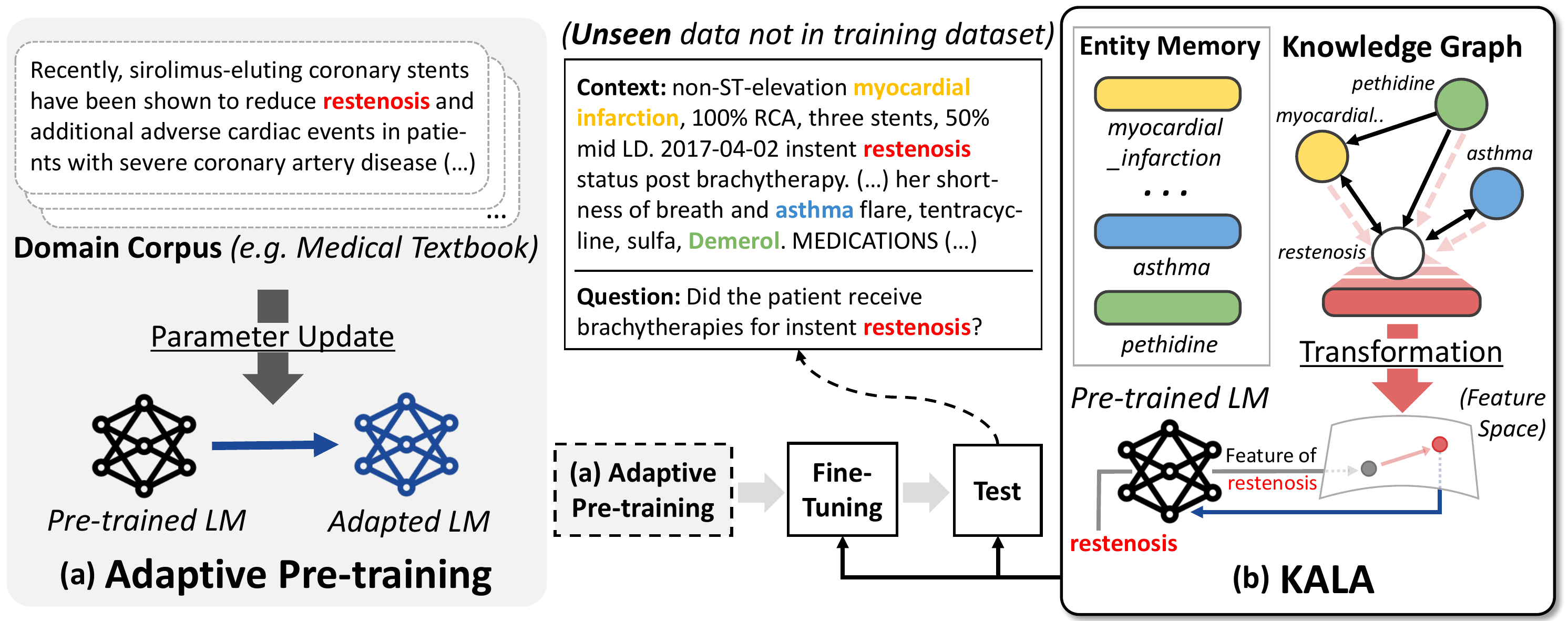}}
    \end{minipage}
    \begin{minipage}{0.19\linewidth}
        \begin{subfigure}[b]{1.0\textwidth}
           \includegraphics[width=1\linewidth]{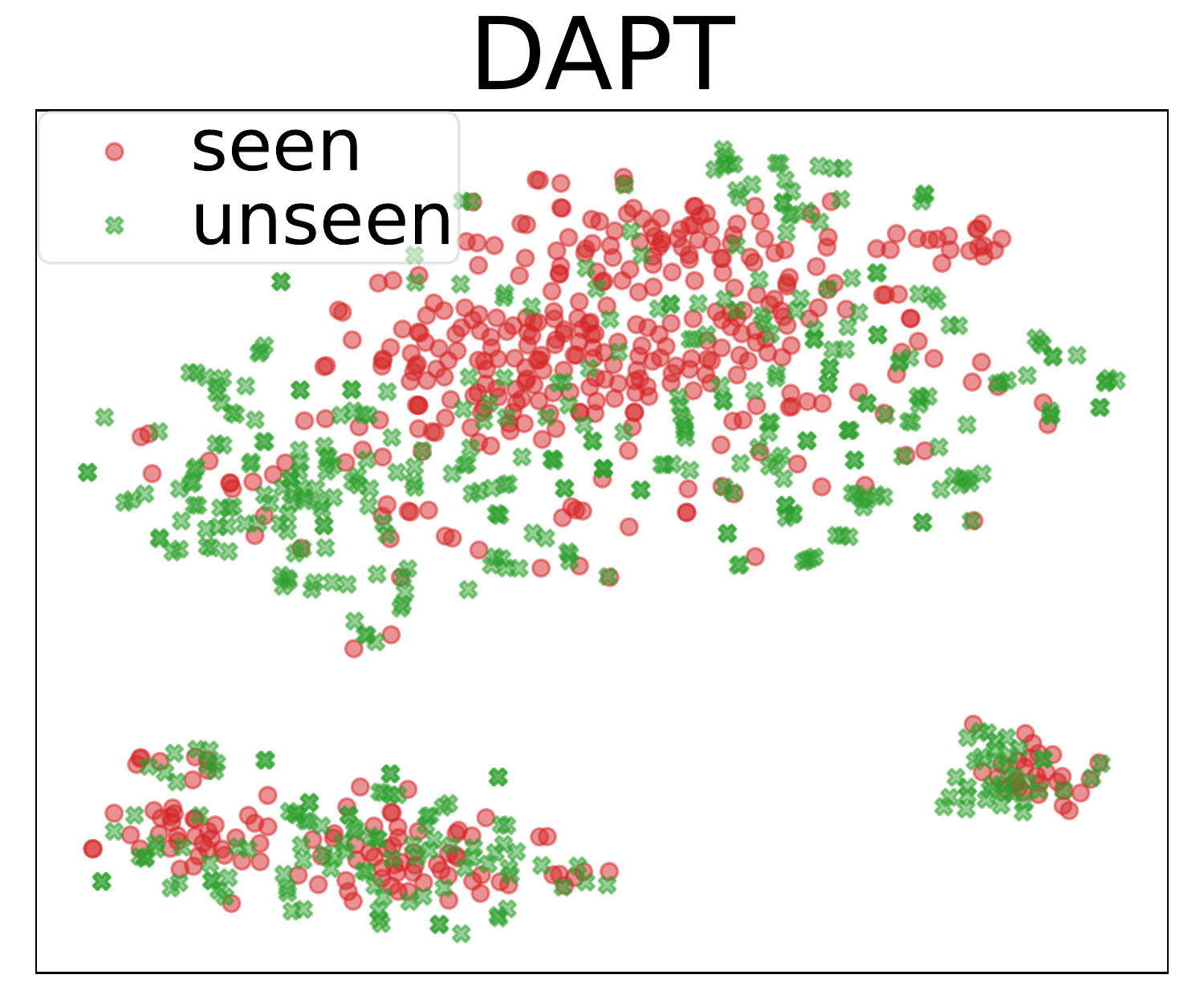}
        \end{subfigure}
        \begin{subfigure}[b]{1.0\textwidth}
           \includegraphics[width=1\linewidth]{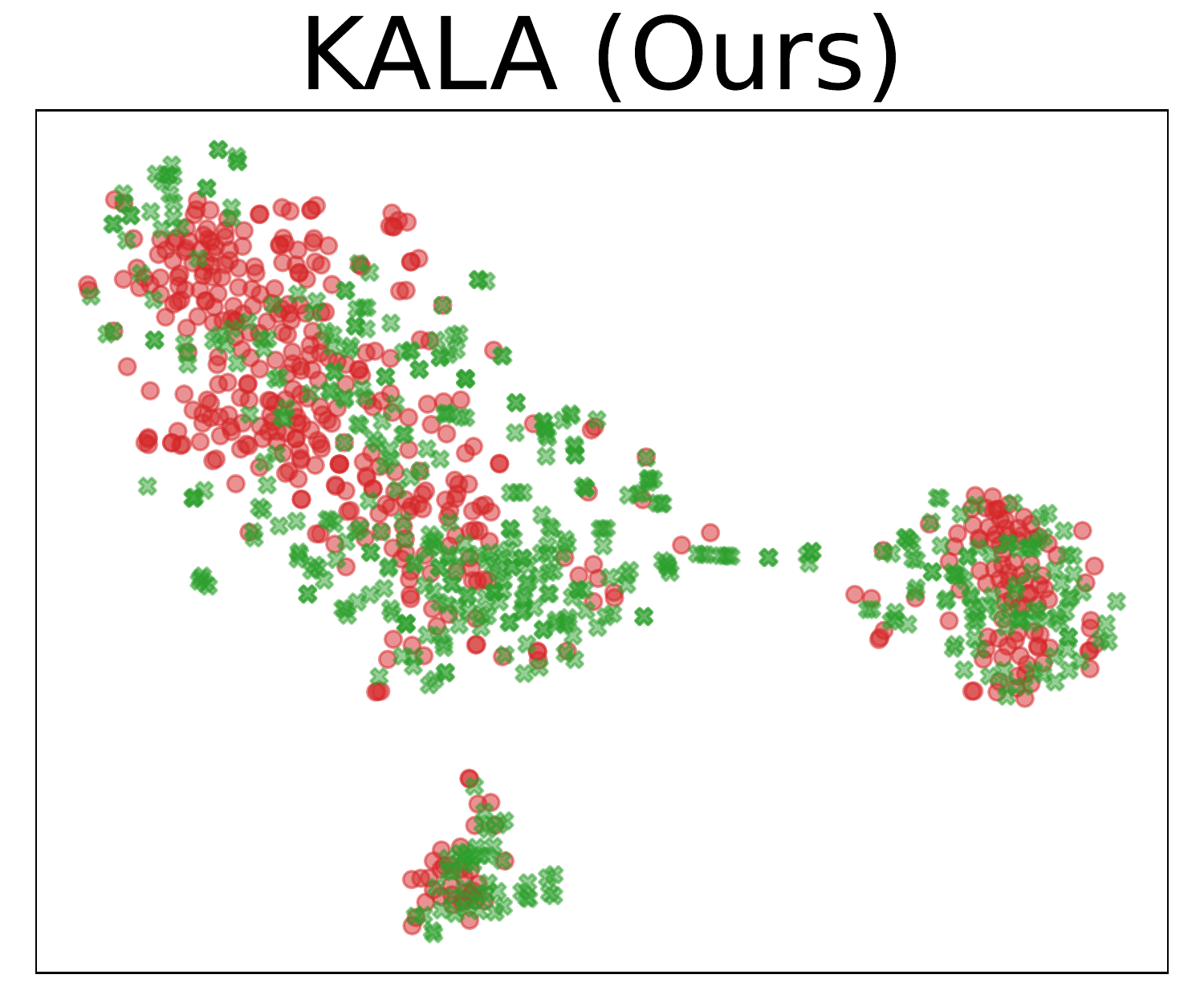}
        \end{subfigure}
    \end{minipage}
    \vskip -0.13in
    \caption{\small \textbf{Concepts (Left).} (a) Adaptive Pre-training updates whole parameters of the PLM through further pre-training on the domain corpus. (b) Our method KALA integrates the external knowledge so that the PLM adapts to the target domain \textbf{only with fine-tuning}, which is realized by the affine transformation on the intermediate feature. 
    \textbf{Visualization of the contextualized representation from the PLM for seen and unseen entities (Right).} Our KALA framework embeds the unseen entities on the embedding space of seen entities by representing them with their relational knowledge over the graph, while the strong DAPT baseline~\cite{dontstoppt} cannot appropriately handle unseen entities that are not given for task fine-tuning.}
    \label{fig:concept}
    \vskip -0.18in
\end{figure*}

This problem, known as \textit{Language Model Adaptation}, can be viewed as a transfer learning problem~\citep{transferlearning, RuderThesis} under domain shift, where the model is pre-trained on the general domain and the labeled distribution is available for the target domain-specific task.
The most prevalent approach to this problem is adaptive pre-training (Figure~\ref{fig:concept}a) which further updates all parameters of the PLM on a large domain-specific or curated task-specific corpus, with the same pre-training strategy (e.g., masked language modeling) before fine-tuning it on the downstream task~\citep{SciBERT, BioBERT, dontstoppt}. 
This continual pre-training of a PLM on the target domain corpus allows it to learn the distribution of the target domain, resulting in improved performance on domain-specific tasks~\citep{TAPT, DAPT}.

While it has shown to be effective, adaptive pre-training has obvious drawbacks. 
First, it is computationally inefficient.
Although a PLM becomes more powerful with the increasing amount of pre-training data~\citep{dontstoppt}, further pre-training on the additional data requires larger memory and computational cost as the dataset size grows~\citep{ProBERT}. 
Besides, it is difficult to adapt the PLM to a new domain without forgetting the general knowledge it obtained from the initial pretraining step, since all pre-trained parameters are continually updated to fit the domain-specific corpus during adaptive pre-training~\citep{RecAdam}. This \emph{catastrophic forgetting} of the task-general knowledge may lead to the performance degradation on the downstream tasks. In Figure~\ref{fig:main}, we show that adaptive pre-training with more training steps could lead to performance degeneration.

Thus, it would be preferable if we could adapt the PLM to the domain-specific task without costly adaptive pre-training. To this end, we aim to integrate the domain-specific knowledge into the PLM directly during the task-specific fine-tuning step, as shown in Figure~\ref{fig:concept}b, eliminating the adaptive pre-training stage.
Specifically, we first note that \textbf{entities} and \textbf{relations} are core building blocks of the domain-specific \textbf{knowledge} that are required to solve for the domain-specific downstream tasks. 
Clinical domain experts, for example, are familiar with medical terminologies and their complex relations. 
Then, to represent the domain knowledge consisting of entities and relations, we introduce the \textit{Entity Memory}, which is the source of entity embeddings but independent of the PLM parameters (See Entity Memory in Figure~\ref{fig:concept}b).
Then, we further exploit the relational structures of the entities by utilizing a Knowledge Graph (KG), which denotes the factual relationships between entities, as shown in Knowledge Graph of Figure~\ref{fig:concept}b.

The remaining step is how to integrate the knowledge into the PLM during fine-tuning. To this end, we propose a novel layer named Knowledge-conditioned Feature  Modulation (KFM, \S\ref{sec:3.2}), which scales and shifts the intermediate hidden representations of PLMs by conditioning them with retrieved knowledge representations.
This knowledge integration scheme has several advantages.
First, it does not modify the original PLM architecture, and thus could be integrated into any PLMs regardless of their architectures. 
Also, it only requires marginal computational and memory overhead, while eliminating the need of excessive further pre-training (Figure~\ref{fig:main}).
Finally, it can effectively handle unseen entities with relational knowledge from the KG, which are suboptimally embedded by adaptive pre-training.
For example, as shown in Figure~\ref{fig:concept}, an entity \textit{restenosis} does not appear in the training dataset for fine-tuning, thus adaptive pre-training only implicitly infers them within the context from the broad domain corpus.
However, we can explicitly represent the unknown entity by aggregating the representations of known entities in the entity memory (i.e., in Figure~\ref{fig:concept}, neighboring entities, such as \textit{asthma} and \textit{pethidine}, are used to represent the unseen entity \textit{restenosis}).

We combine all the previously described components into a novel language model adaptation framework, coined as \textbf{K}nowledge-\textbf{A}ugmented \textbf{L}anguage model \textbf{A}daptation (KALA) (Figure~\ref{fig:overview}). We empirically verify that \textbf{KALA} improves the performance of the PLM over adaptive pre-training on various domains with two knowledge-intensive tasks: Question Answering (QA) and Named Entity Recognition (NER). 
Our contribution is threefold:
\vspace{-0.125in}
\begin{itemize}[itemsep=0.05mm, leftmargin=*]
    \item We propose a novel LM adaptation framework, which augments PLMs with entities and their relations from the target domain, during fine-tuning without any further pre-training. 
    To our knowledge, this is the first work that utilizes the structured knowledge for language model adaptation.
    
    \item To reflect structural knowledge into the PLM, we introduce a novel layer which scales and shifts the intermediate PLM representations with the entity representations contextualized by their related entities according to the KG.

    \item We show that our KALA significantly enhances the model's performance on domain-specific QA and NER tasks, while being significantly more efficient over existing LM adaptation methods.
\end{itemize}

\section{Related Work}
\paragraph{Language Model Adaptation}
Nowadays, transfer learning~\citep{TAPT} is a dominant approach for solving Natural Language Understanding (NLU) tasks.
This strategy first pre-trains a Language Model (LM) on a large and unlabeled corpus, then fine-tunes it on downstream tasks with labeled data~\citep{BERT}.
While this scheme alone achieves impressive performance on various NLU tasks, adaptive pre-training of the PLM on a domain-specific corpus helps the PLM achieve better performance on the domain-specific tasks. 
For example, \citet{BioBERT} demonstrated that a further pre-trained LM on biomedical documents outperforms the original LM on biomedical NLU tasks. 
Also, \citet{dontstoppt} showed that adaptive pre-training of the PLM on the corpus of a target domain (Domain-adaptive Pre-training; DAPT) or a target task (Task-adaptive Pre-training; TAPT) improves its performance on domain-specific tasks. 
However, above approaches generally require a large amount of computational costs for pre-training.

\paragraph{Knowledge-aware LM}
Accompanied with increasing sources of knowledge~\citep{wikidata}, some prior works have proposed to integrate external knowledge into PLMs, to enhance their performance on tasks that require structured knowledge.
For instance, ERNIE~\citep{ERNIE} and KnowBERT~\citep{KnowBERT} incorporate entities as additional inputs in the pre-training stage to obtain a knowledge-aware LM, wherein a pre-trained knowledge graph embedding from Wikidata~\citep{wikidata} is used to represent entities. 
Entity-as-Experts~\citep{EaE} and LUKE~\citep{LUKE} use the entity memory that is pre-trained along with the LMs from scratch.
ERICA~\citep{ERICA} further uses the fact consisting of entities and their relations in the pre-training stage of LMs from scratch.
Previous works aim to integrate external knowledge into the LMs during the pre-training step to obtain a universal knowledge-aware LM that requires additional parameters for millions of entities. In contrast to this, our framework aims to efficiently modify a general PLM for the domain-specific task with a linear modulation layer scheme discussed in Section~\ref{sec:3.2}, during fine-tuning.
\section{Method}
\vspace{-0.05in}
\subsection{Problem Statement}
\label{sec:3.1}
Our goal is to solve Natural Language Understanding (NLU) tasks for a specific domain, with a knowledge-augmented Language Model (LM).
We first introduce the NLU tasks we target, followed by the descriptions of the proposed knowledge-augmented LM.
After that, we formally define the ingredients for structured knowledge integration.

\vspace{-0.05in}
\paragraph{NLU tasks}
The goal of an NLU task is to predict the label $\vy$ of the given input instance $\vx$, where the input $\vx$ contains the sequence of tokens~\citep{BERT}: $\vx = [w_1, w_2, \ldots, w_{|\vx|}]$. 
Then, given a training dataset $\mathcal{D}=\{(\vx^{(i)}, \vy^{(i)})\}_{i=1}^N$, the objective is to maximize the log-likelihood as follows:
\begin{gather*}
    \max_\theta \mathcal{L}(\theta) \coloneqq \max_\theta \sum_{(\vx,\vy)\sim\mathcal{D}} \log p(\vy|\vx; \theta), \\
    p(\vy|\vx; \theta) = g(\mH; \theta_g), \; \mH = f(\vx; \theta_f),
\end{gather*}
where $f$ is an encoder of the PLM which outputs contextualized representation $\mH$ from $\vx$, and $g$ is a decoder which models the probability distribution $p$ of the label $\vy$, with trainable parameters $\theta = \left( \theta_f, \theta_g \right)$.
If the LM is composed of L-layers of transformer blocks~\citep{BERT}, the function $f$ is decomposed to multiple functions $f = [f^{0}, \ldots, f^{L}]$, where each block gets the output of the previous block as the input: $\mH^l = f^l (\mH^{l-1})$.\footnote{$f^{0}$ denotes a word embedding layer which gets $\vx$ as an input, i.e., $\mH^0 = f^0 (\vx)$, for the sake of simplicity.}

\vspace{-0.05in}
\paragraph{Knowledge-Augmented Language Model}
The conventional learning objective defined above might be sufficient for understanding the texts if the tasks require only the general knowledge stored in PLMs. However, it is suboptimal for tackling domain-specific tasks since the general knowledge captured by the parameters $\theta_f$ may not include the knowledge required for solving the domain-specific tasks. Thus, contextualizing the texts by the domain knowledge, captured by the domain-specific entities and their relations, is more appropriate for handling such domain-specific problems.

To this end, we propose a function $h(\cdot;\phi)$ which augments PLMs conditioned on the domain knowledge.
Formally, the objective for a NLU task with our knowledge-augmented LM is given as follows:
\begin{gather*}
    \max_{\theta, \phi} \mathcal{L}(\theta, \phi) \coloneqq \max_{\theta, \phi} \sum_{(\vx,\vy)\sim\mathcal{D}} \log p(\vy|\vx; \theta, \phi), \\
    p(\vy|\vx; \theta, \phi) = g(\tilde{\mH}; \theta_g), \\
    \tilde{\mH}^l = f^l(\mH^{l-1}, h^l(\mH^{l-1}, \mathcal{E}, \mathcal{M}, \mathcal{G}; \phi); \theta_{f^l}),
\end{gather*}
where $\phi$ is parameters for the function $h$, $\mathcal{E}$ is the set of entities, $\mathcal{M}$ is the set of corresponding mentions, and $\mathcal{G}$ is a knowledge graph.
In the following, we will describe the definition of the knowledge-related inputs $\mathcal{E}, \mathcal{M}, \mathcal{G}$, and the details of $h(\cdot, \phi)$.

\vspace{-0.05in}
\paragraph{Definition 1 (Entity and Mention).}
Given a sequence of tokens $\vx = [w_1, \ldots, w_{|\vx|}]$, let $\mathcal{E}$ be a set of entities in $\vx$.
Then an \textbf{entity} $e \in \mathcal{E}$ is composed of one or multiple adjacent tokens within the input text: $[w_{m^\alpha}, \ldots, w_{m^\omega}] \sqsubseteq \vx$\footnote{$E \sqsubseteq E'$ iff $E = E'$, or $E$ is included in $E'$ such that the order of elements in $E$ and $E'$ is the same.}. 
Here, $m=(m^{\alpha}, m^{\omega})$ is a \textbf{mention} that denotes the start and end locations for the entity within the input tokens $\vx$, which term is commonly used for defining entities~\citep{EaE}.
Consequently, for each given input $\vx^{(i)}$, there are a set of entities $\mathcal{E}^{(i)} = \{e_1, \ldots, e_K\}$ and their corresponding mentions $\mathcal{M}^{(i)} = \{m_1, \ldots, m_K\}$.
For example, given an input $\vx = [\text{New, York, is, a, city}]$, we have two entities $\mathcal{E} = \{ \text{\textit{New\_York}, \text{\textit{city}}} \}$ and their associated mentions $\mathcal{M} = \{(1, 2), (4, 4)\}$.

We further construct the entity vocabulary $\mathcal{E}_{\text{train}} = \bigcup_{i=1}^N \mathcal{E}^{(i)}$, which consists of all entities appearing in the training dataset. However, at test time, we may encounter unseen entities that are not in $\mathcal{E}_{\text{train}}$. To tackle this, we regard unknown entities as the null entity $e_\emptyset$, so that $\forall e \in \mathcal{E}_{\text{train}} \cup \{e_\emptyset \}$.

\vspace{-0.05in}
\paragraph{Definition 2 (Entity Memory).}
Given a set of all entities $\mathcal{E}_{\text{train}} \cup \{e_\emptyset \}$, we represent them in the continuous vector (feature) space to learn meaningful entity embeddings.
In order to implement this, we define the \textbf{entity memory} $\mE \in \mathbb{R}^{(|\mathcal{E}_{\text{train}}|+1) \times d}$ that comprises of an entity $e \in \mathbb{R}$ as a key and its embedding $\ve \in \mathbb{R}^{d}$ as its value. Also, to access the value in the entity memory, we define the point-wise memory access function \texttt{EntEmbed} which takes an entity as an input. For instance, $\ve =$ \texttt{EntEmbed}(\textit{New\_York}) returns the embedding of the \textit{New\_York} entity, and $\ve =$ \texttt{EntEmbed}($e_\emptyset$) returns the zero embedding.
This entity memory $\mE$ is the part of the parameter $\phi$ used in function $h$.

\vspace{-0.05in}
\paragraph{Definition 3 (Knowledge Graph).}
Since the entity memory alone cannot represent relational information between entities, we further define a \textbf{Knowledge Graph} (KG) $\mathcal{G}$ that consists of a set of factual triplets $\{(h, r, t)\}$, where the head and the tail entities, $h$ and $t$, are the elements of $\mathcal{E}$, and a relation $r$ is an element of a set of relations $\mathcal{R}$: $h, t \in \mathcal{E}$ and $r \in \mathcal{R}$. We assume that a pre-constructed KG $\mathcal{G}^{(i)}$ is \textbf{given} for each input $x^{(i)}$, and provide the details of the KGs and how to construct them in \textbf{Appendix}~\ref{appendix:kg}.

\begin{figure*}[ht]
    \centering
    \includegraphics[width=1.0\linewidth]{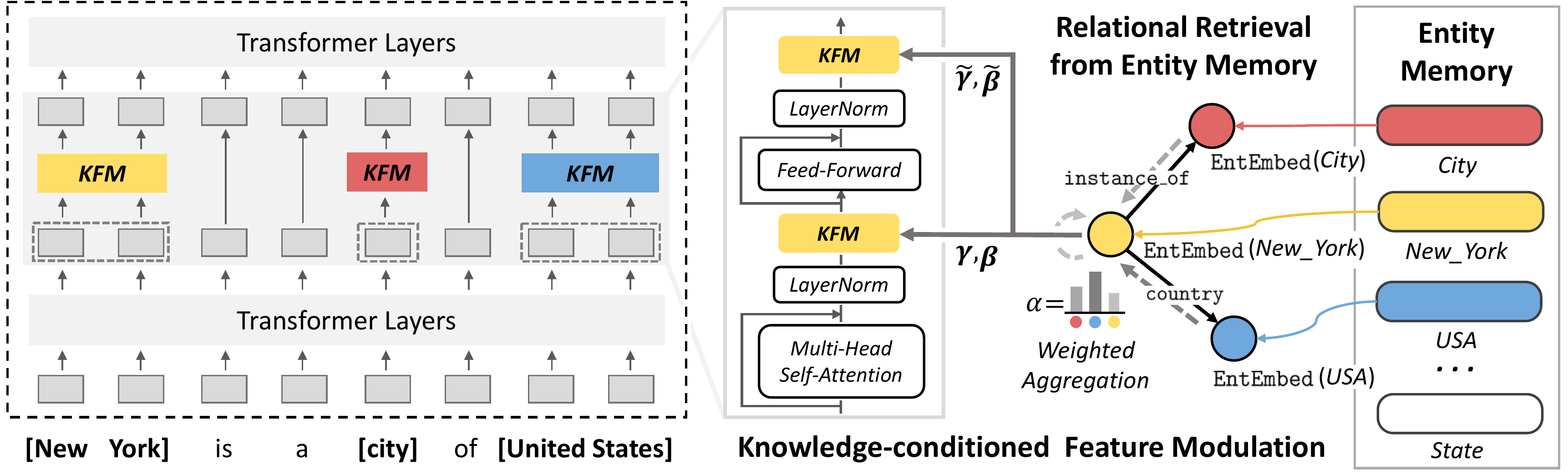}
    \vskip -0.1in
    \caption{\small \textbf{Framework Overview. (Left)} The architecture of a knowledge-augmented LM with our method. Some of the input tokens are annotated as entities with their mentions. 
    \textbf{(Middle)} Inside the transformer block, KFM (\S\ref{sec:3.2}) is applied after the layer normalization as in equation~\ref{eqn:film}, to modulate the hidden representations of tokens in entity mentions.
    \textbf{(Right)} The retrieved embedding of an entity \textit{New\_York} is composed by the weighted aggregation of neighbors through the knowledge graph (\S\ref{sec:3.3}).}
    \vskip -0.1in
    \label{fig:overview}
\end{figure*}

\subsection{Knowledge-conditioned Feature Modulation on Transformer}
\label{sec:3.2}
The remaining problem is how to augment a PLM by conditioning it on the domain-specific knowledge, through the function $h$. 
An effective approach to do so without stacking additional layers on top of the LM is to interleave the knowledge from $h$ with the pre-trained parameters of the language model~\citep{BERT} consisting of transformer layers~\citep{Transformer}. 
Before describing our interleaving method in detail, we first describe the Transformer architecture.

\paragraph{Transformer} 
Given $|x|$ token representations $\mH^{l-1} = [\vh^{l-1}_1, \ldots, \vh^{l-1}_{|\vx|}] \in \mathbb{R}^{{|\vx|} \times d}$ from the layer $l-1$ where $d$ is the embedding size, each transformer block outputs the contextualized representations for all tokens. In detail, the $l$-th block consists of the multi-head self-attention (\textit{Attn}) layer and the residual feed-forward (\textit{FF}) layer as follows:
\begin{align*}
\begin{split}
    &\hat{\mH}^{l} = LN(\mH^{l-1} + Attn(\mH^{l-1})) \\
    &FF(\hat{\mH}^{l}) = \sigma(\hat{\mH}^{l} \cdot \mW_1) \cdot \mW_2, \\
    &\mH^l = LN(\hat{\mH}^{l} + FF(\hat{\mH}^{l})),
\end{split}
\end{align*}
where $LN$ is a layer normalization~\citep{LayerNorm}, $\sigma$ is an activation function~\citep{GELU}, $\mW_2 \in \mathbb{R}^{d' \times d}$ and $\mW_1 \in \mathbb{R}^{d \times d'}$ are weight matrices, and $d'$ is an intermediate hidden size. We omit the bias term for brevity.

\paragraph{Linear Modulation on Transformer} 
An effective yet efficient way to fuse knowledge from different sources without modifying the original model architecture is to scale and shift the features of one source with respect to the data from another source~\citep{Featurewise}.
This scheme of feature-wise affine transformation is effective on various tasks, such as language-conditioned image reasoning~\citep{FiLM} or style-transfer in image generation~\citep{adain}.

Motivated by them, we propose to linearly transform the intermediate features after the layer normalization of the transformer-based PLM, conditioned on the knowledge sources $\mathcal{E}, \mathcal{M}, \mathcal{G}$. We term this method as the \textbf{Knowledge-conditioned Feature Modulation} (KFM), described as follows:
\begin{align}
\begin{split}
    &\bm{\Gamma}, \bm{B}, \tilde{\bm{\Gamma}}, \tilde{\bm{B}} = h^l (\mH^{l-1}, \mathcal{E}, \mathcal{M}, \mathcal{G}; \phi), \\
    &\hat{\mH}^{l} = \bm{\Gamma} \circ LN(\mH^{l-1} + Attn(\mH^{l-1})) + \bm{B}, \\
    &FF(\hat{\mH}^{l}) = \sigma(\hat{\mH}^{l} \cdot \mW_1) \cdot \mW_2, \\
    &\tilde{\mH}^{l} = \tilde{\bm{\Gamma}} \circ LN(\hat{\mH}^{l} + FF(\hat{\mH}^{l})) + \tilde{\bm{B}},
\end{split}
\label{eqn:film}
\raisetag{15pt}
\end{align}
where $\mH^{l-1} \in \mathbb{R}^{|x| \times d}$ is the matrix of hidden representations from the previous layer, $\circ$ denotes the hadamard (element-wise) product, and $\bm{\Gamma} = [\bm{\gamma}_1, \ldots, \bm{\gamma}_{|\vx|}] \in \mathbb{R}^{{|\vx|} \times d}$, $\bm{B} = [\bm{\beta}_1, \ldots, \bm{\beta}_{|\vx|}] \in \mathbb{R}^{{|\vx|} \times d}$. 
$\bm{\Gamma}$ and $\bm{B}$ are learnable modulation parameters from the function $h$, which are conditioned by the entity representation.
For instance, in Figure~\ref{fig:overview}, $\bm{\gamma}$ and $\bm{\beta}$ for token `New' are conditioned on the corresponding entity \textit{New\_York}.
However, if tokens are not part of any entity (e.g., `is'), $\bm{\gamma}$ and $\bm{\beta}$ for such tokens are fixed to $\boldsymbol{1}$ and $\boldsymbol{0}$, respectively.

One notable advantage of our KFM is that multiple tokens associated to the identical entity are affected by the same modulation (e.g., `New' and `York' in Figure~\ref{fig:overview}), which allows the PLM to know which adjacent tokens are in the same entity. This is important for representing the tokens of the domain entity (e.g., `cod' and `on'), since the original PLM might regard them as separate, unrelated tokens (See analysis in \S\ref{sec:case_study} with Figure~\ref{fig:casestudy}). However, with our KFM, the PLM can identify associated tokens and embed them to be close to each other.

Then, how can we design such functional operations in $h$? The easiest way is to retrieve the entity embedding of $e$, associated to the typical token, from the entity memory $\mE$, and then use the retrieved entity embedding as the input to obtain $\bm{\gamma}$ and $\bm{\beta}$ for every entity (See Figure~\ref{fig:overview}). Formally, for each entity $e \in \mathcal{E}$ and its mention $(m^{\alpha}, m^{\omega}) \in \mathcal{M}$,
\begin{align}
\label{eqn:entembed}
    &\vv = \texttt{EntEmbed}(e) \\
    &\bm{\gamma}_j = \boldsymbol{1} + h_1(\vv), \  \bm{\beta}_j = h_2(\vv), \nonumber \\ 
    &\tilde{\bm{\gamma}}_j = \boldsymbol{1} + h_3(\vv), \  \tilde{\bm{\beta}}_j = h_4(\vv),\quad m^{\alpha} \le j \le m^{\omega}, \nonumber
\end{align}
where $\vv$ is the retrieved entity embedding from the entity memory, $h_1, h_2, h_3,$ and $h_4$ are mutually independent Multi-Layer Perceptrons (MLPs) which return a zero vector $\boldsymbol{0}$ if $e = e_\emptyset$.

\subsection{Relational Retrieval from Entity Memory}
\label{sec:3.3}
Although the simple access to the entity memory can retrieve the necessary entity embeddings for the modulation, this approach has obvious drawbacks as it not only fails to reflect the relations with other entities, but also regards unseen entities as the same null entity $e_\emptyset$. 
If so, all unseen entities are inevitably modulated by the same parameters even if they have essentially different meaning.

To tackle these limitations, we further consider the relational information between two entities that are linked with a particular relation. 
For example, the entity \textit{New\_York} alone will not give meaningful information. However, with two associated facts (\textit{New\_York}, \textit{instance of}, \textit{city}) and (\textit{New\_York}, \textit{country}, \textit{USA}), it is clear that \textit{New\_York} is a city in the \textit{USA}.
Motivated by this observation, we propose \textbf{Relational Retrieval} which leverages a KG $\mathcal{G}$ to retrieve entity embeddings from the memory, according to the relations defined in the given KG (See Figure~\ref{fig:overview}, right).

More specifically, our goal is to effectively utilize the relations among entities in $\mathcal{G}$, to improve the \texttt{EntEmbed} function in equation~\ref{eqn:entembed}. We tackle this objective by utilizing a \emph{Graph Neural Network (GNN)} which learns feature representations of each node using a neighborhood aggregation scheme~\cite{GraphSAGE}, as follows: 
\begin{align*}
    \vv = &\texttt{ UPDATE}( \texttt{EntEmbed}(e),\\
          &\texttt{ AGG}(\{ \texttt{EntEmbed}(\hat{e}): \forall \hat{e} \in \mathcal{N}(e;\mathcal{G}) \})),
\end{align*}
where $\mathcal{N}(e;\mathcal{G})$ is a set of neighboring entities of the entity $e$, \texttt{AGG} is the function that aggregates embeddings of neighboring entities of $e$, and \texttt{UPDATE} is the function that updates the representation of $e$ with the aggregated messages from \texttt{AGG}. 

However, simple aggregation (e.g., mean) cannot reflect the relative importance on neighboring nodes, thus we consider the attentive scheme~\cite{GAT, GATv2} for neighborhood aggregation, to allocate weights to the target entity's neighbors by their importance. This scheme is helpful in filtering out less useful relations. Formally, we first define a scoring function $\psi$ that calculates a score for every triplet $(e_i, r_{ij}, e_j)$, which is then used to weigh each node during aggregation:
\begin{gather*}
    \ve_i = \texttt{EntEmbed}(e_i), \; \ve_j = \texttt{EntEmbed}(e_j),\\
    \ve^* = [\ve_i\parallel\vr_{ij}\parallel\ve_j\parallel\vh_{e_i}], \\
    \psi(\ve_i, \vr_{ij}, \ve_j, \vh_{e_i}) = \va^\top \sigma(\mW \cdot \ve^*),
\end{gather*}
where $\sigma$ is a nonlinear activation, $\ve^* \in \mathbb{R}^{4d}$ is concatenated vector where $\parallel$ denotes the concatenation, $\va \in \mathbb{R}^d$ and  $\mW \in \mathbb{R}^{d \times 4d}$ are learnable parameters, $\vr_{ij} \in \mathbb{R}^{d}$ is a embedding of the relation, and $\vh_{e_i} \in \mathbb{R}^d$ is a context representation of the entity $e_i$ obtained from the intermediate hidden states of the LM\footnote{The context representation of the entity is calculated with its mention as follows: $\vh_e = \frac{1}{m^\omega-m^\alpha+1} \sum_{i=m^\alpha}^{m^\omega} \vh^{l-1}_i$}.

The scores obtained from $\psi$ are normalized across all neighbors $e_j \in \mathcal{N}(e_i;\mathcal{G})$ with softmax:
\begin{align*}
    \alpha_{ij} &= \softmax(\psi(\ve_i, \vr_{ij}, \ve_j)) \\
                &= \frac{\exp(\psi(\ve_i, \vr_{ij}, \ve_j))}
                        {\sum_{e_{j'} \in \mathcal{N}(e_i;\mathcal{G})} \exp(\psi(\ve_{i}, \vr_{ij'}, \ve_{j'}))}.
\end{align*}

Then, we update the entity embedding with a weighted average of the neighboring nodes with $\alpha$ as an attention coefficient, denoted as follows:
\begin{equation}
    \vv = \texttt{UPDATE} \left( {\textstyle\sum}_{e_{j'} \in \mathcal{N}(e_i;\mathcal{G})} \alpha_{ij} \cdot \ve_{j'} \right).
\label{eqn:relembed}
\end{equation}
By replacing the \texttt{EntEmbed} function in equation~\ref{eqn:entembed} with the above GNN in equation~\ref{eqn:relembed},
we now represent each entity with its relational information in KG.
This relational retrieval has several advantages over simple retrieval of a single entity from the entity memory. 
First, the relational retrieval with KG can consider richer interactions among entities, as described in Figure~\ref{fig:overview}.

In addition, we can naturally represent an unseen entity -- which is not seen during training but appears at test time -- through neighboring aggregation, which is impossible only with the entity memory. In Figure~\ref{fig:concept}, we provide an illustrative example of the unseen entity representation, where the unseen entity \textit{restenosis} is represented with a weighted sum of representations of its neighboring entities \textit{myocardial\_infarction}, \textit{asthma}, and \textit{pethidine}, which is beneficial when the set of entities for training and test datasets have small overlaps.
\section{Experiment}

\subsection{Tasks and Datasets}
We evaluate our model on two NLU tasks: Question Answering (QA) and Named Entity Recognition (NER).
\textbf{For QA}, we use three domain-specific datasets: NewsQA (News,~\citealp{NewsQA}) and two subsets (Relation, Medication) of EMRQA (Clinical,~\citealp{emrQA}).
We use the Exact-Match (EM) and the F1 score as evaluation metrics.
\textbf{For NER}, we use three datasets from different domains, namely CoNLL-2003 (News,~\citealp{CoNLL}), WNUT-17 (Social Media,~\citealp{WNUT}) and NCBI-Disease (Biomedical,~\citealp{NCBI}). 
We use the F1 score as the evaluation metric. We report statistics and detailed descriptions of each dataset in \textbf{Appendix}~\ref{appendix:dataset}.

\subsection{Baselines}
A direct baseline of our KALA is the adaptive pre-training, which is commonly used to adapt the PLM independent to the choice of a domain and task. Also, to compare ours against a more powerful baseline, we modify a recent method~\citep{RecAdam} that alleviates forgetting of PLM during fine-tuning. Details for each baseline we use are described as follows:

\vspace{-0.05in}
\begin{enumerate}[itemsep=1.0mm, parsep=0pt, leftmargin=*]
    \item \textbf{Vanilla Fine-Tuning (FT)}: A baseline that directly fine-tunes the LM on downstream tasks.
    
    \item \textbf{Fine-Tuning +} \textit{more params}: A baseline with one more transformer layer at the end of the LM. We use this baseline to show that the performance gain of our model does not come from the use of additional parameters. 
    
    \item \textbf{Task-Adaptive Pre-training (TAPT)}: A baseline that further pre-trains the PLM on task-specific corpus as in~\citet{dontstoppt}.
    
    \item \textbf{TAPT + RecAdam}: A baseline that uses RecAdam~\citep{RecAdam} during further pre-training of PLMs (i.e., TAPT), to alleviate catastrophic forgetting of the learned general knowledge in PLMs from adaptive pre-training.
    
    \item \textbf{Domain-Adaptive Pre-training (DAPT)}: A strong baseline that uses a large-scale domain corpus outside the training set during further pre-training~\citep{dontstoppt}, and requires extra data and large computational overhead.
    
    \item \textbf{KALA} (pointwise): A variant of KALA that only uses the \textbf{entity memory} and does not use the knowledge graphs.
    
    \item \textbf{KALA} (relational): Our full model that uses KGs to perform relational retrieval from the entity memory.
\end{enumerate}

\begin{table*}
	\centering
	\small
	\resizebox{0.95\textwidth}{!}{
	\begin{tabular}{lccc}
		\toprule
		{\textbf{Method}} & \textbf{NewsQA} & {\textbf{Relation}} & {\textbf{Medication}} \\
		\midrule[0.8pt]
		\textbf{Fine-Tuning} & {53.06 $\pm$ 0.63  | 67.20 $\pm$ 0.19} & {54.01 $\pm$ 1.14 | 61.43 $\pm$ 1.18} & {12.50 $\pm$ 0.28 | 43.31 $\pm$ 0.67}  \\
		\textbf{+} \textit{more params} & {53.59 $\pm$ 0.99 | 67.79 $\pm$ 0.67} & {54.06 $\pm$ 1.35 | 62.07 $\pm$ 1.44} & {12.46 $\pm$ 0.25 | 42.74 $\pm$ 0.91}  \\
		\textbf{TAPT} & {53.47 $\pm$ 1.69 | 67.59 $\pm$ 1.44} & {53.57 $\pm$ 2.05 | 60.87 $\pm$ 2.52} & {12.58 $\pm$ 0.42 | 43.82 $\pm$ 1.10}  \\
		\textbf{+ RecAdam} & {53.95 $\pm$ 1.02 | 67.89 $\pm$ 0.75} & {54.88 $\pm$ 1.94 | 62.54 $\pm$ 2.14} & {12.63 $\pm$ 0.30 | 43.86 $\pm$ 0.87} \\
		\textbf{DAPT}$^\dagger$ & {53.68 $\pm$ 0.94 | 67.76 $\pm$ 0.61} & {55.29 $\pm$ 1.74 | 62.25 $\pm$ 1.80} & {12.67 $\pm$ 0.27 | 43.26 $\pm$ 0.88} \\
		\midrule[0.5pt]
		\textbf{KALA} \textit{(point-wise)} & {53.41 $\pm$ 0.74 | 67.30 $\pm$ 0.45} & {\textbf{56.13} $\pm$ 0.85 | \textbf{64.69} $\pm$ 0.92} & {12.01 $\pm$ 0.47 | 42.97 $\pm$ 0.70} \\
		\textbf{KALA} \textit{(relational)} & {\textbf{54.25} $\pm$ 0.63 | \textbf{68.27} $\pm$ 0.63} & {55.96 $\pm$ 1.37 | 64.22 $\pm$ 1.15} & {\textbf{12.75} $\pm$ 0.61 | \textbf{44.19} $\pm$ 0.46}  \\
		\bottomrule
	\end{tabular}
	}
 	\vspace{-0.075in}
	\caption{\small Experimental results of the extractive QA task on three different datasets with the BERT-base. The reported results are means and standard deviations of performances over five different runs with Exact Match / F1 score as a metric. The numbers in bold fonts denote the best score. $\dagger$ indicates the method under an \textbf{extremely high} computational resource setting (See Figure~\ref{fig:main}).}
	\label{qa-exp}
 	\vspace{-0.1in}
\end{table*}

\begin{table}
	\centering
	\small
	\resizebox{1.0\columnwidth}{!}{
	\begin{tabular}{lccc}
		\toprule
		{\textbf{Method}} & {\textbf{CoNLL}-2003} & {\textbf{WNUT}-17} & {\textbf{NCBI}-Disease} \\
		\midrule[0.8pt]
		\textbf{Fine-Tuning} & {90.58 $\pm$ 0.19} & {45.70 $\pm$ 1.25} & {84.42 $\pm$ 0.58} \\
		\textbf{+} \textit{more params} & {90.75
		$\pm$ 0.23} & {46.42
		$\pm$ 0.55} & {84.70 $\pm$ 0.49} \\
		\textbf{TAPT} & {90.61
		$\pm$ 0.73} & {45.39 $\pm$ 0.77} & {84.39 $\pm$ 0.73} \\
		\textbf{+ RecAdam} & {90.69
		$\pm$ 0.30} & {46.73 $\pm$ 0.94} & {84.99 $\pm$ 0.88} \\
		\textbf{DAPT}$^\dagger$ & {90.30
		$\pm$ 0.39} & {48.29 $\pm$ 1.08} & {84.68 $\pm$ 1.63} \\
		\midrule[0.5pt]
		\textbf{KALA} \textit{(point-wise)} & {90.96 $\pm$ 0.21} & {47.33 $\pm$ 0.82} & {85.10 $\pm$ 0.73} \\
		\textbf{KALA} \textit{(relational)} & {\textbf{91.02} $\pm$ 0.29} & {\textbf{48.35} $\pm$ 0.92} & {\textbf{85.77} $\pm$ 0.43} \\
		\bottomrule
	\end{tabular}
	}
 	\vspace{-0.1in}
	\caption{\small Experimental results of the NER task on three different datasets with the BERT-base. The reported results are means and standard deviations over five different runs with an F1 score as a metric. The numbers in bold fonts denote the best score. $\dagger$ indicates the baseline under an \textbf{extremely high} computational resource setting (See Figure~\ref{fig:main}).}
	\label{ner-exp}
 	\vspace{-0.1in}
\end{table}

\subsection{Experimental Setup}
We use the uncased BERT-base~\citep{BERT} as the base PLM for all our experiments on QA and NER tasks.
For more details on training and implementation, please see the \textbf{Appendix}~\ref{appendix:setup}.

\subsection{Experimental Results}

\paragraph{Performance on QA and NER tasks} 
On both extractive QA and NER tasks, our KALA outperforms all baselines, including TAPT and TAPT+RedcAdam~\citep{dontstoppt, RecAdam}, as shown in Table~\ref{qa-exp} and~\ref{ner-exp}. These results show that our KALA is highly effective for the language model adaptation task.
KALA also largely outperforms DAPT~\citep{dontstoppt} which is trained with extra data and requires a significantly higher computational cost compare to KALA (See Figure~\ref{fig:main} for the plot of efficiency, discussed in Section~\ref{sec:efficiency}).

\paragraph{Effect of Using more Parameters}
One may suspect whether the performance of our KALA comes from the increment of parameters. However, the experimental results in Table~\ref{qa-exp} and~\ref{ner-exp} show that increasing the parameters for PLM during fine-tuning (+ more params) yields marginal performance improvements over naive fine-tuning.
This result confirms that the performance improvement of KALA is not due to the increased number of parameters.

\paragraph{Importance of Relational Retrieval}
The performance gap between KALA (relational) and KALA (point-wise) shows the effectiveness of relational retrieval for language model adaptation, which allows us to incorporate relational knowledge into the PLM. The relational retrieval also helps address unseen entities, as discussed in Section~\ref{sec:unseen}. 
\section{Analysis and Discussion}

\subsection{Ablation Studies}
\label{sec:5.1}
We perform an ablation study to see how much each component contributes to the performance gain.
\paragraph{KFM Parameters}
We first analyze the effect of feature modulation parameters (i.e., gamma and beta) in transformers by ablating a subset of them in Table~\ref{table:ablationfilm}, in which we observe that using both gamma and beta after both layer normalization on a transformer layer obtains the best performance.

\begin{figure}[!t]
    \begin{minipage}{0.54\linewidth}
        \centering
        \resizebox{1.0\textwidth}{!}{
        \renewcommand{\tabcolsep}{1.0mm}
        \begin{tabular}{lcc}
		\toprule
		\textbf{KFM (\S3.2)}     & \multicolumn{2}{c}{\textbf{NewsQA}}  \\
		\textbf{Components}      & EM          & F1          \\ 
		\midrule[0.8pt]
		\textbf{None (Fine-tuning)} & {53.06} & {67.20}\\
		\textbf{+ $\bm{\Gamma}, \bm{\tilde{\Gamma}}$} \textit{(gamma only)}& {54.10} & {67.98}\\
		\textbf{+ $\mB, \tilde{\mB}$} \textit{(beta only)} & {53.74} &  {67.69}\\
		\textbf{+ $\bm{\Gamma}, \mB$} \textit{(first only)} & {53.77} & {67.88}\\
		\textbf{+ $ \bm{\tilde{\Gamma}}, \tilde{\mB}$} \textit{(second only)} & {53.89} & {67.49}\\
		\midrule[0.8pt]
		\textbf{+ $\bm{\Gamma}, \mB, \bm{\tilde{\Gamma}}, \tilde{\mB}$ (\textbf{final})} & {\textbf{54.25}} &  {\textbf{68.27}}\\
		\bottomrule
    	\end{tabular}
    	}
    	\vspace{-0.08in}
    	\captionof{table}{\small An ablation study of the KFM parameters $\bm{\Gamma}$, $\mB$, $\bm{\tilde{\Gamma}}$, $\tilde{\mB}$. We report the average results over five different runs.}
    	\label{table:ablationfilm}
    \end{minipage}
    \hfill
    \begin{minipage}{0.44\linewidth}
        \centering
        \resizebox{1\textwidth}{!}{
        \renewcommand{\arraystretch}{1.1}
        \renewcommand{\tabcolsep}{1.0mm}
        \begin{tabular}{lcc}
		\toprule
		\textbf{Architecture} & \multicolumn{2}{c}{\textbf{NewsQA}}  \\
		\textbf{Variants (\S5.2)}                   & EM          & F1          \\ 
		\midrule[0.8pt]
		\textbf{ERNIE} & {53.35} &  {67.49}\\
		\textbf{Adapter} & {53.32} & {67.38}\\
		\textbf{KT-Net} & {53.15} & {67.01} \\
		\textbf{EaE} & {53.00} & {67.40}\\ 
		\textbf{ERICA} & {51.99} & {66.40}\\ 
		\midrule[0.8pt]
		\textbf{KALA (ours)} & {\textbf{54.25}} & {\textbf{68.27}}\\
		\bottomrule
    	\end{tabular}
    	}
    	\vspace{-0.08in}
    	\captionof{table}{\small Experimental results on knowledge integration architecture variants, averaged over five runs.}
	\label{table:variants}
	\end{minipage}
    \vspace{-0.15in}
\end{figure}

\paragraph{Architectural Variants}
We now examine the effectiveness of the proposed knowledge conditioning scheme in our KALA framework. To this end, we use or adapt the knowledge integration methods from previous literature, to compare their effectiveness. Specifically, we couple the following five components with KALA: Entity-as-Experts~\citep{EaE}, Adapter~\citep{HoulsbyAdapter}, KT-Net~\citep{KTNet}, ERNIE~\citep{ERNIE}, and ERICA~\citep{ERICA}. Note that, most of them were proposed for improving pre-training from scratch, while we adapt them for fine-tuning under our KALA framework (The details are given in \textbf{Appendix}~\ref{appendix:variant}). As shown in Table~\ref{table:variants}, our KFM used in KALA outperforms all variants, demonstrating the effectiveness of feature modulation in the middle of transformer layers for fine-tuning.

\begin{table*}
	\centering
	\small
	\resizebox{0.95\textwidth}{!}{
	\begin{tabular}{lcccc}
		\toprule
		{\textbf{Method}} & \textbf{NewsQA} & {\textbf{Relation}} & \textbf{WNUT}-17 & \textbf{NCBI}-Disease \\
		\midrule[0.8pt]
		\textbf{Fine-Tuning} & {57.21 $\pm$ 0.56 | 71.91 $\pm$ 0.35} & {46.61 $\pm$ 2.75 | 53.89 $\pm$ 2.92} & {55.00 $\pm$ 1.66} & {86.91 $\pm$ 1.08}  \\
		\textbf{+} \textit{more params} & {\textbf{58.07} $\pm$ 1.19 | 72.38 $\pm$ 1.04} & {45.12 $\pm$ 0.86 | 53.22 $\pm$ 1.27} & {56.62 $\pm$ 0.26} & {87.21 $\pm$ 0.26}  \\
		\textbf{TAPT} & {57.24 $\pm$ 0.53 | 71.77 $\pm$ 0.34} & {45.66 $\pm$ 2.20 | 53.23 $\pm$ 2.38} & {55.46 $\pm$ 1.90} & {86.24 $\pm$ 0.76}  \\
		\midrule[0.5pt]
		\textbf{KALA} \textit{(relational)} & {58.01 $\pm$ 0.57 | \textbf{72.70} $\pm$ 0.25} & {\textbf{47.40} $\pm$ 1.67 | \textbf{55.13} $\pm$ 1.26} & {\textbf{56.96} $\pm$ 0.27} & {\textbf{87.72} $\pm$ 0.27}\\
		\bottomrule
	\end{tabular}
	}
 	\vspace{-0.075in}
	\caption{\small Experimental results of the extractive QA and NER tasks on four different datasets -- NewsQA, Relation, WNUT-17 and NCBI-Disease -- with the RoBERTa-base. The reported results are means and standard deviations over five different runs. We use Exact Match and F1 score as a metric for QA, and F1 score for NER. The numbers in bold fonts denote the best score.}
	\label{table:roberta}
 	\vspace{-0.1in}
\end{table*}
\subsection{Robustness to Other PLMs}
Although we believe our experimental results on Table~\ref{qa-exp},~\ref{ner-exp} with BERT~\cite{BERT} are enough to show the effectiveness of KALA across different pre-trained language models (PLMs), one might be curious that KALA can work on even other PLMs such as RoBERTa~\citep{RoBERTa}. Thus, to address such concerns, we additionally conduct experiments on RoBERTa. As shown in Table~\ref{table:roberta}, we observe that our KALA outperforms all baselines except for one case (Fine-Tuning + \textit{more params} on NewsQA). Thus, we believe that our KALA would be useful to any PLMs, not depending on specific PLMs.

\begin{figure}[t!]
    \begin{minipage}{0.325\linewidth}
        \centering
        \includegraphics[width=1\linewidth]{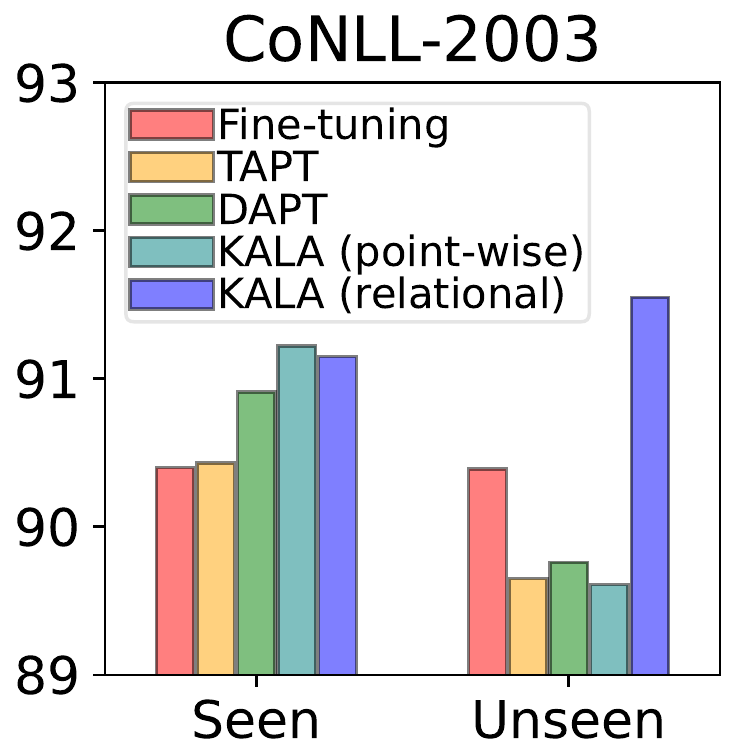}
    \end{minipage}
    \begin{minipage}{0.325\linewidth}
        \centering
        \includegraphics[width=1\linewidth]{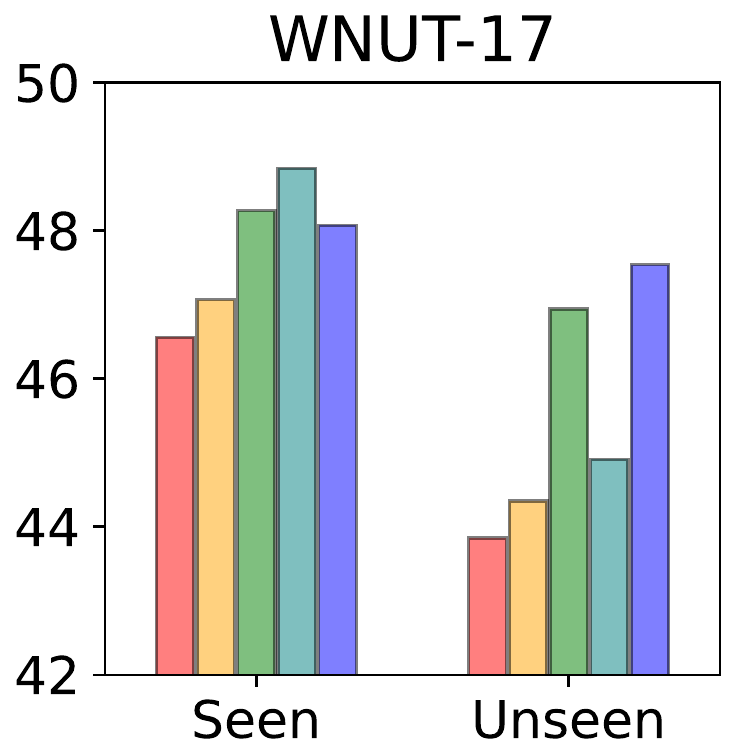}
    \end{minipage}
    \begin{minipage}{0.325\linewidth}
        \centering
        \includegraphics[width=1\linewidth]{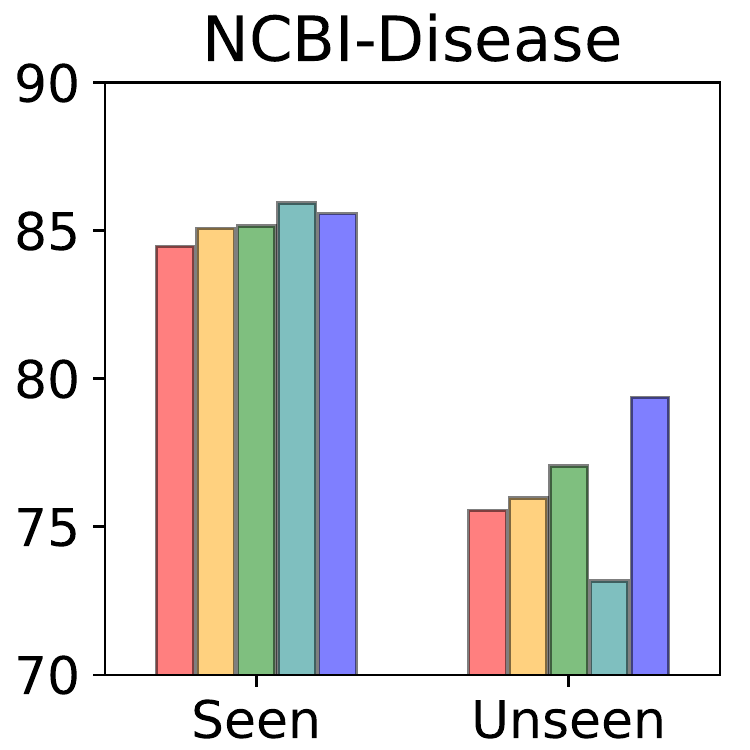}
    \end{minipage}
    \vspace{-0.1in}
    \caption{\small Results on seen and unseen, where Seen denotes the context having less than 3 unseen entities, otherwise Unseen. Note that DAPT uses extra datasets in addition to the training dataset, thus the Unseen for other models could be considered as the Seen for DAPT.}
    \label{fig:unseen_bar}
    \vspace{-0.175in}
\end{figure}

\subsection{Efficiency}
\label{sec:efficiency}
Figure~\ref{fig:main} illustrates the performance and training FLOPs of KALA against baselines on the NewsQA dataset.
We observe that the performance of TAPT decreases with the increased number of iterations, which could be due to forgetting of the knowledge from the PLM. On the other hand, DAPT, while not suffering from performance loss, requires huge computational costs as it trains on 112 times larger data for further pre-training (See \textbf{Appendix}~\ref{appendix:training} for detailed explanations on training data).
On the other hand, our KALA outperforms DAPT without using external data, while requiring 17 times fewer computational costs, which shows that KALA is not only effective but also highly efficient.

To further compare the efficiency in various aspects, we report GPU memory, training wall time, and training FLOPs for baselines and ours in Table~\ref{table:efficiency}. Through this, we verify that our KALA is more efficient to train for language model adaptation settings than baselines.
Note that the resource requirement of KALA could be further reduced by adjusting the size of the entity memory (e.g., removing less frequent entities).
Therefore, to show the flexibility of our KALA on the typical resource constraint, we provide the experimental results on two different settings (i.e., tuning the number of entities in the entity memory) -- KALA with memory size of 200 and 62.8k (full memory) in \textbf{Appendix}~\ref{appendix:efficiecy-detail}.

\begin{table}
	\centering
	\small
	\resizebox{0.975\columnwidth}{!}{
		\begin{tabular}{lcccc}
		\toprule
		{\textbf{Method}} &   GPU Mem.  & Approx. Wall Time  & FLOPs ($10^{16}$) \\
		\midrule[0.8pt]
		\textbf{Fine-Tuning}  & {8 GB} & {3 hrs} & {$9.5$} \\
		+ \textit{more params}   & {8.8 GB} & {3 hrs} & {$10.1$} \\
		\textbf{TAPT}  & {8 GB} & {3.8 hrs} & {$10.1$} \\
		\textbf{DAPT}  & {48 GB} & {40 hrs <} & {$182.0$} \\
		\midrule[0.8pt]
		\textbf{KALA} (ours, 0.2k) & {8.4 GB} & {3 hrs} & {$9.97$} \\
		\textbf{KALA} (ours, 62.8k) & {9.2 GB} & {3 hrs} & {$10.5$} \\
		\bottomrule
	\end{tabular}
	}
	\vspace{-0.05in}
	\captionof{table}{\small Efficiency comparisons of GPU memory, Wall Time, and FLOPs on the NewsQA dataset. The number 0.2k and 62.8k indicate the size of entity memory used in our KALA.}
	\vspace{-0.15in}
	\label{table:efficiency}
\end{table}

\subsection{Effectiveness on Unseen Entities}
\label{sec:unseen}
One remarkable advantage of our KALA is its ability to represent an unseen entity by aggregating features of its neighbors from a given KG. To analyze this, we first divide all contexts into one of Seen and Unseen, where Seen denotes the context with less than 3 unseen entities, and then measure the performance on the two subsets. 
As shown in Figure~\ref{fig:unseen_bar}, we observe that the performance gain of KALA over the baselines is much larger on the Unseen subset, which demonstrates the effectiveness of KALA's relational retrieval scheme to represent unseen entities.
DAPT also largely outperforms fine-tuning and TAPT as it is trained on an extremely large external corpus for adaptive pre-training.
However, KALA even outperforms DAPT in most cases, verifying that our knowledge-augmentation method is more effective for tackling domain-specific tasks.
The visualization of embeddings of seen and unseen entities in Figure~\ref{fig:concept} shows that KALA embeds the unseen entities more closely to the seen entities\footnote{We quantitatively measure the mean of cosine distance of each unseen entity to its nearest seen entity, observing that KALA embeds unseen 1.5 times more closer to seen than DAPT (i.e., 0.07 for KALA vs 0.11 for DAPT for distance).}, which explains KALA's good performance on the Unseen subset.

\begin{figure*}[t!]
\begin{minipage}{0.72\linewidth}
    \begin{minipage}{0.42\linewidth}
        \centering
        \resizebox{\linewidth}{!}{
            \begin{tabular}{lccl}
            \toprule
            \makecell[l]{Context:
                    A \BLUE{nonsense mutation} in \\
                    \BLUE{exon} 17 ( \BLUE{codon} 556 ) of the RB1 \\
                    \BLUE{gene} was found to be present \\
                    homozygously in both the \RED{retinal} \\ 
                    and the pineal tumours.
            } \\
            \midrule
            Fact: (retinal, instance of, gene) \\
            \bottomrule
            \end{tabular}
        }
    \end{minipage}
    \begin{minipage}{0.28\linewidth}
        \centering
        \includegraphics[width=1\linewidth]{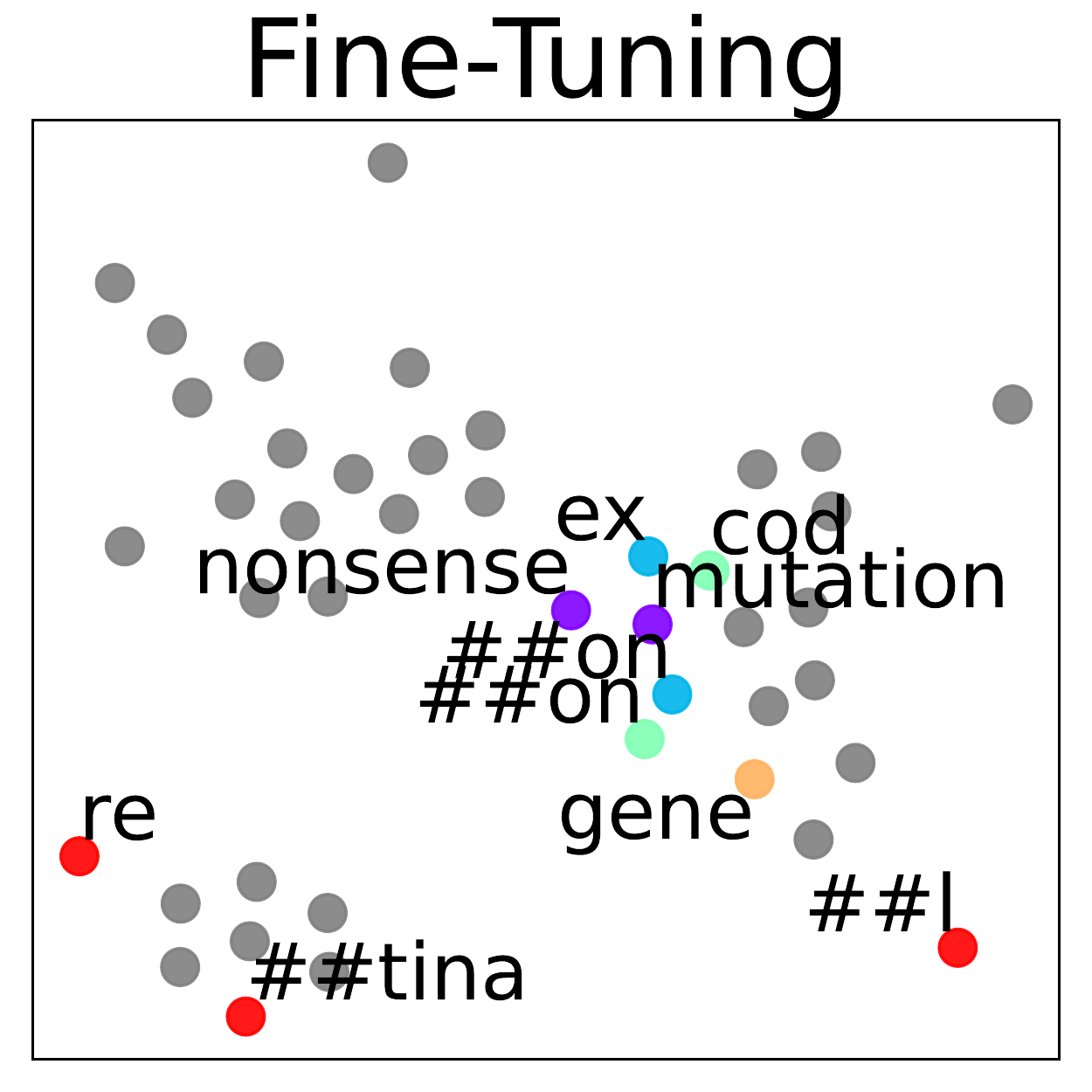}
    \end{minipage}
    \begin{minipage}{0.28\linewidth}
        \centering
        \includegraphics[width=1\linewidth]{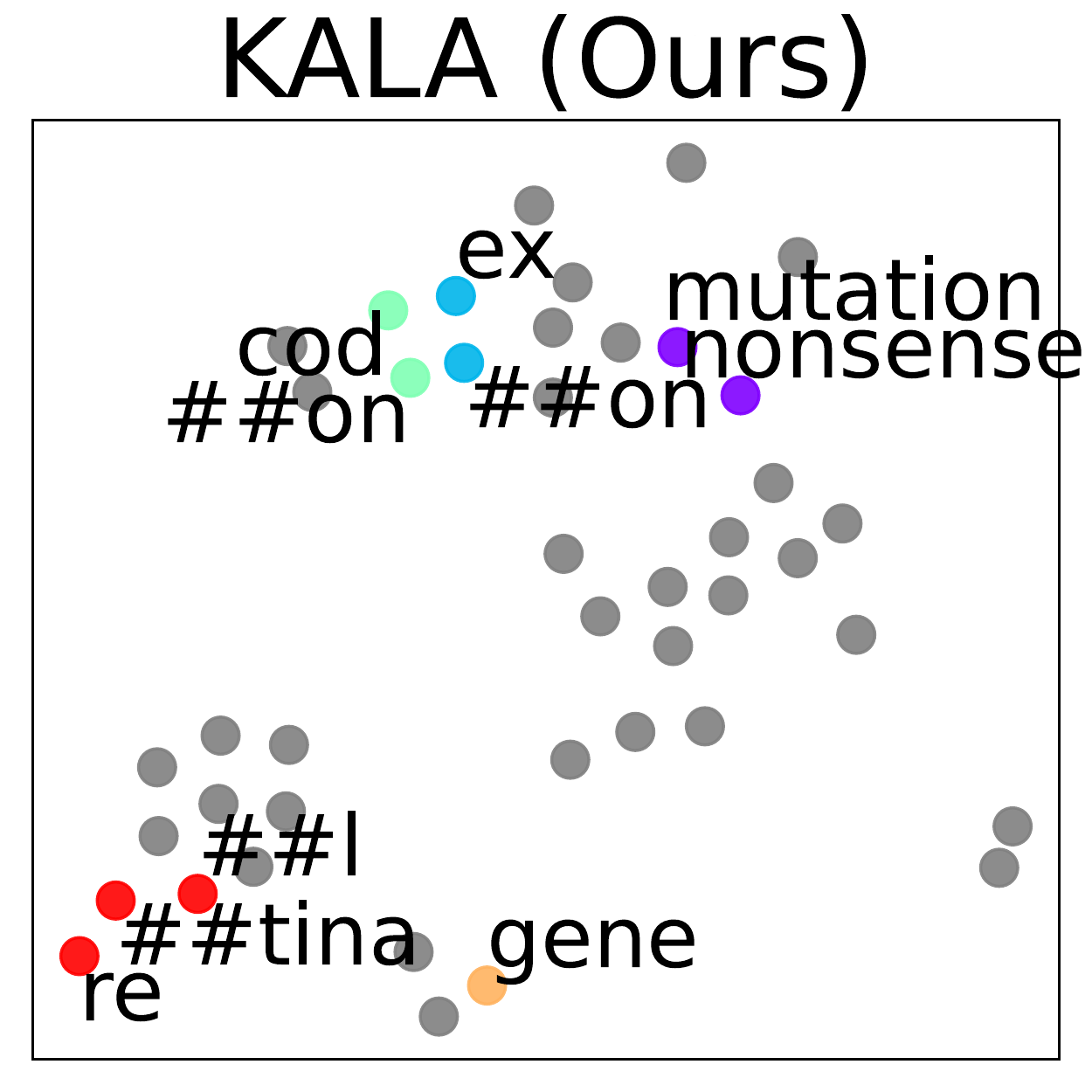}
    \end{minipage}
    \vspace{-0.05in}
    \caption{\small A case study on one context of the NCBI-Disease dataset. A left table shows the context and its fact, and a right figure shows a visualization of token representations. Text in blue and red denote the seen and unseen entities, respectively.}
    \label{fig:casestudy}
\end{minipage}
\hfill
\begin{minipage}{0.26\linewidth}
    \centering
% 	\small
    \resizebox{1.0\columnwidth}{!}{
% 	\resizebox{0.75\columnwidth}{!}{
	\renewcommand{\arraystretch}{1.05}
    \renewcommand{\tabcolsep}{3.0mm}
    \begin{tabular}{lcc}
    	\toprule
    	                                   & \multicolumn{2}{c}{\textbf{NewsQA}}  \\
    	\textbf{T5-small}                   & EM          & F1          \\ 
    	\midrule[0.8pt]
    	\textbf{Fine-tuning} & {48.96} &  {64.24}\\
    	\textbf{TAPT}        & {48.66} &  {64.30} \\
    	+ \textbf{RecAdam}   & {48.37} &  {63.41} \\
    	\midrule[0.8pt]
    	\textbf{KALA (ours)} & {\textbf{51.78}} & {\textbf{66.88}}\\ 
    	\bottomrule
    \end{tabular}
	}
	\vspace{-0.05in}
	\caption{\small Experimental results on generative question answering with T5-small as a PLM and NewsQA as a dataset.}
	\label{table:generative-qa}
	 \vspace{-0.15in}
\end{minipage}
% \vspace{-0.2in}
\end{figure*}

\subsection{Case Study}
\label{sec:case_study}
To better see how our KFM (\S\ref{sec:3.2}) works, we show the context and its fact, and then visualize representations from the PLM modulated by the KFM.
As shown in Figure~\ref{fig:casestudy} right, the token `\#\#on' is not aligned with their corresponding tokens, such as `ex' (for \textit{exon}) and `cod' (for \textit{codon}), in the baseline. However, with our feature modulation that transforms multiple tokens associated with the single entity equally, the two tokens (e.g., (`ex', `\#\#on')), composing one entity, are closely embedded. Also, while the baseline cannot handle the unseen entity consisting of three tokens: `re', `\#\#tina', and `\#\#l', KALA embeds them closely by representing the unseen \textit{retinal} from the representation of its neighborhood \textit{gene} derived by the domain knowledge -- (\textit{retinal}, \textit{instance of}, \textit{gene}).

\subsection{Extension to Generative Model}
\label{sec:generative}
Our KALA framework is also applicable to encoder-decoder PLMs by applying the KFM to the encoder. Therefore, we further validate KALA's effectiveness on the encoder-decoder PLMs on the generative QA task~\citep{SWEP} with T5-small~\citep{T5}. Table~\ref{table:generative-qa} shows that KALA largely outperforms baselines even with such a generative PLM.

\section{Conclusion}
\vspace{-0.075in}
In this paper, we introduced KALA, a novel framework for language model adaptation, which modulates the intermediate representations of a PLM by conditioning it with the entity memory and the relational facts from KGs.
We validated KALA on various domains of QA and NER tasks, on which KALA significantly outperforms relevant baselines while being computationally efficient.
We demonstrate that the success of KALA comes from both KFM and relational retrieval, allowing the PLM to recognize entities but also handle unseen ones that might frequently appear in domain-specific tasks.
There are many other avenues for future work, including the application of KALA on pre-training of knowledge-augmented PLMs from scratch.

\section*{Ethical Statements}
\vspace{-0.075in}
Enhancing the domain converge of pre-traind language models (PLMs) with external knowledge is increasingly important, since the PLMs cannot observe all the data during training and cannot memorize all the necessary knowledge for solving down-stream tasks. Our KALA contributes to this problem by augmenting domain knowledge graphs for PLMs. However, we have to still consider the accurateness of knowledge, i.e., the fact in the knowledge graph may not be correct, which affects the model to generate incorrect answers. Also, the model's prediction performance is still far from optimal. Thus, we should be aware of model's failure from errors in knowledge and prediction, especially on high-risk domains (e.g., biomedicine). 

\section*{Acknowledgement}
\vspace{-0.075in}
This work was supported by 
Institute of Information \& communications Technology Planning \& Evaluation (IITP) grant funded by the Korea government (MSIT) (No.2019-0-00075, Artificial Intelligence Graduate School Program (KAIST) and No. 2021-0-02068, Artificial Intelligence Innovation Hub)),
AITRICS,
Samsung Electronics (IO201214-08145-01),
and the Engineering Research Center Program through the National Research Foundation of Korea (NRF) funded by the Korean Government MSIT (NRF-2018R1A5A1059921).
\bibliography{anthology,custom}

\begin{thebibliography}{52}
\expandafter\ifx\csname natexlab\endcsname\relax\def\natexlab#1{#1}\fi

\bibitem[{Ba et~al.(2016)Ba, Kiros, and Hinton}]{LayerNorm}
Lei~Jimmy Ba, Jamie~Ryan Kiros, and Geoffrey~E. Hinton. 2016.
\newblock \href {http://arxiv.org/abs/1607.06450} {Layer normalization}.
\newblock \emph{arXiv preprint}, arXiv:1607.06450.

\bibitem[{Bai et~al.(2021)Bai, Ritter, and Xu}]{ProBERT}
Fan Bai, Alan Ritter, and Wei Xu. 2021.
\newblock \href {https://aclanthology.org/2021.emnlp-main.409} {Pre-train or
  annotate? domain adaptation with a constrained budget}.
\newblock In \emph{Proceedings of the 2021 Conference on Empirical Methods in
  Natural Language Processing, {EMNLP} 2021, Virtual Event / Punta Cana,
  Dominican Republic, 7-11 November, 2021}, pages 5002--5015.

\bibitem[{Beltagy et~al.(2019)Beltagy, Lo, and Cohan}]{SciBERT}
Iz~Beltagy, Kyle Lo, and Arman Cohan. 2019.
\newblock \href {https://doi.org/10.18653/v1/D19-1371} {Scibert: {A} pretrained
  language model for scientific text}.
\newblock In \emph{Proceedings of the 2019 Conference on Empirical Methods in
  Natural Language Processing and the 9th International Joint Conference on
  Natural Language Processing, {EMNLP-IJCNLP} 2019, Hong Kong, China, November
  3-7, 2019}, pages 3613--3618.

\bibitem[{Brody et~al.(2021)Brody, Alon, and Yahav}]{GATv2}
Shaked Brody, Uri Alon, and Eran Yahav. 2021.
\newblock \href {http://arxiv.org/abs/2105.14491} {How attentive are graph
  attention networks?}
\newblock \emph{arXiv preprint}, arXiv:2105.14491.

\bibitem[{Brown et~al.(2020)Brown, Mann, Ryder, Subbiah, Kaplan, Dhariwal,
  Neelakantan, Shyam, Sastry, Askell, Agarwal, Herbert{-}Voss, Krueger,
  Henighan, Child, Ramesh, Ziegler, Wu, Winter, Hesse, Chen, Sigler, Litwin,
  Gray, Chess, Clark, Berner, McCandlish, Radford, Sutskever, and
  Amodei}]{GPT3}
Tom~B. Brown, Benjamin Mann, Nick Ryder, Melanie Subbiah, Jared Kaplan,
  Prafulla Dhariwal, Arvind Neelakantan, Pranav Shyam, Girish Sastry, Amanda
  Askell, Sandhini Agarwal, Ariel Herbert{-}Voss, Gretchen Krueger, Tom
  Henighan, Rewon Child, Aditya Ramesh, Daniel~M. Ziegler, Jeffrey Wu, Clemens
  Winter, Christopher Hesse, Mark Chen, Eric Sigler, Mateusz Litwin, Scott
  Gray, Benjamin Chess, Jack Clark, Christopher Berner, Sam McCandlish, Alec
  Radford, Ilya Sutskever, and Dario Amodei. 2020.
\newblock \href
  {https://proceedings.neurips.cc/paper/2020/hash/1457c0d6bfcb4967418bfb8ac142f64a-Abstract.html}
  {Language models are few-shot learners}.
\newblock In \emph{Advances in Neural Information Processing Systems 33: Annual
  Conference on Neural Information Processing Systems 2020, NeurIPS 2020,
  December 6-12, 2020, virtual}.

\bibitem[{Carlson et~al.(2010)Carlson, Betteridge, Kisiel, Settles, Jr., and
  Mitchell}]{NELL}
Andrew Carlson, Justin Betteridge, Bryan Kisiel, Burr Settles, Estevam
  R.~Hruschka Jr., and Tom~M. Mitchell. 2010.
\newblock \href {http://www.aaai.org/ocs/index.php/AAAI/AAAI10/paper/view/1879}
  {Toward an architecture for never-ending language learning}.
\newblock In \emph{Proceedings of the Twenty-Fourth {AAAI} Conference on
  Artificial Intelligence, {AAAI} 2010, Atlanta, Georgia, USA, July 11-15,
  2010}.

\bibitem[{Chen et~al.(2020)Chen, Hou, Cui, Che, Liu, and Yu}]{RecAdam}
Sanyuan Chen, Yutai Hou, Yiming Cui, Wanxiang Che, Ting Liu, and Xiangzhan Yu.
  2020.
\newblock \href {https://doi.org/10.18653/v1/2020.emnlp-main.634} {Recall and
  learn: Fine-tuning deep pretrained language models with less forgetting}.
\newblock In \emph{Proceedings of the 2020 Conference on Empirical Methods in
  Natural Language Processing, {EMNLP} 2020, Online, November 16-20, 2020},
  pages 7870--7881.

\bibitem[{Clark et~al.(2020)Clark, Luong, Le, and Manning}]{ELECTRA}
Kevin Clark, Minh{-}Thang Luong, Quoc~V. Le, and Christopher~D. Manning. 2020.
\newblock \href {https://openreview.net/forum?id=r1xMH1BtvB} {{ELECTRA:}
  pre-training text encoders as discriminators rather than generators}.
\newblock In \emph{8th International Conference on Learning Representations,
  {ICLR} 2020, Addis Ababa, Ethiopia, April 26-30, 2020}.

\bibitem[{Derczynski et~al.(2017)Derczynski, Nichols, van Erp, and
  Limsopatham}]{WNUT}
Leon Derczynski, Eric Nichols, Marieke van Erp, and Nut Limsopatham. 2017.
\newblock \href {https://doi.org/10.18653/v1/w17-4418} {Results of the
  {WNUT2017} shared task on novel and emerging entity recognition}.
\newblock In \emph{Proceedings of the 3rd Workshop on Noisy User-generated
  Text, NUT@EMNLP 2017, Copenhagen, Denmark, September 7, 2017}, pages
  140--147. Association for Computational Linguistics.

\bibitem[{Devlin et~al.(2019)Devlin, Chang, Lee, and Toutanova}]{BERT}
Jacob Devlin, Ming{-}Wei Chang, Kenton Lee, and Kristina Toutanova. 2019.
\newblock \href {https://doi.org/10.18653/v1/n19-1423} {{BERT:} pre-training of
  deep bidirectional transformers for language understanding}.
\newblock In \emph{Proceedings of the 2019 Conference of the North American
  Chapter of the Association for Computational Linguistics: Human Language
  Technologies, {NAACL-HLT} 2019, Minneapolis, MN, USA, June 2-7, 2019, Volume
  1 (Long and Short Papers)}, pages 4171--4186.

\bibitem[{Dogan et~al.(2014)Dogan, Leaman, and Lu}]{NCBI}
Rezarta~Islamaj Dogan, Robert Leaman, and Zhiyong Lu. 2014.
\newblock \href {https://doi.org/10.1016/j.jbi.2013.12.006} {{NCBI} disease
  corpus: {A} resource for disease name recognition and concept normalization}.
\newblock \emph{J. Biomed. Informatics}, 47:1--10.

\bibitem[{Dumoulin et~al.(2018)Dumoulin, Perez, Schucher, Strub, Vries,
  Courville, and Bengio}]{Featurewise}
Vincent Dumoulin, Ethan Perez, Nathan Schucher, Florian Strub, Harm~de Vries,
  Aaron Courville, and Yoshua Bengio. 2018.
\newblock \href {https://distill.pub/2018/feature-wise-transformations}
  {Feature-wise transformations}.
\newblock \emph{Distill}.

\bibitem[{F{\'{e}}vry et~al.(2020)F{\'{e}}vry, Soares, FitzGerald, Choi, and
  Kwiatkowski}]{EaE}
Thibault F{\'{e}}vry, Livio~Baldini Soares, Nicholas FitzGerald, Eunsol Choi,
  and Tom Kwiatkowski. 2020.
\newblock \href {https://doi.org/10.18653/v1/2020.emnlp-main.400} {Entities as
  experts: Sparse memory access with entity supervision}.
\newblock In \emph{Proceedings of the 2020 Conference on Empirical Methods in
  Natural Language Processing, {EMNLP} 2020, Online, November 16-20, 2020},
  pages 4937--4951.

\bibitem[{Fey and Lenssen(2019)}]{torch-geometric}
Matthias Fey and Jan~E. Lenssen. 2019.
\newblock \href {http://arxiv.org/abs/1903.02428} {Fast graph representation
  learning with {PyTorch Geometric}}.
\newblock In \emph{ICLR Workshop on Representation Learning on Graphs and
  Manifolds}.

\bibitem[{Gururangan et~al.(2021)Gururangan, Lewis, Holtzman, Smith, and
  Zettlemoyer}]{demix}
Suchin Gururangan, Mike Lewis, Ari Holtzman, Noah~A. Smith, and Luke
  Zettlemoyer. 2021.
\newblock \href {http://arxiv.org/abs/2108.05036} {Demix layers: Disentangling
  domains for modular language modeling}.
\newblock \emph{arXiv preprint}, arXiv:2108.05036.

\bibitem[{Gururangan et~al.(2020)Gururangan, Marasovic, Swayamdipta, Lo,
  Beltagy, Downey, and Smith}]{dontstoppt}
Suchin Gururangan, Ana Marasovic, Swabha Swayamdipta, Kyle Lo, Iz~Beltagy, Doug
  Downey, and Noah~A. Smith. 2020.
\newblock \href {https://doi.org/10.18653/v1/2020.acl-main.740} {Don't stop
  pretraining: Adapt language models to domains and tasks}.
\newblock In \emph{Proceedings of the 58th Annual Meeting of the Association
  for Computational Linguistics, {ACL} 2020, Online, July 5-10, 2020}, pages
  8342--8360.

\bibitem[{Hamilton et~al.(2017)Hamilton, Ying, and Leskovec}]{GraphSAGE}
William~L. Hamilton, Zhitao Ying, and Jure Leskovec. 2017.
\newblock \href
  {https://proceedings.neurips.cc/paper/2017/hash/5dd9db5e033da9c6fb5ba83c7a7ebea9-Abstract.html}
  {Inductive representation learning on large graphs}.
\newblock In \emph{Advances in Neural Information Processing Systems 30: Annual
  Conference on Neural Information Processing Systems 2017, December 4-9, 2017,
  Long Beach, CA, {USA}}, pages 1024--1034.

\bibitem[{Han and Eisenstein(2019)}]{DAPT}
Xiaochuang Han and Jacob Eisenstein. 2019.
\newblock \href {https://doi.org/10.18653/v1/D19-1433} {Unsupervised domain
  adaptation of contextualized embeddings for sequence labeling}.
\newblock In \emph{Proceedings of the 2019 Conference on Empirical Methods in
  Natural Language Processing and the 9th International Joint Conference on
  Natural Language Processing, {EMNLP-IJCNLP} 2019, Hong Kong, China, November
  3-7, 2019}, pages 4237--4247.

\bibitem[{Hendrycks and Gimpel(2016)}]{GELU}
Dan Hendrycks and Kevin Gimpel. 2016.
\newblock \href {http://arxiv.org/abs/1606.08415} {Bridging nonlinearities and
  stochastic regularizers with gaussian error linear units}.
\newblock \emph{arXiv preprint}, arXiv:1606.08415.

\bibitem[{Houlsby et~al.(2019)Houlsby, Giurgiu, Jastrzebski, Morrone,
  de~Laroussilhe, Gesmundo, Attariyan, and Gelly}]{HoulsbyAdapter}
Neil Houlsby, Andrei Giurgiu, Stanislaw Jastrzebski, Bruna Morrone, Quentin
  de~Laroussilhe, Andrea Gesmundo, Mona Attariyan, and Sylvain Gelly. 2019.
\newblock \href {http://proceedings.mlr.press/v97/houlsby19a.html}
  {Parameter-efficient transfer learning for {NLP}}.
\newblock In \emph{Proceedings of the 36th International Conference on Machine
  Learning, {ICML} 2019, 9-15 June 2019, Long Beach, California, {USA}}, pages
  2790--2799.

\bibitem[{Howard and Ruder(2018)}]{TAPT}
Jeremy Howard and Sebastian Ruder. 2018.
\newblock \href {https://aclanthology.org/P18-1031/} {Universal language model
  fine-tuning for text classification}.
\newblock In \emph{Proceedings of the 56th Annual Meeting of the Association
  for Computational Linguistics, {ACL} 2018, Melbourne, Australia, July 15-20,
  2018, Volume 1: Long Papers}, pages 328--339.

\bibitem[{Huang and Belongie(2017)}]{adain}
Xun Huang and Serge~J. Belongie. 2017.
\newblock \href {https://doi.org/10.1109/ICCV.2017.167} {Arbitrary style
  transfer in real-time with adaptive instance normalization}.
\newblock In \emph{{IEEE} International Conference on Computer Vision, {ICCV}
  2017, Venice, Italy, October 22-29, 2017}, pages 1510--1519. {IEEE} Computer
  Society.

\bibitem[{Jin et~al.(2020)Jin, Pan, Oufattole, Weng, Fang, and
  Szolovits}]{MedExam}
Di~Jin, Eileen Pan, Nassim Oufattole, Wei{-}Hung Weng, Hanyi Fang, and Peter
  Szolovits. 2020.
\newblock \href {http://arxiv.org/abs/2009.13081} {What disease does this
  patient have? {A} large-scale open domain question answering dataset from
  medical exams}.
\newblock \emph{arXiv preprint}, arXiv:2009.13081.

\bibitem[{Lee et~al.(2020)Lee, Yoon, Kim, Kim, Kim, So, and Kang}]{BioBERT}
Jinhyuk Lee, Wonjin Yoon, Sungdong Kim, Donghyeon Kim, Sunkyu Kim, Chan~Ho So,
  and Jaewoo Kang. 2020.
\newblock \href {https://doi.org/10.1093/bioinformatics/btz682} {Biobert: a
  pre-trained biomedical language representation model for biomedical text
  mining}.
\newblock \emph{Bioinform.}, 36(4):1234--1240.

\bibitem[{Lee et~al.(2021)Lee, Kang, Lee, and Hwang}]{SWEP}
Seanie Lee, Minki Kang, Juho Lee, and Sung~Ju Hwang. 2021.
\newblock \href {https://doi.org/10.18653/v1/2021.acl-long.434} {Learning to
  perturb word embeddings for out-of-distribution {QA}}.
\newblock In \emph{Proceedings of the 59th Annual Meeting of the Association
  for Computational Linguistics and the 11th International Joint Conference on
  Natural Language Processing, {ACL/IJCNLP} 2021, (Volume 1: Long Papers),
  Virtual Event, August 1-6, 2021}, pages 5583--5595.

\bibitem[{Liu et~al.(2020)Liu, Ott, Goyal, Du, Joshi, Chen, Levy, Lewis,
  Zettlemoyer, and Stoyanov}]{RoBERTa}
Yinhan Liu, Myle Ott, Naman Goyal, Jingfei Du, Mandar Joshi, Danqi Chen, Omer
  Levy, Mike Lewis, Luke Zettlemoyer, and Veselin Stoyanov. 2020.
\newblock Roberta: {A} robustly optimized {BERT} pretraining approach.
\newblock \emph{arXiv preprint}, arXiv:1907.11692.

\bibitem[{Loshchilov and Hutter(2019)}]{AdamW}
Ilya Loshchilov and Frank Hutter. 2019.
\newblock \href {https://openreview.net/forum?id=Bkg6RiCqY7} {Decoupled weight
  decay regularization}.
\newblock In \emph{7th International Conference on Learning Representations,
  {ICLR} 2019, New Orleans, LA, USA, May 6-9, 2019}.

\bibitem[{Micikevicius et~al.(2018)Micikevicius, Narang, Alben, Diamos, Elsen,
  Garc{\'{\i}}a, Ginsburg, Houston, Kuchaiev, Venkatesh, and Wu}]{fp16}
Paulius Micikevicius, Sharan Narang, Jonah Alben, Gregory~F. Diamos, Erich
  Elsen, David Garc{\'{\i}}a, Boris Ginsburg, Michael Houston, Oleksii
  Kuchaiev, Ganesh Venkatesh, and Hao Wu. 2018.
\newblock \href {https://openreview.net/forum?id=r1gs9JgRZ} {Mixed precision
  training}.
\newblock In \emph{{ICLR} 2018, Vancouver, BC, Canada, April 30 - May 3, 2018,
  Conference Track Proceedings}.

\bibitem[{Miller(1995)}]{WordNet}
George~A. Miller. 1995.
\newblock \href {http://doi.acm.org/10.1145/219717.219748} {Wordnet: {A}
  lexical database for english}.
\newblock \emph{Commun. {ACM}}, 38(11):39--41.

\bibitem[{Nair and Hinton(2010)}]{ReLU}
Vinod Nair and Geoffrey~E. Hinton. 2010.
\newblock \href {https://icml.cc/Conferences/2010/papers/432.pdf} {Rectified
  linear units improve restricted boltzmann machines}.
\newblock In \emph{Proceedings of the 27th International Conference on Machine
  Learning (ICML-10), June 21-24, 2010, Haifa, Israel}, pages 807--814.

\bibitem[{Pampari et~al.(2018)Pampari, Raghavan, Liang, and Peng}]{emrQA}
Anusri Pampari, Preethi Raghavan, Jennifer~J. Liang, and Jian Peng. 2018.
\newblock \href {https://doi.org/10.18653/v1/d18-1258} {emrqa: {A} large corpus
  for question answering on electronic medical records}.
\newblock In \emph{Proceedings of the 2018 Conference on Empirical Methods in
  Natural Language Processing, Brussels, Belgium, October 31 - November 4,
  2018}, pages 2357--2368.

\bibitem[{Paszke et~al.(2019)Paszke, Gross, Massa, Lerer, Bradbury, Chanan,
  Killeen, Lin, Gimelshein, Antiga, Desmaison, Kopf, Yang, DeVito, Raison,
  Tejani, Chilamkurthy, Steiner, Fang, Bai, and Chintala}]{Pytorch}
Adam Paszke, Sam Gross, Francisco Massa, Adam Lerer, James Bradbury, Gregory
  Chanan, Trevor Killeen, Zeming Lin, Natalia Gimelshein, Luca Antiga, Alban
  Desmaison, Andreas Kopf, Edward Yang, Zachary DeVito, Martin Raison, Alykhan
  Tejani, Sasank Chilamkurthy, Benoit Steiner, Lu~Fang, Junjie Bai, and Soumith
  Chintala. 2019.
\newblock \href
  {https://proceedings.neurips.cc/paper/2019/hash/bdbca288fee7f92f2bfa9f7012727740-Abstract.html}
  {Pytorch: An imperative style, high-performance deep learning library}.
\newblock In \emph{Advances in Neural Information Processing Systems 32}, pages
  8024--8035.

\bibitem[{Perez et~al.(2018)Perez, Strub, de~Vries, Dumoulin, and
  Courville}]{FiLM}
Ethan Perez, Florian Strub, Harm de~Vries, Vincent Dumoulin, and Aaron~C.
  Courville. 2018.
\newblock \href
  {https://www.aaai.org/ocs/index.php/AAAI/AAAI18/paper/view/16528} {Film:
  Visual reasoning with a general conditioning layer}.
\newblock In \emph{Proceedings of the Thirty-Second {AAAI} Conference on
  Artificial Intelligence, (AAAI-18), the 30th innovative Applications of
  Artificial Intelligence (IAAI-18), and the 8th {AAAI} Symposium on
  Educational Advances in Artificial Intelligence (EAAI-18), New Orleans,
  Louisiana, USA, February 2-7, 2018}, pages 3942--3951.

\bibitem[{Peters et~al.(2019)Peters, Neumann, IV, Schwartz, Joshi, Singh, and
  Smith}]{KnowBERT}
Matthew~E. Peters, Mark Neumann, Robert L.~Logan IV, Roy Schwartz, Vidur Joshi,
  Sameer Singh, and Noah~A. Smith. 2019.
\newblock \href {https://doi.org/10.18653/v1/D19-1005} {Knowledge enhanced
  contextual word representations}.
\newblock In \emph{Proceedings of the 2019 Conference on Empirical Methods in
  Natural Language Processing and the 9th International Joint Conference on
  Natural Language Processing, {EMNLP-IJCNLP} 2019, Hong Kong, China, November
  3-7, 2019}, pages 43--54.

\bibitem[{Qin et~al.(2021)Qin, Lin, Takanobu, Liu, Li, Ji, Huang, Sun, and
  Zhou}]{ERICA}
Yujia Qin, Yankai Lin, Ryuichi Takanobu, Zhiyuan Liu, Peng Li, Heng Ji, Minlie
  Huang, Maosong Sun, and Jie Zhou. 2021.
\newblock \href {https://doi.org/10.18653/v1/2021.acl-long.260} {{ERICA:}
  improving entity and relation understanding for pre-trained language models
  via contrastive learning}.
\newblock In \emph{Proceedings of the 59th Annual Meeting of the Association
  for Computational Linguistics and the 11th International Joint Conference on
  Natural Language Processing, {ACL/IJCNLP} 2021, (Volume 1: Long Papers),
  Virtual Event, August 1-6, 2021}, pages 3350--3363.

\bibitem[{Raffel et~al.(2020)Raffel, Shazeer, Roberts, Lee, Narang, Matena,
  Zhou, Li, and Liu}]{T5}
Colin Raffel, Noam Shazeer, Adam Roberts, Katherine Lee, Sharan Narang, Michael
  Matena, Yanqi Zhou, Wei Li, and Peter~J. Liu. 2020.
\newblock \href {http://jmlr.org/papers/v21/20-074.html} {Exploring the limits
  of transfer learning with a unified text-to-text transformer}.
\newblock \emph{J. Mach. Learn. Res.}, 21:140:1--140:67.

\bibitem[{Ruder(2019)}]{RuderThesis}
Sebastian Ruder. 2019.
\newblock \href {https://ruder.io/thesis/neural_transfer_learning_for_nlp.pdf}
  {\emph{Neural Transfer Learning for Natural Language Processing}}.
\newblock Ph.D. thesis, National University of Ireland, Galway.

\bibitem[{Sang and Meulder(2003)}]{CoNLL}
Erik F. Tjong~Kim Sang and Fien~De Meulder. 2003.
\newblock \href {https://aclanthology.org/W03-0419/} {Introduction to the
  conll-2003 shared task: Language-independent named entity recognition}.
\newblock In \emph{Proceedings of the Seventh Conference on Natural Language
  Learning, CoNLL 2003, Held in cooperation with {HLT-NAACL} 2003, Edmonton,
  Canada, May 31 - June 1, 2003}, pages 142--147. {ACL}.

\bibitem[{Shazeer and Stern(2018)}]{Adafactor}
Noam Shazeer and Mitchell Stern. 2018.
\newblock \href {http://proceedings.mlr.press/v80/shazeer18a.html} {Adafactor:
  Adaptive learning rates with sublinear memory cost}.
\newblock In \emph{Proceedings of the 35th International Conference on Machine
  Learning, {ICML} 2018, Stockholmsm{\"{a}}ssan, Stockholm, Sweden, July 10-15,
  2018}, pages 4603--4611.

\bibitem[{Sukhbaatar et~al.(2015)Sukhbaatar, Szlam, Weston, and Fergus}]{MemNN}
Sainbayar Sukhbaatar, Arthur Szlam, Jason Weston, and Rob Fergus. 2015.
\newblock \href
  {https://proceedings.neurips.cc/paper/2015/hash/8fb21ee7a2207526da55a679f0332de2-Abstract.html}
  {End-to-end memory networks}.
\newblock In \emph{Advances in Neural Information Processing Systems 28: Annual
  Conference on Neural Information Processing Systems 2015, December 7-12,
  2015, Montreal, Quebec, Canada}, pages 2440--2448.

\bibitem[{Trischler et~al.(2017)Trischler, Wang, Yuan, Harris, Sordoni,
  Bachman, and Suleman}]{NewsQA}
Adam Trischler, Tong Wang, Xingdi Yuan, Justin Harris, Alessandro Sordoni,
  Philip Bachman, and Kaheer Suleman. 2017.
\newblock \href {https://doi.org/10.18653/v1/w17-2623} {Newsqa: {A} machine
  comprehension dataset}.
\newblock In \emph{Proceedings of the 2nd Workshop on Representation Learning
  for NLP, Rep4NLP@ACL 2017, Vancouver, Canada, August 3, 2017}, pages
  191--200.

\bibitem[{Vaswani et~al.(2017)Vaswani, Shazeer, Parmar, Uszkoreit, Jones,
  Gomez, Kaiser, and Polosukhin}]{Transformer}
Ashish Vaswani, Noam Shazeer, Niki Parmar, Jakob Uszkoreit, Llion Jones,
  Aidan~N. Gomez, Lukasz Kaiser, and Illia Polosukhin. 2017.
\newblock \href
  {https://proceedings.neurips.cc/paper/2017/hash/3f5ee243547dee91fbd053c1c4a845aa-Abstract.html}
  {Attention is all you need}.
\newblock In \emph{Advances in Neural Information Processing Systems 30: Annual
  Conference on Neural Information Processing Systems 2017, December 4-9, 2017,
  Long Beach, CA, {USA}}, pages 5998--6008.

\bibitem[{Velickovic et~al.(2018)Velickovic, Cucurull, Casanova, Romero,
  Li{\`{o}}, and Bengio}]{GAT}
Petar Velickovic, Guillem Cucurull, Arantxa Casanova, Adriana Romero, Pietro
  Li{\`{o}}, and Yoshua Bengio. 2018.
\newblock \href {https://openreview.net/forum?id=rJXMpikCZ} {Graph attention
  networks}.
\newblock In \emph{6th International Conference on Learning Representations,
  {ICLR} 2018, Vancouver, BC, Canada, April 30 - May 3, 2018, Conference Track
  Proceedings}.

\bibitem[{Verga et~al.(2021)Verga, Sun, Soares, and Cohen}]{FaE}
Pat Verga, Haitian Sun, Livio~Baldini Soares, and William~W. Cohen. 2021.
\newblock \href {https://doi.org/10.18653/v1/2021.naacl-main.288} {Adaptable
  and interpretable neural memoryover symbolic knowledge}.
\newblock In \emph{Proceedings of the 2021 Conference of the North American
  Chapter of the Association for Computational Linguistics: Human Language
  Technologies, {NAACL-HLT} 2021, Online, June 6-11, 2021}, pages 3678--3691.

\bibitem[{Vrandecic and Kr{\"{o}}tzsch(2014)}]{wikidata}
Denny Vrandecic and Markus Kr{\"{o}}tzsch. 2014.
\newblock \href {https://doi.org/10.1145/2629489} {Wikidata: a free
  collaborative knowledgebase}.
\newblock \emph{Commun. {ACM}}, 57(10):78--85.

\bibitem[{Wolf et~al.(2020)Wolf, Debut, Sanh, Chaumond, Delangue, Moi, Cistac,
  Rault, Louf, Funtowicz, Davison, Shleifer, von Platen, Ma, Jernite, Plu, Xu,
  Scao, Gugger, Drame, Lhoest, and Rush}]{transformers}
Thomas Wolf, Lysandre Debut, Victor Sanh, Julien Chaumond, Clement Delangue,
  Anthony Moi, Pierric Cistac, Tim Rault, R{\'{e}}mi Louf, Morgan Funtowicz,
  Joe Davison, Sam Shleifer, Patrick von Platen, Clara Ma, Yacine Jernite,
  Julien Plu, Canwen Xu, Teven~Le Scao, Sylvain Gugger, Mariama Drame, Quentin
  Lhoest, and Alexander~M. Rush. 2020.
\newblock \href {https://doi.org/10.18653/v1/2020.emnlp-demos.6} {Transformers:
  State-of-the-art natural language processing}.
\newblock In \emph{{EMNLP} 2020 - Demos, Online, November 16-20, 2020}, pages
  38--45.

\bibitem[{Yamada et~al.(2020)Yamada, Asai, Shindo, Takeda, and
  Matsumoto}]{LUKE}
Ikuya Yamada, Akari Asai, Hiroyuki Shindo, Hideaki Takeda, and Yuji Matsumoto.
  2020.
\newblock \href {https://doi.org/10.18653/v1/2020.emnlp-main.523} {{LUKE:} deep
  contextualized entity representations with entity-aware self-attention}.
\newblock In \emph{Proceedings of the 2020 Conference on Empirical Methods in
  Natural Language Processing, {EMNLP} 2020, Online, November 16-20, 2020},
  pages 6442--6454.

\bibitem[{Yang et~al.(2019)Yang, Wang, Liu, Liu, Lyu, Wu, She, and Li}]{KTNet}
An~Yang, Quan Wang, Jing Liu, Kai Liu, Yajuan Lyu, Hua Wu, Qiaoqiao She, and
  Sujian Li. 2019.
\newblock \href {https://doi.org/10.18653/v1/p19-1226} {Enhancing pre-trained
  language representations with rich knowledge for machine reading
  comprehension}.
\newblock In \emph{Proceedings of the 57th Conference of the Association for
  Computational Linguistics, {ACL} 2019, Florence, Italy, July 28- August 2,
  2019, Volume 1: Long Papers}, pages 2346--2357.

\bibitem[{Yao et~al.(2019)Yao, Ye, Li, Han, Lin, Liu, Liu, Huang, Zhou, and
  Sun}]{DocRED}
Yuan Yao, Deming Ye, Peng Li, Xu~Han, Yankai Lin, Zhenghao Liu, Zhiyuan Liu,
  Lixin Huang, Jie Zhou, and Maosong Sun. 2019.
\newblock \href {https://doi.org/10.18653/v1/p19-1074} {Docred: {A} large-scale
  document-level relation extraction dataset}.
\newblock In \emph{Proceedings of the 57th Conference of the Association for
  Computational Linguistics, {ACL} 2019, Florence, Italy, July 28- August 2,
  2019, Volume 1: Long Papers}, pages 764--777.

\bibitem[{Yosinski et~al.(2014)Yosinski, Clune, Bengio, and
  Lipson}]{transferlearning}
Jason Yosinski, Jeff Clune, Yoshua Bengio, and Hod Lipson. 2014.
\newblock \href
  {https://proceedings.neurips.cc/paper/2014/hash/375c71349b295fbe2dcdca9206f20a06-Abstract.html}
  {How transferable are features in deep neural networks?}
\newblock In \emph{Advances in Neural Information Processing Systems 27: Annual
  Conference on Neural Information Processing Systems 2014, December 8-13 2014,
  Montreal, Quebec, Canada}, pages 3320--3328.

\bibitem[{Yue et~al.(2020)Yue, Gutierrez, and Sun}]{CliniRC}
Xiang Yue, Bernal~Jimenez Gutierrez, and Huan Sun. 2020.
\newblock \href {https://doi.org/10.18653/v1/2020.acl-main.410} {Clinical
  reading comprehension: {A} thorough analysis of the emrqa dataset}.
\newblock In \emph{Proceedings of the 58th Annual Meeting of the Association
  for Computational Linguistics, {ACL} 2020, Online, July 5-10, 2020}, pages
  4474--4486.

\bibitem[{Zhang et~al.(2019)Zhang, Han, Liu, Jiang, Sun, and Liu}]{ERNIE}
Zhengyan Zhang, Xu~Han, Zhiyuan Liu, Xin Jiang, Maosong Sun, and Qun Liu. 2019.
\newblock \href {https://doi.org/10.18653/v1/p19-1139} {{ERNIE:} enhanced
  language representation with informative entities}.
\newblock In \emph{Proceedings of the 57th Conference of the Association for
  Computational Linguistics, {ACL} 2019, Florence, Italy, July 28- August 2,
  2019, Volume 1: Long Papers}, pages 1441--1451.

\end{thebibliography}
\bibliographystyle{acl_natbib}

\clearpage
\newpage
\appendix

\begin{figure*}[t]
    \centering
    \includegraphics[width=1.0\linewidth]{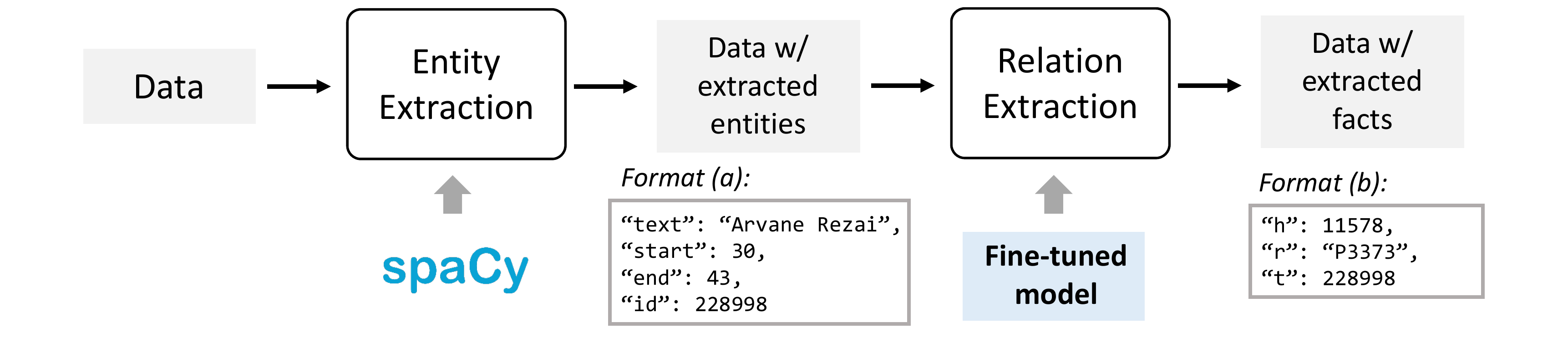}
    \vskip -0.225in
    \caption{\small Visual diagram of the KG construction pipeline used in this work. The entity format is composed of its corresponding text in the data, its character-level mention boundary, and its wikidata id. The fact format is composed of the head, relation, and tail, where head and tail entities are represented with their wikidata ids following the entity format.}
    \label{fig:kg_pipeline}
\end{figure*}
\begin{table*}[ht]
\centering
\resizebox{0.98\textwidth}{!}{
\begin{tabular}{lcccccc}
\toprule
\multicolumn{1}{l}{Hyperparameters} & NewsQA & Relation & Medication & CoNLL-2003 & WNUT-17 & NCBI-Disease \\ 
\midrule
LM for Relation Extraction         & \multicolumn{6}{c}{BERT-base-uncased} \\
Threshold on Relation Extraction   & \multicolumn{6}{c}{0.1}               \\
Size of Entity Memory              & 62823 &  5724  & 4635 & 10288  & 101 & 3502  \\
\midrule
The location of KFM                & 11    &  11    & 11   & 8  & 9, 11 & 8, 10  \\
\bottomrule
\end{tabular}
}
\vspace{-0.1in}
\caption{\small \textbf{Hyperparamters for Knowledge Graph (Top) and KALA (Bottom)} on six datasets we used. The reported performances on main paper are measured with the above settings.}
\label{table:hyper-kg}
\vspace{-0.15in}
\end{table*}

\section{Details on KG Construction}
\label{appendix:kg}
In this work, we propose to use the Knowledge Graph (KG) that can define the relational information among entities that only appear in each dataset. However, unfortunately, most of the task datasets do not contain such relational facts on its context, thus we need to construct them manually to obtain the knowledge graph. In this section, we explain the way of constructing the knowledge graph that we used, consisting of facts of entities for each context in the task dataset. 

Relation extraction is the way how we obtain the factual knowledge from the text of the target dataset. To do so, we first need to extract entities and their corresponding mentions from the text, and then link it to the existing entities in wikidata~\cite{wikidata}. In order to do this, we use the existing library named as spaCy\footnote{https://spacy.io/}, and open-sourced implementation of Entity Linker\footnote{https://github.com/egerber/spaCy-entity-linker}. To sum up, in our work, a set of entities $\mathcal{E}^{(i)}$ and corresponding mentions $\mathcal{M}^{(i)}$ for the given input $\vx^{(i)}$ are obtained through this step.
Regarding a concrete example, please see format (a) in Figure~\ref{fig:kg_pipeline}. In the example, ``Text'' indicates the entity mention within the input $\vx$, the ``start'' and ``end'' indicates its mention position denoted as $(m^\alpha, m^\omega)$, and ``id'' indicates the wikidata id for the entity identification used in the next step.

To extract the relation among entities that we obtained above, we use the scheme of Relation Extraction (RE). In other words, we use the trained RE model to build our own knowledge base (KB) instead of using the existing KG directly from the existing general-domain KB\footnote{We faced several problems here. First of all, most KBs such as Wikidata are less informative, especially for the entities included in the domain-specific context (e.g., News, Medical records). It only has a few facts for each context of domain-specific tasks, although we can find a lot of entities included in the context. Second, the entity linker is imperfect. Due to the wrongly linked entity to the wikidata, even existing relations in the KG are ignored a lot. Therefore, we instead use a trained neural network to effectively extract the relations between entities, instead of direct querying to obtain facts.}. Specifically, we first fine-tune the BERT-base model~\citep{BERT} for 2 epochs with 600k distantly supervised data used in~\citet{ERICA}, where the Wikipedia document and the Wikidata triplets are aligned. 
Then, we use the fine-tuned BERT model to extract the relations between entity pairs in the text.
We use the model with a simple bilinear layer on top of it, which is widely used scheme in the relation extraction literature~\citep{DocRED}.
For an example of the extracted fact, please see format (b) in Figure~\ref{fig:kg_pipeline}. In the example, ``h'' denotes the wikidata id of the head entity, ``r'' denotes the wikidata id of the extracted relation, and ``t'' denotes the wikidata id of the tail entity.
In the relation extraction, the model returns the categorical distribution over the top 100 frequent relations. In general, the relation of top-1 probability is used as the relation for the corresponding entity pair. 
However, this approach sometimes results in predicting \texttt{no\_relation} on most entity pairs. Thus, to obtain more relations, we further use the relation of top-2 probability in the case where \texttt{no\_relation} has a top-1 probability but the top-2 probability is larger than a certain threshold (e.g., $> 0.1$).
In Figure~\ref{fig:kg_pipeline}, we summarize our KG construction pipeline. In Table~\ref{table:hyper-kg}, we report the hyperparameters related to our KG construction.

\section{Experimental Setup}
\label{appendix:setup}
In this section, we introduce the detailed setups for our models and baselines used in Table~\ref{qa-exp},~\ref{ner-exp}, and~\ref{table:variants}.

\begin{table*}[!htb]
\centering
\resizebox{0.98\textwidth}{!}{
    \begin{tabular}{lccccccccc}
    \toprule            
                        &  \multicolumn{3}{c}{\textbf{Training}}    &  \multicolumn{3}{c}{\textbf{Validation}}    & \multicolumn{3}{c}{\textbf{Test}} \\ 
    Dataset             &  \# Context & C. Length &  \# Question &  \# Context & C. Length &  \# Question &  
                        \# Context & C. Length &  \# Question \\
    \midrule
    \textbf{NewsQA}     & 11428 & 655.7 & 74160  &  -   &   -    &  -   & 106  & 625.8  & 674  \\
    \textbf{Relation}   & 296   & 1386.1 & 6162   & 42   & 1206.6 & 321  & 85   & 1467.7  & 802  \\
    \textbf{Medication} & 182   & 1737.3 & 7518   & 26   & 1626.5 & 1858 & 53   & 2005.0  & 4005 \\
    \bottomrule
    \end{tabular}
}
\vspace{-0.1in}
\caption{\small \textbf{QA dataset statistics.} We report the number of contexts and questions (i.e., \# Context and \# Question), with the average length of contexts (i.e., C. Length) where the length is measured as the number of tokens after wordpiece tokenization.}
\label{table:qastat}
\vspace{-0.1in}
\end{table*}

\subsection{Implementation Details}
\label{appendix:implement}
We use the Pytorch~\citep{Pytorch} for the implementation of all models. Also, to easily implement the language model, we use the huggingface library~\citep{transformers} containing various transformer-based pre-trained language models (PLMs) and their checkpoints.

\paragraph{Details for KALA}
In this paragraph, we describe the implementation details of the components, such as four linear layers in the proposed KFM, architectural specifications in the attention-based GNN, and initialization of both the entity memory and relational embeddings, in the following. 
In terms of the functions $h_1, h_2, h_3,$ and $h_4$ in the KFM of Equation~\ref{eqn:entembed}, we use two linear layers with the ReLU~\citep{ReLU} activation function, where the dimension is set to $768$.

For relational retrieval, we implement the novel GNN model based on GATv2~\citep{GATv2} provided by the torch-geometric package~\citep{torch-geometric}.
Specifically, we stack two GNN layers with the RELU activation function and also use the dropout with a probability of 0.1.
For attention in our GNN, we mask the nodes of the null entity, so that the attention score becomes zero for them. Moreover, to obtain the context representation of the entity (See Footnote 3 in the main paper) used in the GNN attention, we use the scatter operation\footnote{https://github.com/rusty1s/pytorch\_scatter} for reduced computational cost.

For Entity Memory, we experimentally found that initializing the embeddings of the entity memory with the contextualized features obtained from the pre-trained language model could be helpful. Therefore, the dimension of the entity embedding is set to the same as the language model $d=768$.
For relation embeddings, we randomly initialize them, where the dimension size is set to 128.

\paragraph{Location of KLM in the PLM} Note that, the number and location of the KFM layers inside the PLM are hyperparameters. However, we empirically found that inserting one to three KFM layers at the end of the PLM (i.e., after the 9th - 11th layers of the BERT-base language model) is beneficial to the performance (See Appendix~\ref{appendix:loc} for experiments on diverse layer locations).

\subsection{Dataset Details}
\label{appendix:dataset}
Here we describe the dataset details with its statistics for two different tasks: extractive question answering (QA) and named entity recognition (NER).

\begin{table*}[ht]
\centering
\resizebox{0.98\textwidth}{!}{
\renewcommand{\arraystretch}{0.95}
\begin{tabular}{lccccccc}
\toprule
\textbf{Hyperparameters} & \textbf{NewsQA} & \textbf{Relation} & \textbf{Medication} & \textbf{CoNLL-2003} & \textbf{WNUT-17} & \textbf{NCBI-Disease} & \textbf{Generative} NewsQA \\ 
\midrule
\multicolumn{8}{c}{\textbf{Fine-tuning}} \\
\midrule
Language Model              & \multicolumn{6}{c}{BERT-base-uncased}                                & T5-small          \\
Maximum Sequence Length     & 384    &  384     & 384        & 128        & 128     & 128          & 512               \\
Batch Size                  & 12     &  12      & 12         & 32         & 32      & 32           & 64                \\
Training Epochs             & 2      &  2       & 2          & 20         & 20      & 20           & 4                 \\
Optimizer                   &\multicolumn{6}{c}{AdamW}                                             & Adafactor         \\
Learning rate               & 3e-5   &  3e-5    & 3e-5       & 5e-5       & 5e-5    & 5e-5         & 1e-4              \\
Weight Decay                & 0.01   &  0.01    & 0.01       & 0          & 0       & 0            & -                 \\
LR decay Warmup rate        & 0.06   &  0.06    & 0.06       & 0          & 0       & 0            & -                 \\
Half Precision              & Yes    & Yes      & Yes        & No         & No      & No           & No                \\    
\midrule
\multicolumn{8}{c}{\textbf{Task-Adaptive Pre-training (TAPT)}} \\
\midrule
Maximum Sequence Length     & 384    &  384     & 384        & 128        & 128     & 128          & 384               \\
Batch Size                  & 12     &  12      & 12         & 32         & 32      & 32           & 64                \\
Training Epochs             & 1      &  1       & 1          & 3          & 3       & 3            & 4                 \\
Training Epochs (RecAdam)   & 3      &  1       & 1          & 3          & 3       & 3            & 4                 \\
Optimizer                   &\multicolumn{6}{c}{AdamW}                                             & Adafactor         \\
Learning rate               &\multicolumn{6}{c}{5e-5}                                              & 1e-3              \\
Weight Decay                & 0.01   &  0.01    & 0.01       & 0          & 0       & 0            & -                 \\
LR decay Warmup rate        & 0.06   &  0.06    & 0.06       & 0          & 0       & 0            & -                 \\
Half Precision              & Yes    & Yes      & Yes        & No         & No      & No           & No                \\ 
\bottomrule
\end{tabular}
}
\vspace{-0.1in}
\caption{\small \textbf{Hyperparamters for Fine-tuning (Top) and TAPT (Bottom)} on six datasets (+ generative QA) we used for reporting the performances in the main paper. Note that the Fine-tuning setup is applied to all methods including KALA.}
\label{table:hyper}
\vspace{-0.125in}
\end{table*}
\begin{table}
\centering
\resizebox{0.475\textwidth}{!}{
    \begin{tabular}{lcccccccc}
    \toprule
                        &  \multicolumn{2}{c}{\textbf{Training}} & \multicolumn{2}{c}{\textbf{Validation}} & \multicolumn{2}{c}{\textbf{Test}}  \\ 
    Dataset             &  \# Context & C. Length &  \# Context  & C. Length     &  \# Context & C. Length             \\
    \midrule
    \textbf{CoNLL}-2003 & 14,041 & 19.95 &  3,250 & 21.36 & 3,453 & 18.77             \\
    \textbf{WNUT}-17    & 3,394 & 31.32 & 1,009 & 19.28 & 1,287 & 30.58             \\
    \textbf{NCBI}-Disease & 5,433 & 34.36 & 924 & 35.00 & 941 & 35.50           \\
    \bottomrule
    \end{tabular}
}
\vspace{-0.1in}
\caption{\small \textbf{NER dataset statistics.} We report the number of contexts (i.e., \# Context), with the average length of them (i.e., C. Length) on training, validation, and test sets.}
\vspace{-0.2in}
\label{table:nerstat}
\end{table}

\paragraph{Question Answering}
We evaluate models on three domain-specific datasets: NewsQA, Relation, and Medication. 
Notably, NewsQA~\citep{NewsQA} is curated from CNN news articles. Relation and Medication are originally part of the emrQA~\citep{emrQA}, which is an automatically constructed question answering dataset based on the electrical medical record from n2c2 challenges\footnote{https://portal.dbmi.hms.harvard.edu/projects/n2c2-nlp/}. However, \citet{CliniRC} extract two major subsets by dividing the entire dataset into Relation and Medication and suggest the usage of sampled questions from the original emrQA dataset. Following the suggestion of~\citet{CliniRC}, we use only 1\% of generated questions of Relation for training, validation, and testing. Also, we only use 1\% of generated questions of Medication for training and use 5\% of generated questions of Medication for validation and testing. Since the original emrQA is automatically generated based on templates, the quality is poor -- it means that the original emrQA dataset was inappropriate to evaluate the ability of the model to reason over the clinical text since the most of questions can be answered by the simple text matching. To overcome this limitation, \citet{CliniRC} suggests two ways to make the task more difficult. First, they divide the question templates into easy and hard versions and then use the hard question only. Second, they suggest replacing medical terminologies in the question of the test set into synonyms to avoid the trivial question which can be solvable with a simple text matching. We use both methods to Relation and Medication datasets to report the performance of every model. For more details on Relation and Medication datasets, please refer to the original paper~\citep{CliniRC}.
The statistics of training, validation, and test sets on all QA datasets are provided in Table~\ref{table:qastat}.

\begin{table}[th]
\centering
\resizebox{0.98\linewidth}{!}{
\begin{tabular}{lcc}
\toprule
\textbf{Hyperparameters}    & \textbf{News} & \textbf{Medical Textbook} \\ 
\midrule
\multicolumn{3}{c}{\textbf{Domain-Adaptive Pre-training (DAPT)}} \\
\midrule
The number of text (by lines)& 10M    &  100k     \\
The number of text (by words)& 618M    &  12.8M     \\
The size of data (by volume)& 3.5G   &  86M      \\
Maximum Sequence Length     & \multicolumn{2}{c}{384}      \\
Batch Size                  & \multicolumn{2}{c}{64}      \\
Training Epochs             & \multicolumn{2}{c}{50}       \\
Maximum Steps               & \multicolumn{2}{c}{12.5k}    \\
Optimizer                   & \multicolumn{2}{c}{AdamW}   \\
Learning rate               & \multicolumn{2}{c}{5e-5}     \\
Weight Decay                & \multicolumn{2}{c}{0.01}    \\
LR decay Warmup rate        & \multicolumn{2}{c}{0.06}    \\
Half Precision              & \multicolumn{2}{c}{Yes}    \\
Applied Dataset             & \makecell{NewsQA\\CoNLL-2003\\WNUT-17} & \makecell{Relation\\Medication\\NCBI-Disease} \\
\bottomrule
\end{tabular}
}
\vspace{-0.1in}
\caption{\small \textbf{Hyperparamters for DAPT} on two domains we used for reporting the performances in the main paper.}
\label{table:hyper-dapt}
\vspace{-0.2in}
\end{table}

\paragraph{Named Entity Recognition}
We use three different domain-specific datasets for evaluating our KALA on NER tasks: CoNLL-2003~\cite{CoNLL} (News), WNUT-17~\cite{WNUT} (Social Network Service) and NCBI-Disease~\cite{NCBI} (Biomedical). The CoNLL-2003 is constructed from the manually curated 1,393 English news articles, including 301.4k tokens, which has 9 class labels. The WNUT-17 dataset consists of 65,124 emerging and rare entities from social media (e.g., Twitter, Reddit, YouTube, to name a few), which has 13 class labels. The NCBI-Disease dataset consists of the 793 PubMed articles from the biomedical domain, which contains 6,892 disease mentions and 790 disease concepts, and also has 3 class labels. The statistics of training, validation, and test sets are provided in Table~\ref{table:nerstat}.

\subsection{Training details}
\label{appendix:training}
All experiments are constrained to be done with a single 12GB Geforce RTX 2080 Ti GPU for fairness in terms of memory and the availability on the academic budget, except for the DAPT and generative QA which use a single 48GB Quadro 8000 GPU. KALA training needs 3 hours in wall time with a single GPU. For all experiments, we select the best checkpoint on the validation set.
For the summary of training setups, please see Table~\ref{table:hyper} and~\ref{table:hyper-dapt}.

\paragraph{Fine-tuning Setup} In the following three paragraphs, we explain the setting of fine-tuning for QA, NER, and generative QA tasks.
For all experiments on extractive QA tasks, we fine-tune the Pre-trained Language Model (PLM) for 2 epochs with the weight decay of 0.01, learning rate of 3e-5, maximum sequence length of 384, batch size of 12, linear learning rate decay of 0.06 warmup rate, and half precision~\citep{fp16}.

For all experiments on NER tasks, we fine-tune the PLM for 20 epochs, where the learning rate is set to 5e-5, maximum sequence length is set to 128, and batch size is set to 32.
We use AdamW~\citep{AdamW} as an optimizer using BERT-base as the PLM.

For the generative QA task in Table~\ref{table:generative-qa}, we fine-tune the T5-small~\citep{T5} for 4 epochs with the learning rate of 1e-4, maximum sequence length of 512, and batch size of 64. We also use the Adafactor~\citep{Adafactor} optimizer. Instead of training with the same optimizer as in BERT for QA and NER, we instead use the independent AdamW optimizer with the learning rate of 1e-4 and weight decay of 0.01 to train the KALA module with T5.

\paragraph{Adaptive Pre-training Setup}
In this paragraph, we describe the experimental settings of adaptive pre-training baselines, namely TAPT, TAPT (+ RecAdam), and DAPT. For QA tasks, we further pre-train the PLM for \{1,3,5,10\} epochs and then report the best performance among them. Specifically, reported TAPT result on NewsQA, Relation, and Medication are obtained by 1 epoch of further pre-training. We use the weight decay of 0.01, learning rate of 5e-5, maximum sequence length of 384, batch size of 12, and linear learning rate decay of 0.06 warmup rate, with a half-precision. Also, the masking ratio for the pre-training objective is set to 0.15, following the existing strategy introduced in the original BERT paper~\citep{BERT}.

For NER tasks, we further pre-train the PLM for 3 epochs across all datasets. In particular, the learning rate is set to 5e-5, batch size is set to 32, and the maximum sequence length is set to 128. 
We also use AdamW~\citep{AdamW} as the optimizer for all experiments.

In the case of T5-small for generative QA in Table~\ref{table:generative-qa}, we further pre-train the PLM for 4 epochs with the learning rate of 0.001, batch size of 64, maximum sequence length of 384, and Adafactor~\citep{Adafactor} optimizer.

Regarding the setting of TAPT (+ RecAdam) on all tasks, we follow the best setting in the original paper~\citep{RecAdam} -- sigmoid as an annealing function with annealing parameters: $k = 0.5$, $t_0 = 250$, and the pretraining coefficient of 5000.

For training with DAPT, we need an external corpus having a large amount of data for adaptive pre-training. Thus, we first choose the datasets of two domains -- News and Medical. Specifically, as the source of corpus for the News domain, we use the sampled set of 10 million News from the RealNews dataset used in~\citet{demix}. As the source of corpus for the Medical domain, we use the set of approximately 100k passages from the Medical textbook provided in~\citet{MedExam}. The size of pre-training data used in DAPT is much larger than TAPT. In other words, for experiments on NewsQA, TAPT only uses fine-tuning contexts containing 5.8 million words from the NewsQA training dataset, while DAPT uses more than a hundred times larger data -- enormous contexts containing about 618 million words from the RealNews database. For both News and Medical domains, we further pre-train the BERT-base model for 50 epochs with the batch size of 64, to match the similar computational cost used in~\citet{dontstoppt}. Other experimental details are the same as TAPT described above.

\subsection{Architectural Variant Details}
\label{appendix:variant}
In this subsection, we describe the details of architectural variants reported in Section~\ref{sec:5.1}. For all variants, we use the same KGs used in KALA.

\textbf{Entity-as-Experts} (\citet{EaE}; EaE) utilizes the entity memory similar to our work, but they use the parametric dense retrieval more like the memory neural network~\citep{MemNN}. Similar to~\citet{EaE, FaE}, we change the formulation of query and memory retrieval by using the mention representation of the entity from the intermediate hidden states of PLMs, which is formally defined as follows:
\begin{align}
\label{eqn:mentionembed}
    &\vh_e = \frac{1}{m^\omega-m^\alpha+1} \sum_{i=m^\alpha}^{m^\omega} \vh^{l-1}_i, \\
    &\vv = \softmax (\vh_e \cdot \mE^\top) \cdot \mE, \nonumber
\end{align}
where $\vh_e$ represents the average of token representations of the entity mention $m=(m^\alpha, m^\omega)$. We also give the supervised retrieval loss (\texttt{ELLoss} in~\citet{EaE}), when training the EaE model. With this retrieval, EaE also can represent the unseen entity $e \notin \mathcal{E}_{train}$ if we know the mention boundary of the given entity on the context. We believe it is expected to work well, if the entity memory is pre-trained on the enormous text along with the pre-training of the language model from the scratch. However, it might underperform for the language model adaptation scenario, since it can fall into the problem of circular reasoning -- the PLM does not properly represent the unseen entity, but it should predict which entity it is similar from the representation.
Regarding the integration of the knowledge from the entity memory into the PLM, the retrieved entity representation $\vv$ is simply added~\citep{KnowBERT} to the hidden representations $\mH$ after the transformer block as follows:
\begin{equation}
    \tilde{\mH}^l = \mH^l + h(\vv) 
    \label{eqn:EaE}
\end{equation}
where $h$ is Multi-Layer Perceptrons (MLPs).

\textbf{Adapter}~\citep{HoulsbyAdapter} is introduced to fine-tune the PLM only with a few trainable parameters, instead of fine-tuning the whole parameters of the PLM. To adapt this original implementation into our KALA framework, we replace our Knowledge-conditioned Feature Modulation with it, where the Adapter is used as the knowledge integration module. We interleave the layer of Adapter after the feed-forward layer ($FF$) and before the residual connection of the transformer block. Also, instead of only providing the LM hidden states as an input, we concatenate the knowledge representation in Equation~\ref{eqn:relembed} to the LM hidden states. Note that we fine-tune the whole parameters following our KALA setting, unlike fine-tuning the parameters of only Adapter layers in~\citet{HoulsbyAdapter}.

\textbf{ERNIE}~\citep{ERNIE} is a notable PLM model that utilizes the external KB as an input for the language model. The key feature of ERNIE can be summarized into two folds. First, they use the multi-head self-attention scheme~\citep{Transformer} to contextualize the input entities. Second, ERNIE fuses the entity representation at the end of the PLM by adding it to the corresponding language representation. We assume that those two features are important points of ERNIE. Therefore, instead of using a Graph Neural Network (GNN) layer, we use a multi-head self-attention layer to contextualize the entity embeddings. Then, we add it to a representation of the entity from the PLM, which is the same as the design in equation~\ref{eqn:EaE}.

\textbf{KT-Net}~\citep{KTNet} uses knowledge as an external input in the fine-tuning stage for extractive QA. Since they have a typical layer for integrating existing KB~\citep{WordNet,NELL} with the PLM, we only adopt the self-matching layer as the architecture variant of the KFM layer used in our KALA framework. The computation of the self-matching matrix in KT-Net is costly, i.e., it requires a large computational cost that is approximately 12 times larger than KALA.

\textbf{ERICA}~\citep{ERICA} uses contrastive learning in LM pre-training to reflect the relational knowledge into the language model. We use the Entity Discrimination task from ERICA on the primary task of fine-tuning. We would like to note that, as reported in Section 5 of the original paper~\citep{ERICA}, the use of ERICA on fine-tuning has no effect, since the size and diversity of entities and relations in downstream training data are limited. Such limited information rather harms the performance, as it can hinder the generalization. In other words, contrastive learning cannot reflect the entity and relation in the test dataset.

\subsection{FLOPs Computation}
In this subsection, we give detailed descriptions of how the FLOPs in Figure~\ref{fig:main} are measured.
We majorly follow the script from the ELECTRA~\citep{ELECTRA} repository to compute the approximated FLOPs for all models including ours.
For FLOPs computation of our KALA, we additionally include the FLOPs of the entity embedding layer, linear layers for $h_1, h_2, h_3, h_4$, and GNN layer.
Since the GNN layer is implemented based on the sparse implementation, we first calculate the FLOPs of the message propagation over one edge, and then multiply it to the average number of edges per node. Also, in terms of the computation on mentions, we consider the maximum sequence length of the context rather than the average number of mentions, to set the upper bound of FLOPs for our KALA. Note that, in NewsQA training data, the average number of nodes is 57, the average number of edges for each node is 0.64, and the average number of mentions in the context is 92.68.

\section{Additional Experimental Results}
In this section, we provide the analyses on the forgetting of TAPT, entity memory, number of entities and facts, location of the KLM layer, and values of Gamma and Beta.

\begin{figure}[t]
    \centering
    \includegraphics[width=1.0\linewidth]{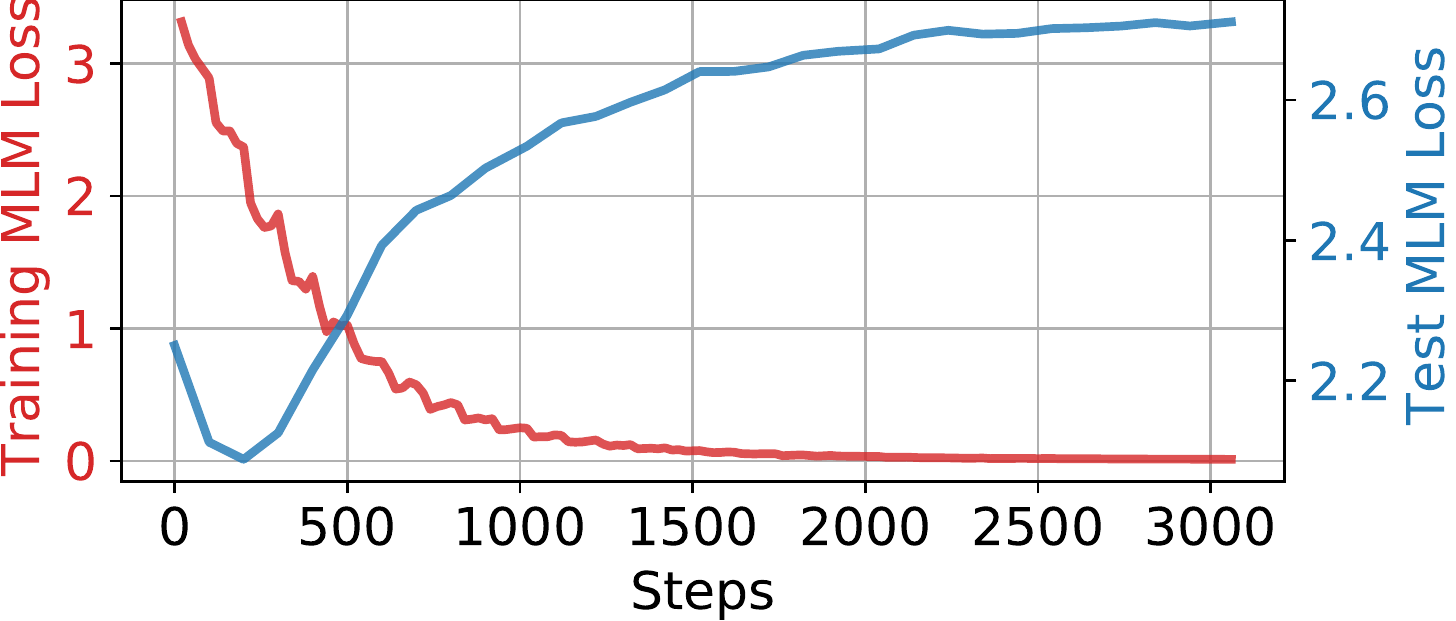}
    \vskip -0.1in
    \caption{\small Masked Language Model loss from Task-Adaptive Pre-Training on the domain-specific training dataset (Relation) and the general domain test dataset (Sampled wikipedia).}
    \label{fig:mlm_loss}
\end{figure}

\subsection{Analysis on forgetting of TAPT}
In Figure~\ref{fig:main}, we observe that the performance of TAPT decreases as the number of training steps increases. To get a concrete intuition on this particular phenomenon, we analysis what happens in the Pre-trained Language Model (PLM), when we further pre-train it on the task-specific corpus. 
Specifically, in Figure~\ref{fig:mlm_loss}, we visualize the Masked Language Model (MLM) loss of TAPT on both domain-specific corpus from the Relation dataset and general corpus from the sampled Wikipedia documents during the adaptive pre-traing. 
As Figure~\ref{fig:mlm_loss} shows, the test MLM loss increases while the training MLM loss persistently increases as the training step increases. This result indicates that TAPT on domain-specific corpus may yield the catastrophic forgetting of the general knowledge in the PLM.

\begin{figure}[t]
    \centering
    \includegraphics[width=1.0\linewidth]{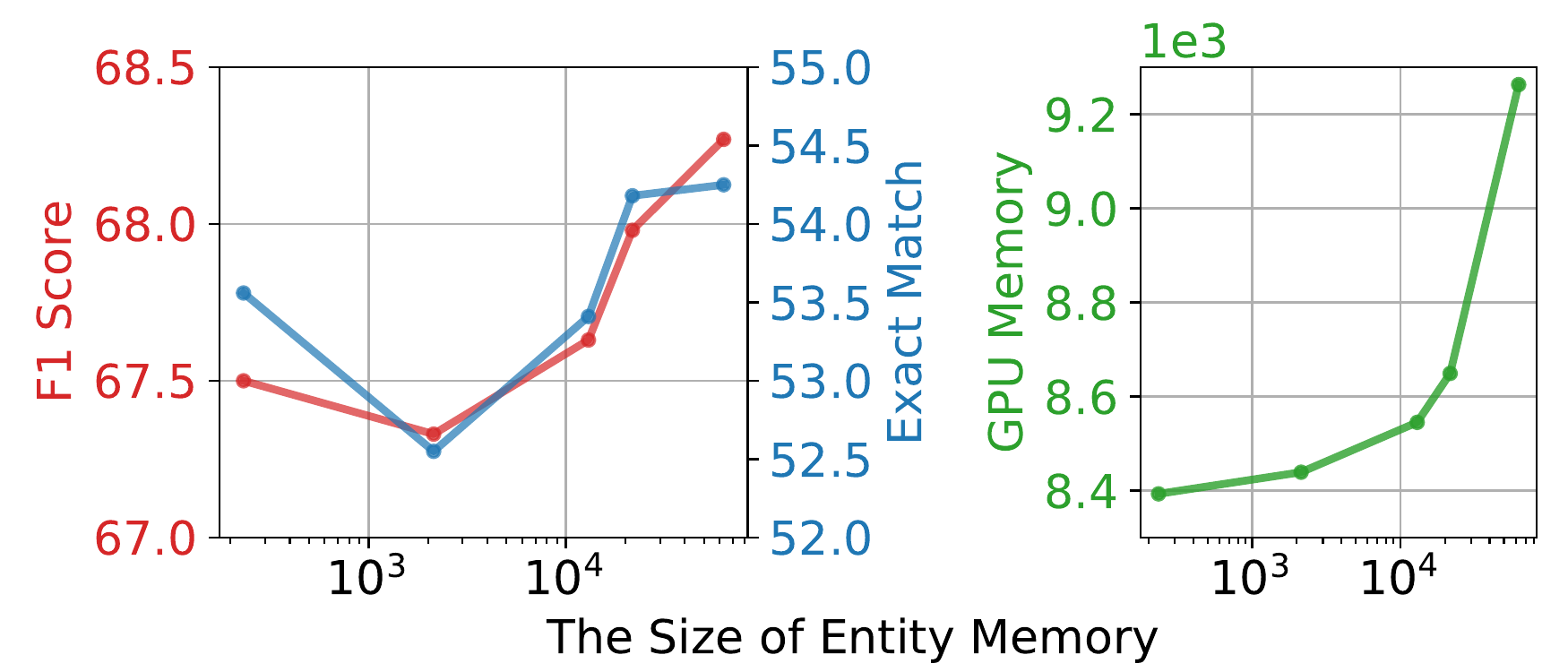}
    \vspace{-0.3in}
    \caption{\small The performance (F1 score and Exact Match) and the GPU memory usage on NewsQA dataset with varying the size of elements in the entity memory.}
    \label{fig:memory}
    \vspace{-0.2in}
\end{figure}

\subsection{Effects of the Size of Entity Memory}
\label{appendix:memorysize}
In this subsection, we analyze how the size of entity memory affects the performance of our KALA.
In Figure~\ref{fig:memory}, we plot the performance of KALA on the NewsQA dataset by varying the number of entity elements in the memory.
Note that, we reduce the size of the entity memory by eliminating the entity appearing fewer times. Thus, the results are obtained by only considering the entities that appear more than $[1000, 100, 10, 5, 0]$ times, e.g., 0 means the model with full entity memory.
As shown in Figure~\ref{fig:memory}, we observe that the size of the entity memory is larger, the performance of our KALA is better in general. 
However, interestingly, we also observe that the smallest size of the entity memory shows decent performance, which might be due to the fact that some parameters in the entity memory are stale. For more discussions on it including visualization, please refer to Appendix~\ref{appendix:frequency}.
Finally, we would like to note that, in Figure~\ref{fig:main}, we report the performance of our KALA in the case of $[1000, 5, 0]$ (i.e., considering entities appearing more than $[1000, 5, 0]$ times).

\begin{figure}[t!]
    \begin{minipage}{0.495\linewidth}
        \centering
        \includegraphics[width=1\linewidth]{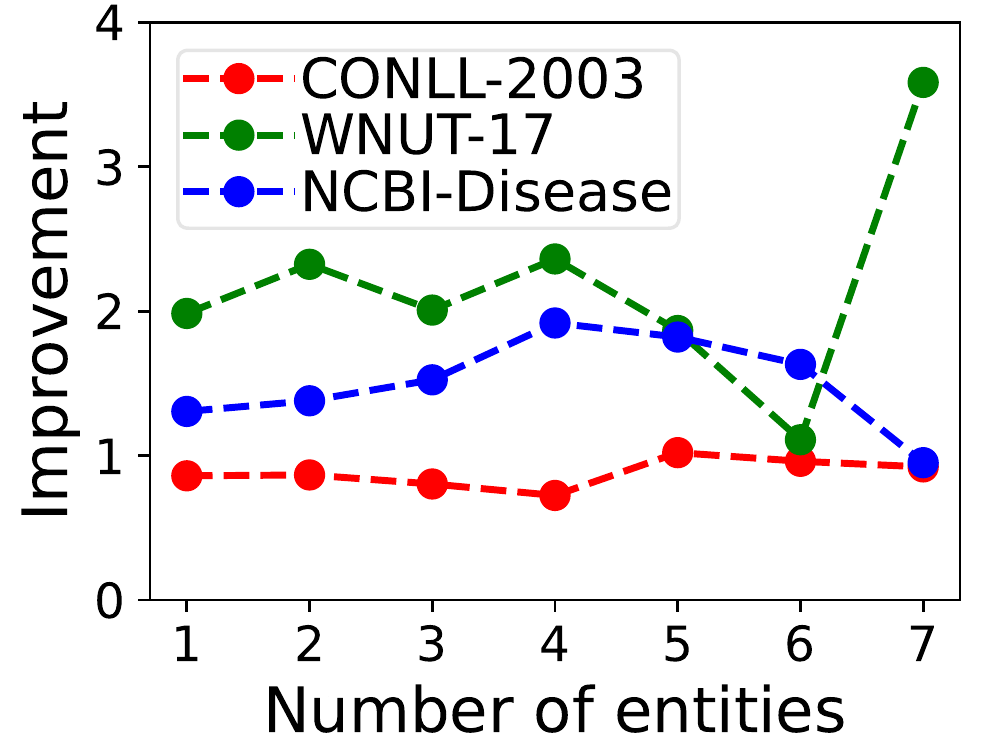}
    \end{minipage}
    \begin{minipage}{0.495\linewidth}
        \centering
        \includegraphics[width=1\linewidth]{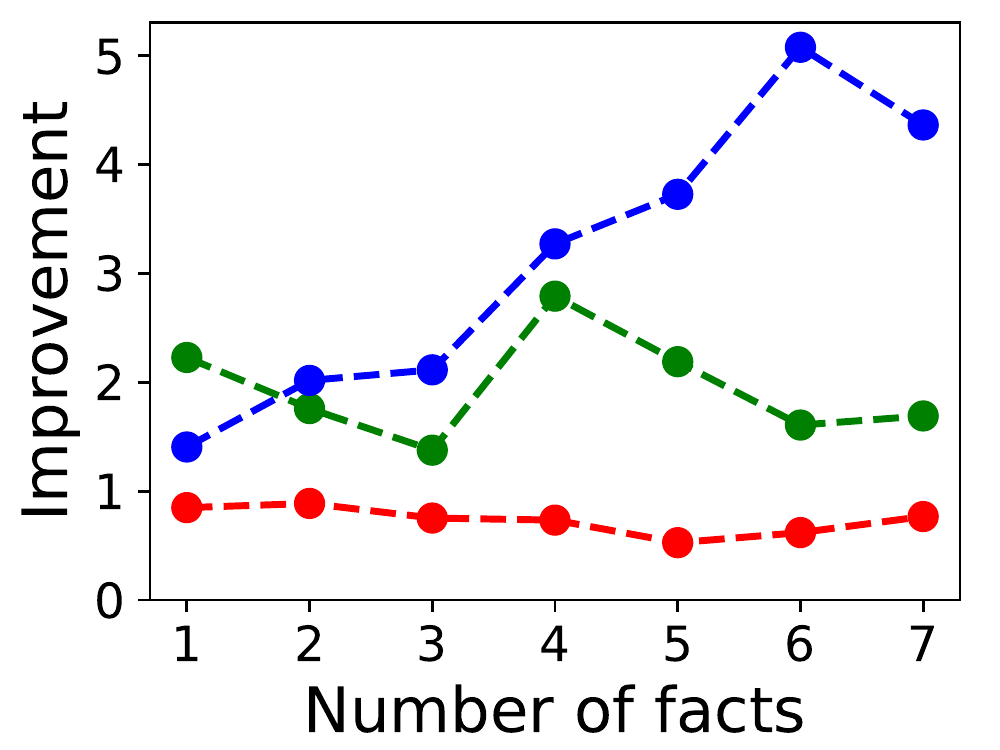}
    \end{minipage}
    \vspace{-0.1in}
    \caption{\small Performance improvements of our KALA from simple fine-tuning, with varying the number of entities and facts in the context on Named Entity Recognition tasks.}
    \label{fig:entity_fact}
\end{figure}
\begin{figure}[t]
    \centering
    \includegraphics[width=1.0\linewidth]{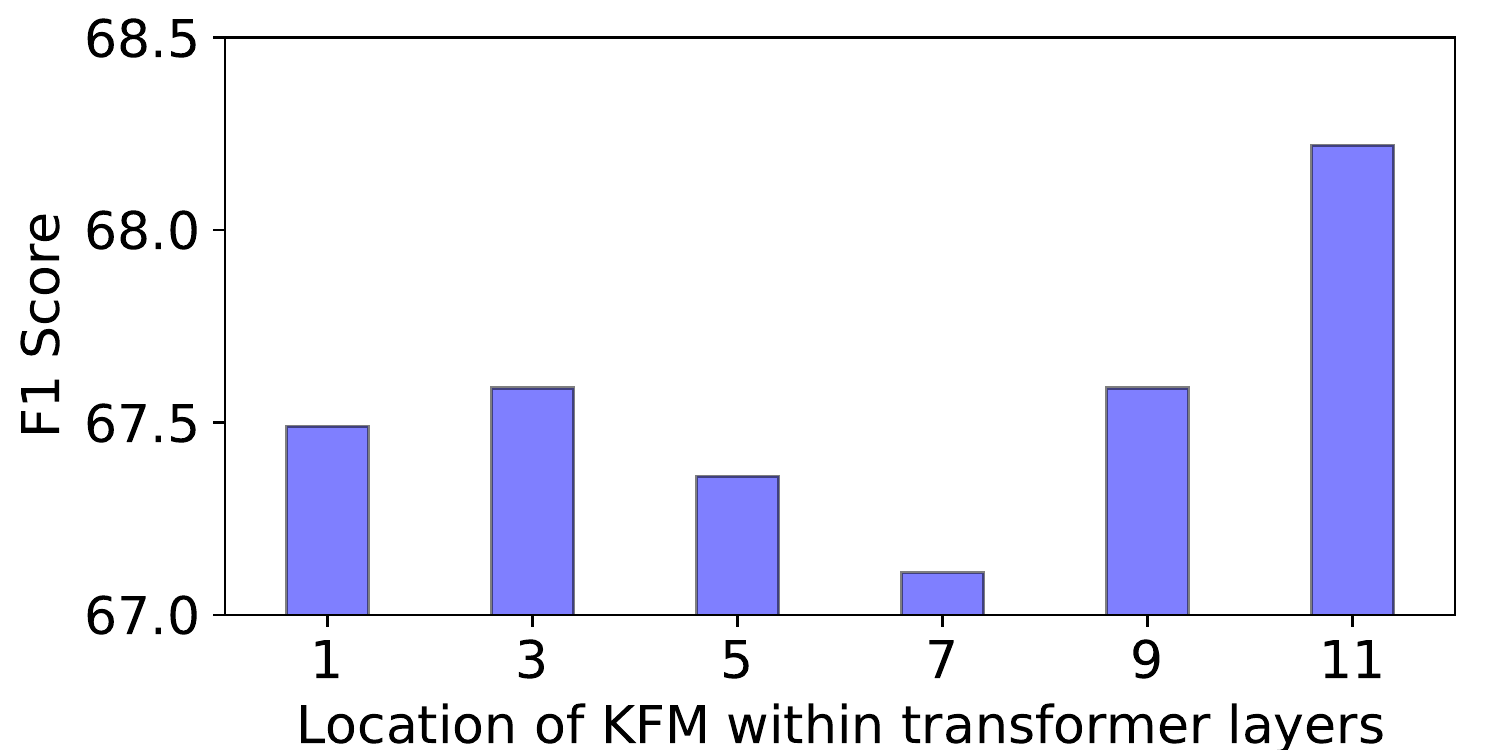}
    \vskip -0.1in
    \caption{\small The performance of our KALA with varying the location of the KFM layer inside the BERT-base model. y-axis denotes the F1 score on NewsQA and x-axis denotes the location of the KFM layer. For instance, 11 means the case where the KFM layer is appended in the 11th transformer layer of BERT-base.}
    \label{fig:loc}
    \vskip -0.2in
\end{figure}

\subsection{Effects of the Number of Entity and Fact}
In this subsection, we aim to analyze which numbers of entities and facts per context are appropriate to achieve good performance in NER tasks. Specifically, we first collect the contexts having more than or equal to the $k$ number of entities (or facts), and then calculate the performance difference from our KALA to the fine-tuning baseline. As shown in Figure~\ref{fig:entity_fact}, while there are no obvious patterns, performance improvements from the baseline are consistent across a varying number of entities and facts. This result suggests that our KALA is indeed beneficial when entities and facts are given to the model, whereas the appropriate number of entities and facts to obtain the best performance against the baseline is different across datasets.

\subsection{Effects of the Location of KFM}
\label{appendix:loc}
In the main paper and Appendix~\ref{appendix:implement}, we describe that the location of the KFM layer inside the PLM architecture is the hyperparameter.
However, someone might wonder which location of KFM yields the best performance, and what is the reason for this. 
Therefore, in this section, we analyze where we obtain the best performance in various locations of the KFM layer on the NewsQA dataset. 
Specifically, in Figure~\ref{fig:loc}, we show the performance of our KALA with varying the location of the KFM layer insider the BERT-base model.
The results demonstrate that the model with the KFM on the last layer of the BERT-base outperforms all the other choices. This might be because, as the final layer of the PLM is generally considered as the most task-specific layer, our KFM interleaved in the latest layer of BERT expressively injects the task-specific information from the entity memory and KGs, to such a task-specific layer.

\begin{figure*}[t!]
    \begin{minipage}{0.24\linewidth}
        \centering
        \includegraphics[width=1\linewidth]{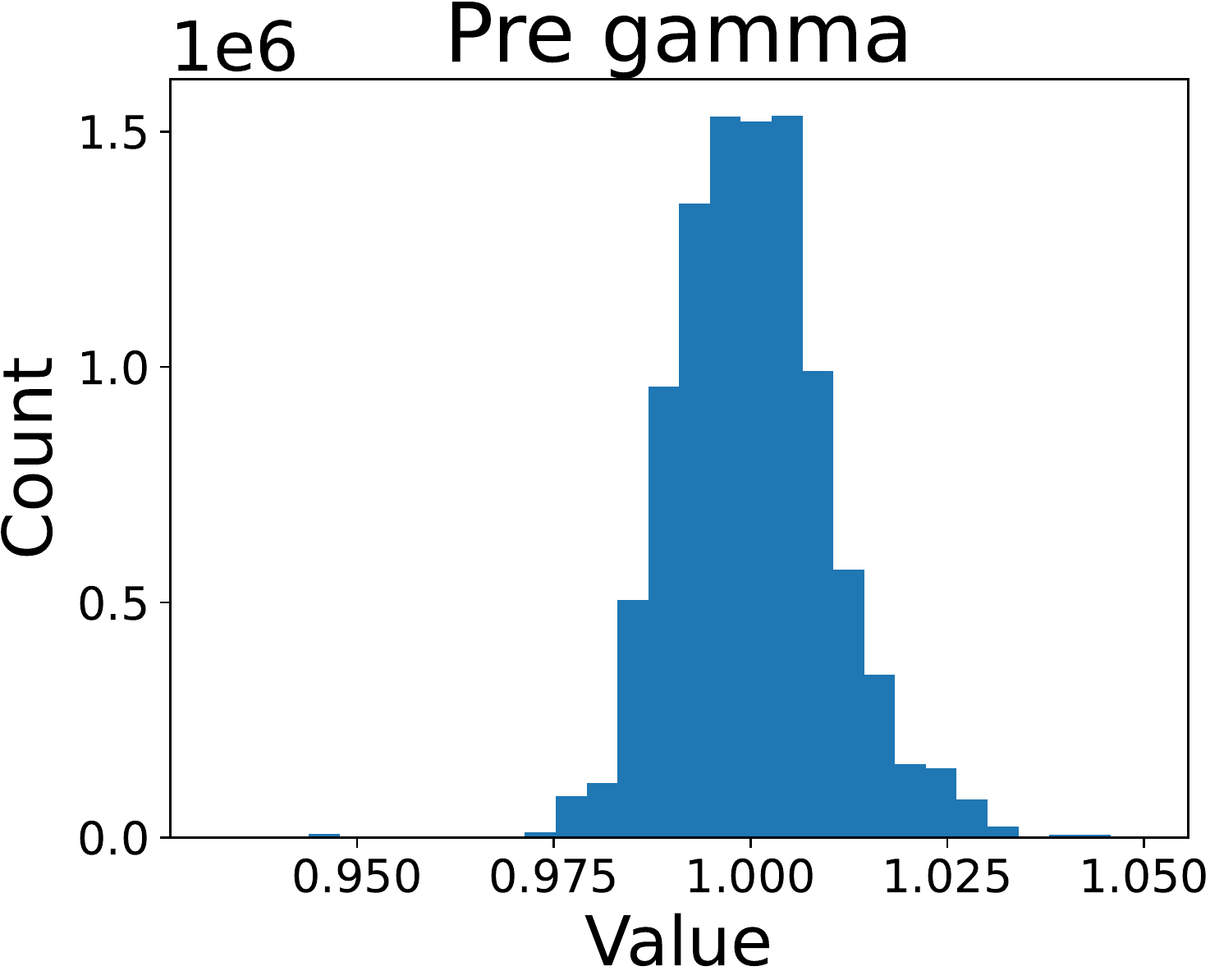}
    \end{minipage}
    \hfill
    \begin{minipage}{0.24\linewidth}
        \centering
        \includegraphics[width=1\linewidth]{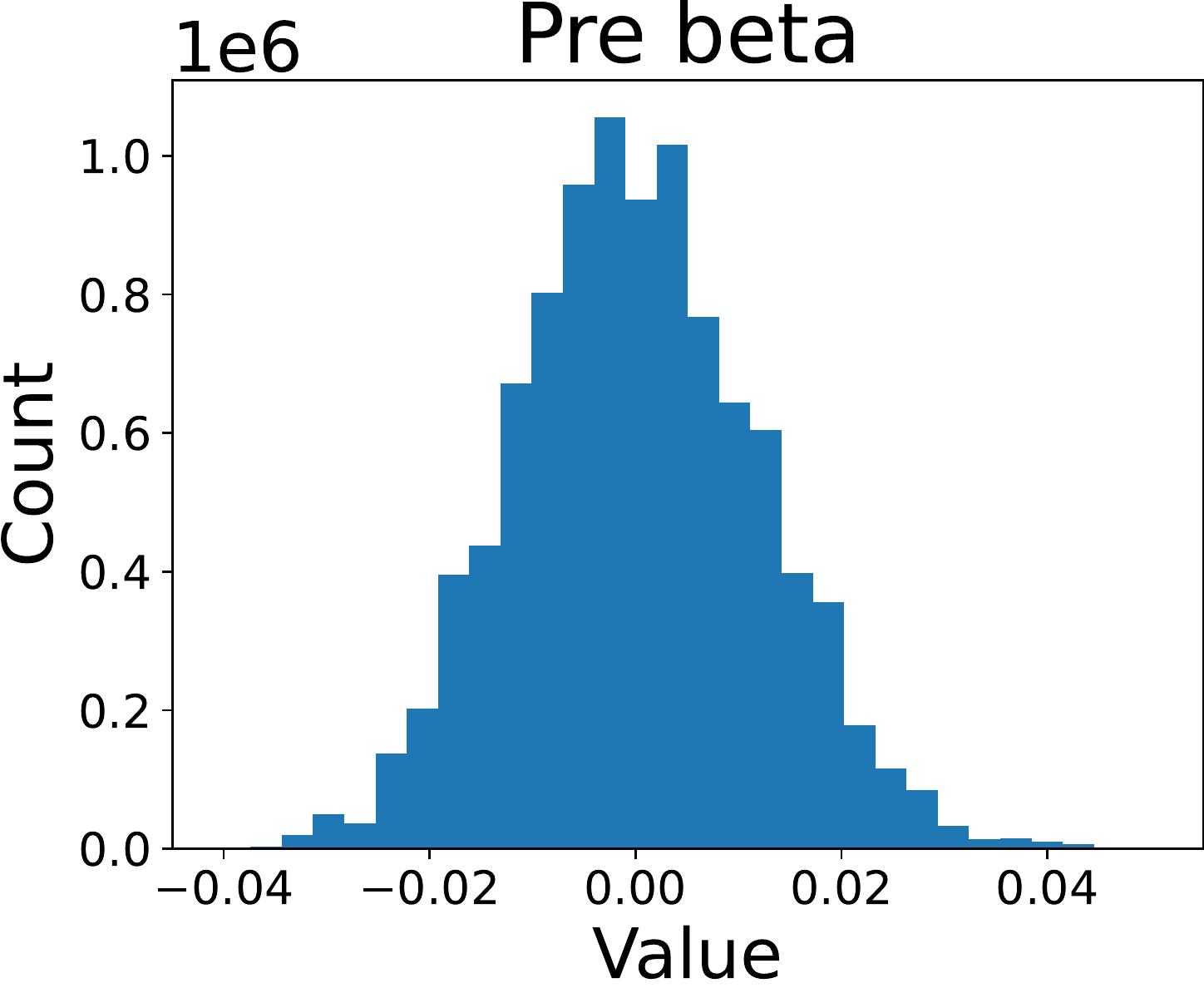}
    \end{minipage}
    \hfill
    \begin{minipage}{0.24\linewidth}
        \centering
        \includegraphics[width=1\linewidth]{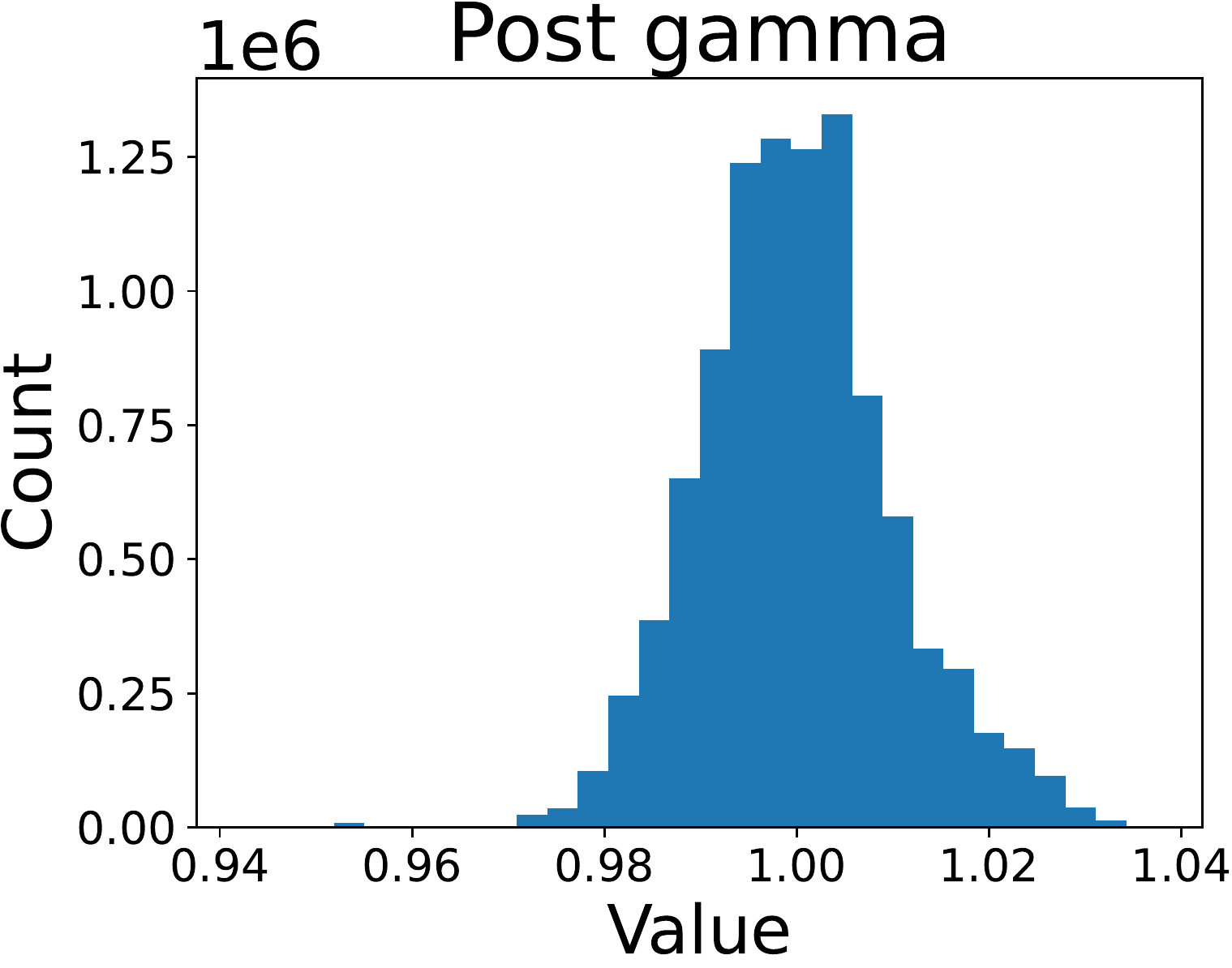}
    \end{minipage}
    \hfill
    \begin{minipage}{0.24\linewidth}
        \centering
        \includegraphics[width=1\linewidth]{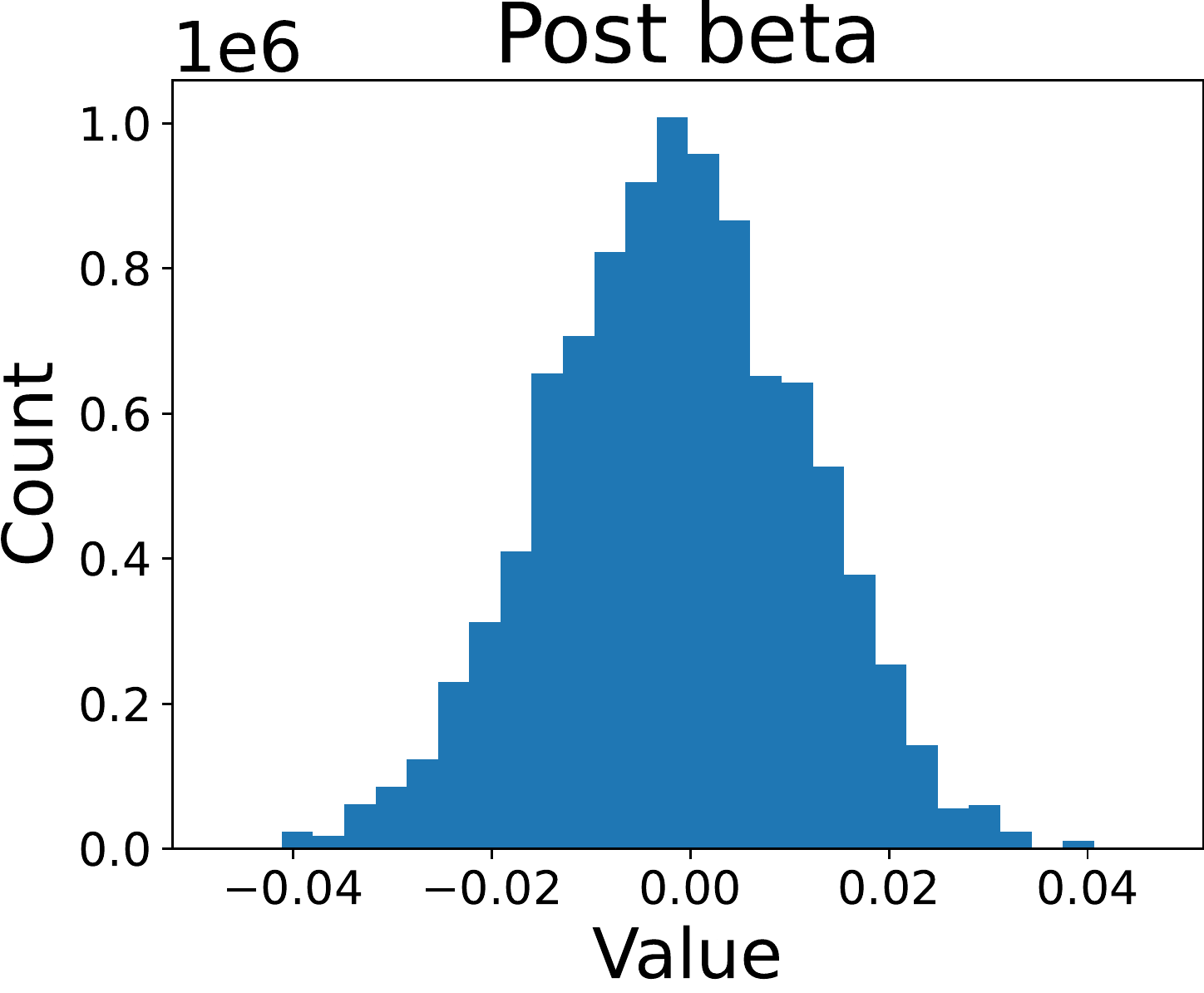}
    \end{minipage}
    \vspace{-0.15in}
    \caption{\small Histogram of values of gamma and beta on the CoNLL-2003 dataset.}
    \label{fig:gamma_beta}
    \vspace{-0.1in}
\end{figure*}
\begin{figure}[t!]
    \centering
    \vspace{-0.05in}
    \begin{minipage}{0.45\linewidth}
        \centering
        \includegraphics[width=1\linewidth]{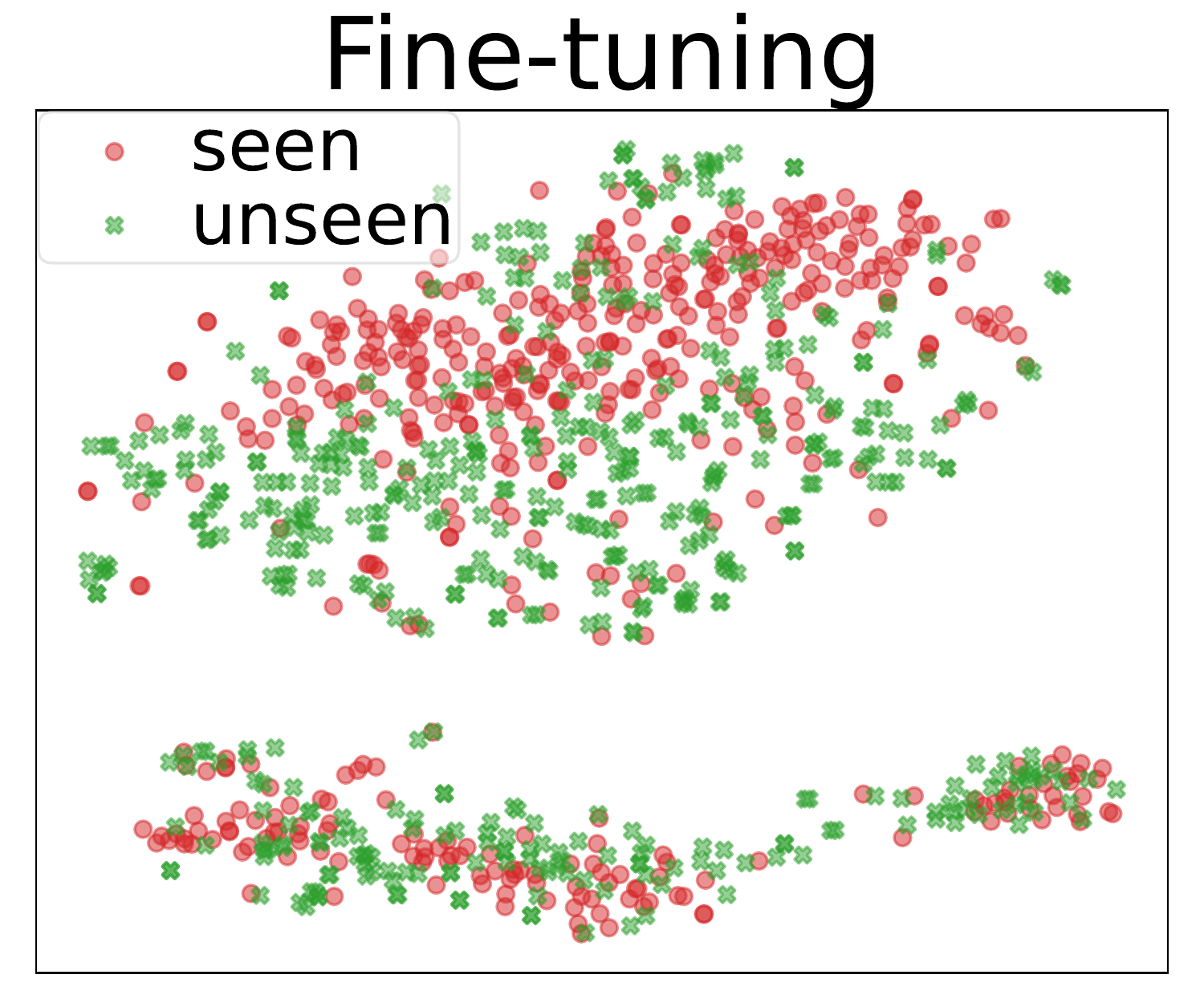}
    \end{minipage}
    \begin{minipage}{0.45\linewidth}
        \centering
        \includegraphics[width=1\linewidth]{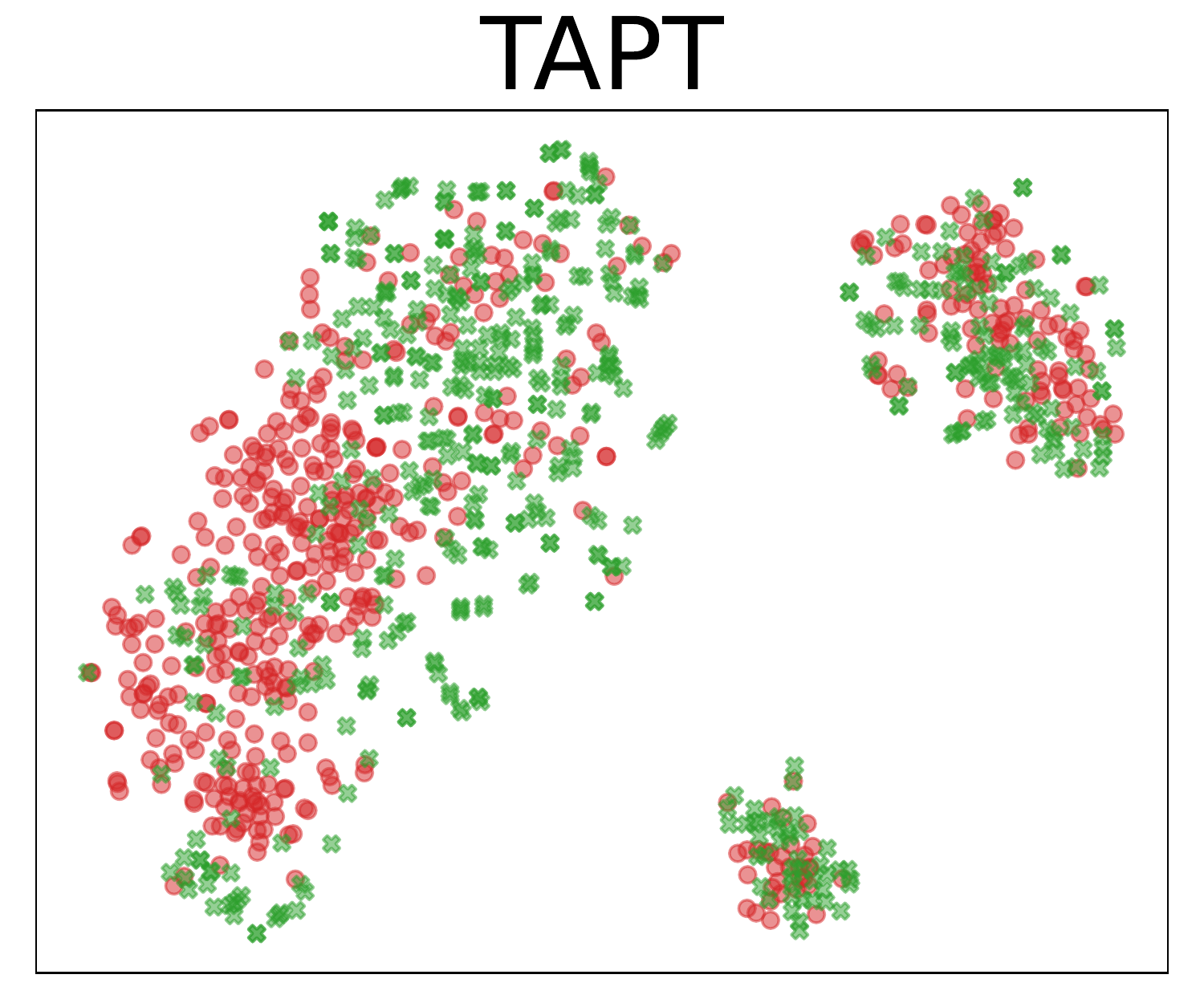}
    \end{minipage}
    \\
    \begin{minipage}{0.45\linewidth}
        \centering
        \includegraphics[width=1\linewidth]{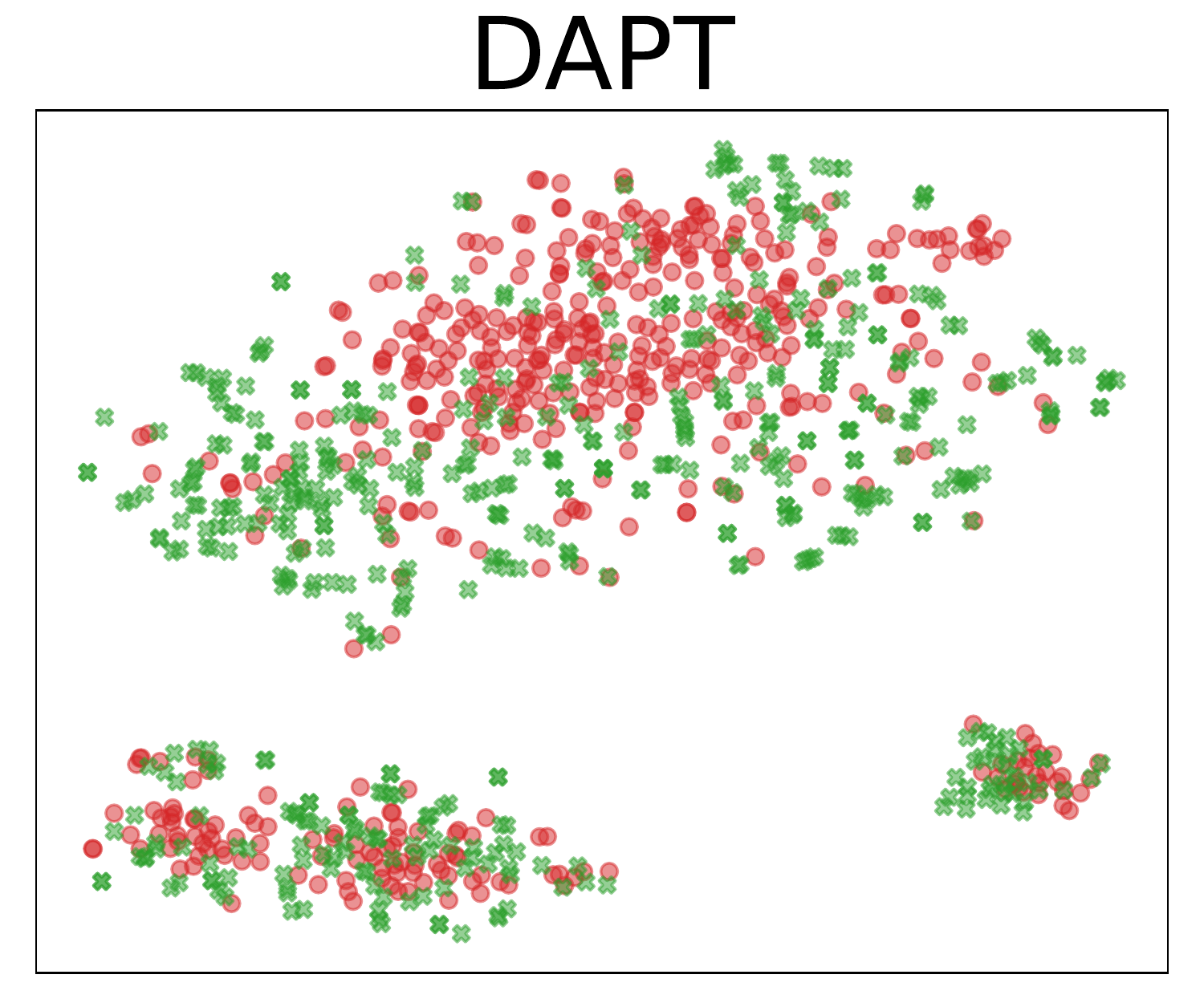}
    \end{minipage}
    \begin{minipage}{0.45\linewidth}
        \centering
        \includegraphics[width=1\linewidth]{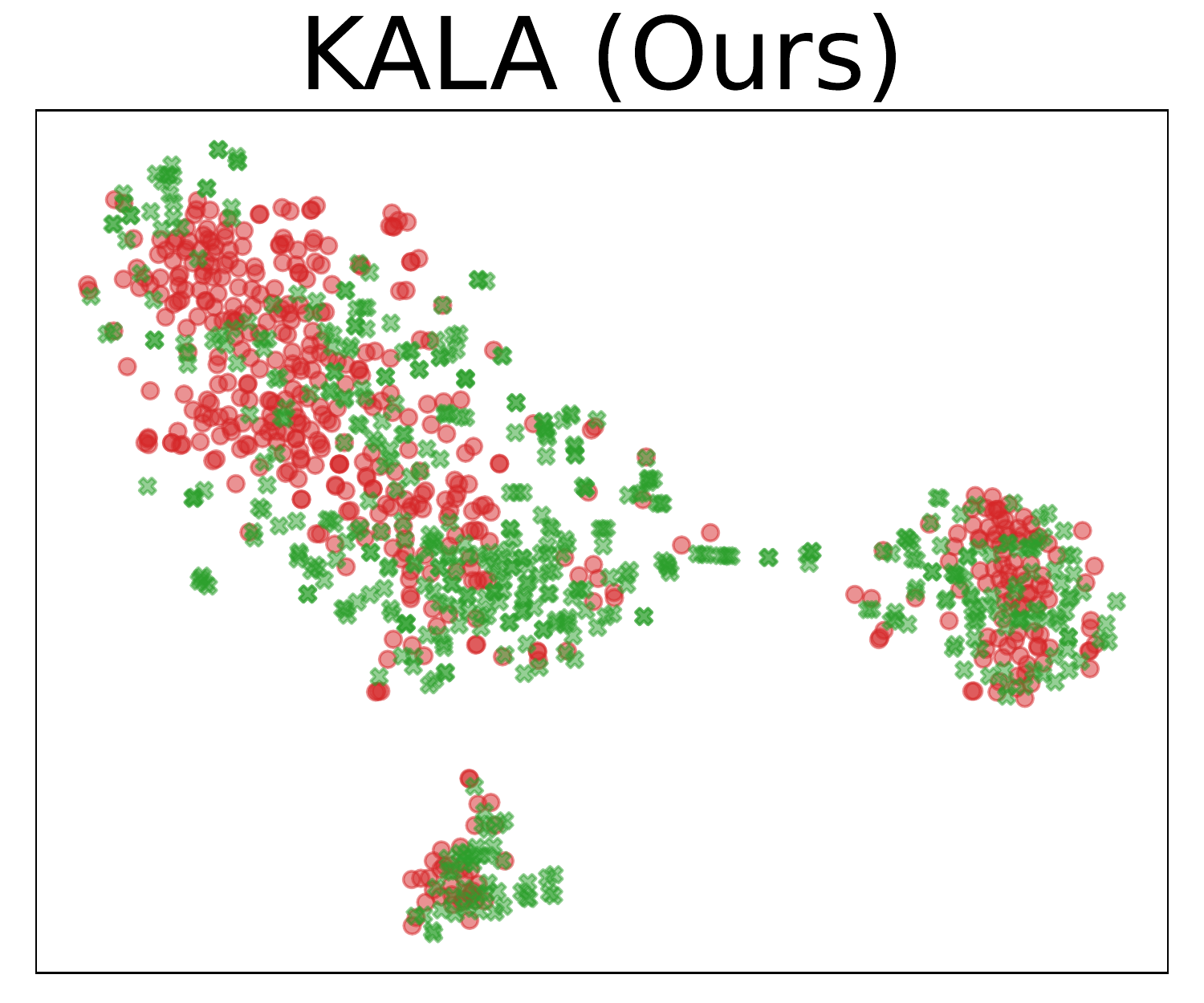}
    \end{minipage}
    \vspace{-0.1in}
    \caption{\small Visualization of contextual representations for seen and unseen entities on the NCBI-Disease dataset.}
    \label{fig:unseen_embedding}
    \vspace{-0.15in}
\end{figure}

\subsection{Analysis on Values of Gamma and Beta}
To see how much amount of value on gamma and beta is used to shift and scale the intermediate hidden representations in transformer layers, we visualize the modulation values, namely gamma and beta, in Figure~\ref{fig:gamma_beta}. We first observe that, as shown in Figure~\ref{fig:gamma_beta}, the distribution of values of gamma and beta approximately follow the Gaussian distribution, with zero mean for beta and one mean for gamma. Also, we notice that the scale of values remain nearly around the mean point, which suggests that the small amount of shifting to intermediate hidden representations on transformer layers is enough to contribute to the performance gain, as we can see in the main results of Table~\ref{qa-exp},~\ref{ner-exp}.

\subsection{Detailed Efficiency Comparison}
\label{appendix:efficiecy-detail}
While we provide the efficiency on FLOPs in Figure~\ref{fig:main}, we further provide the efficiency on GPU memory, wall time, and FLOPs for training each method in Table~\ref{table:efficiency}. Specifically, we measure the computational cost on the NewsQA dataset with BERT-base, where we use the single Geforce RTX 2080 Ti GPU on the same machine. For our KALA, as we can flexibly manage the cost of GPU memory by reducing the number of entities in entity memory (See Figure~\ref{fig:memory} with Appendix~\ref{appendix:memorysize} for more analysis on the effects of the size of entity memory), we provide the experimental results on two settings -- KALA with memory size 0.2k and 62.8k (full memory). As shown in Table~\ref{table:efficiency}, we confirm that the computational cost of our KALA with the full memory is similar to the cost of the \textit{more params} baseline that uses one additional transformer layer on top of BERT-base. However, by reducing the number of entities in the memory, we can achieve better efficiency than more params in terms of GPU memory and FLOPs. Also, we observe that the training cost (i.e., Wall Time and FLOPs) of TAPT and DAPT is high, especially on DAPT, thus we verify that our KALA is more efficient to train for domain adaptation settings.

\section{Additional Visualization Results}
Here we provide the frequency distribution of entities, additional case studies, and more illustrations of textual examples and embedding spaces.

\begin{figure}[t!]
    \begin{minipage}{0.325\linewidth}
        \centering
        \includegraphics[width=1\linewidth]{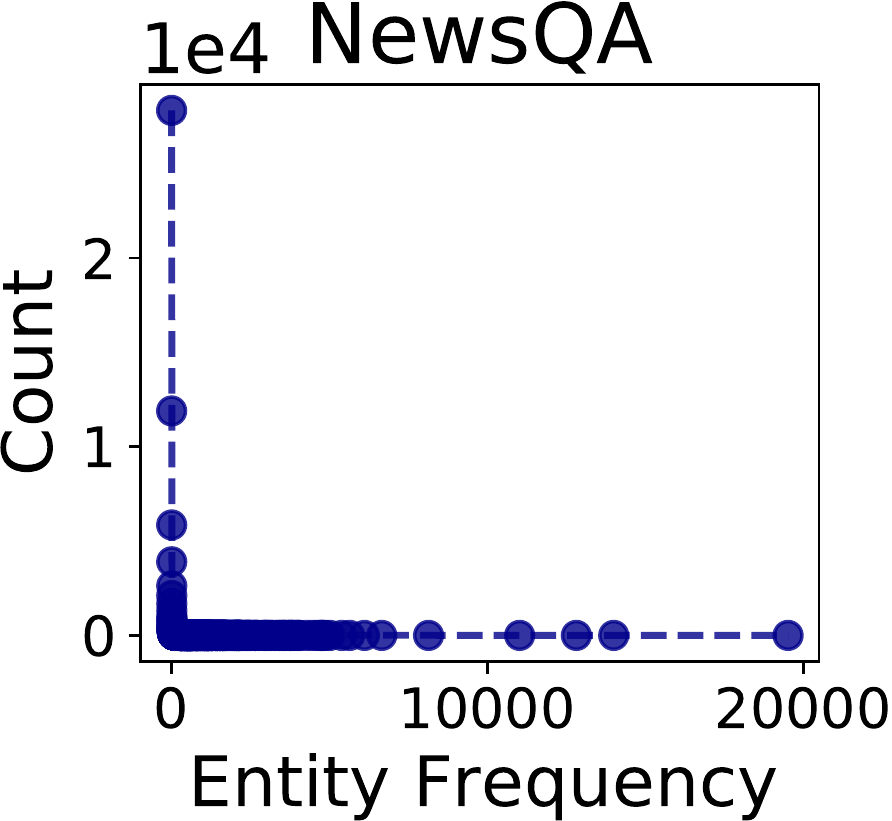}
    \end{minipage}
    \begin{minipage}{0.325\linewidth}
        \centering
        \includegraphics[width=1\linewidth]{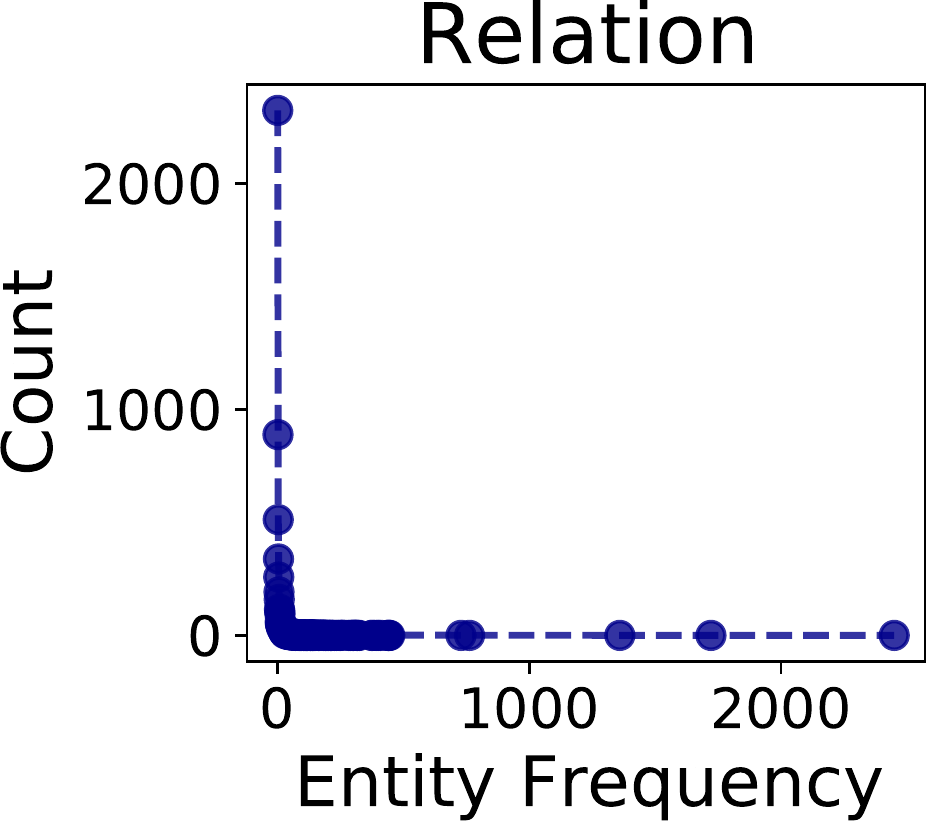}
    \end{minipage}
    \begin{minipage}{0.325\linewidth}
        \centering
        \includegraphics[width=1\linewidth]{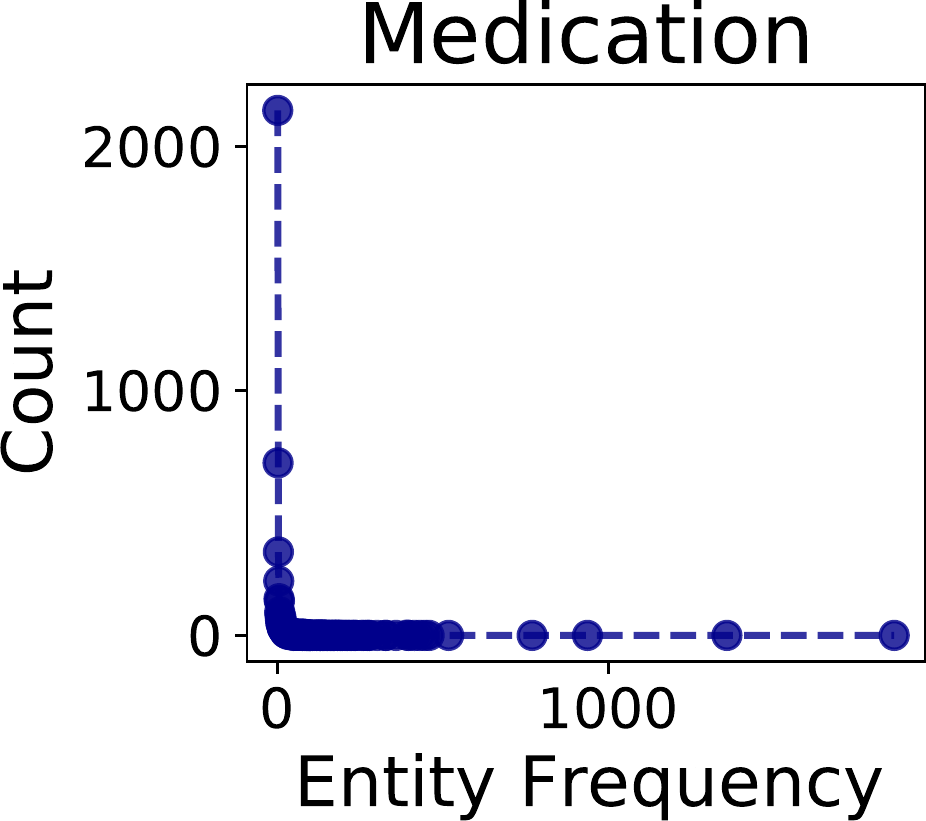}
    \end{minipage}
    \vspace{-0.1in}
    \caption{\small Distribution of frequency of entities on QA datasets: NewsQA, Relation, and Medication, where almost all entities appear less than 10 times, while an extremely few numbers of entities appear very frequently.}
    \label{fig:longtail/qa}
    \vspace{-0.05in}
\end{figure}

\subsection{Additional Representation Visualization}
While we already show the contextualized representations of seen and unseen entities in the latent space in Figure~\ref{fig:concept} right, we further visualize them including the missing baselines of Figure~\ref{fig:concept}, such as Fine-tuning or TAPT, in Figure~\ref{fig:unseen_embedding} on the NCBI-Disease dataset. Similar to Figure~\ref{fig:concept}, we observe that all baselines fail to closely embed the unseen entities in the representation space of seen entities.
While this visualization result does not give a strong evidence of why our KALA outperforms other baselines, we clearly observe that KALA is beneficial to represent unseen entities in the feature space of seen entities, which suggests that such an advantage of our KALA helps the PLM to generalize over the test dataset, where the context contains unseen entities.

\subsection{Entity Frequency Distribution}
\label{appendix:frequency}
\begin{figure}[t!]
    \begin{minipage}{0.325\linewidth}
        \centering
        \includegraphics[width=1\linewidth]{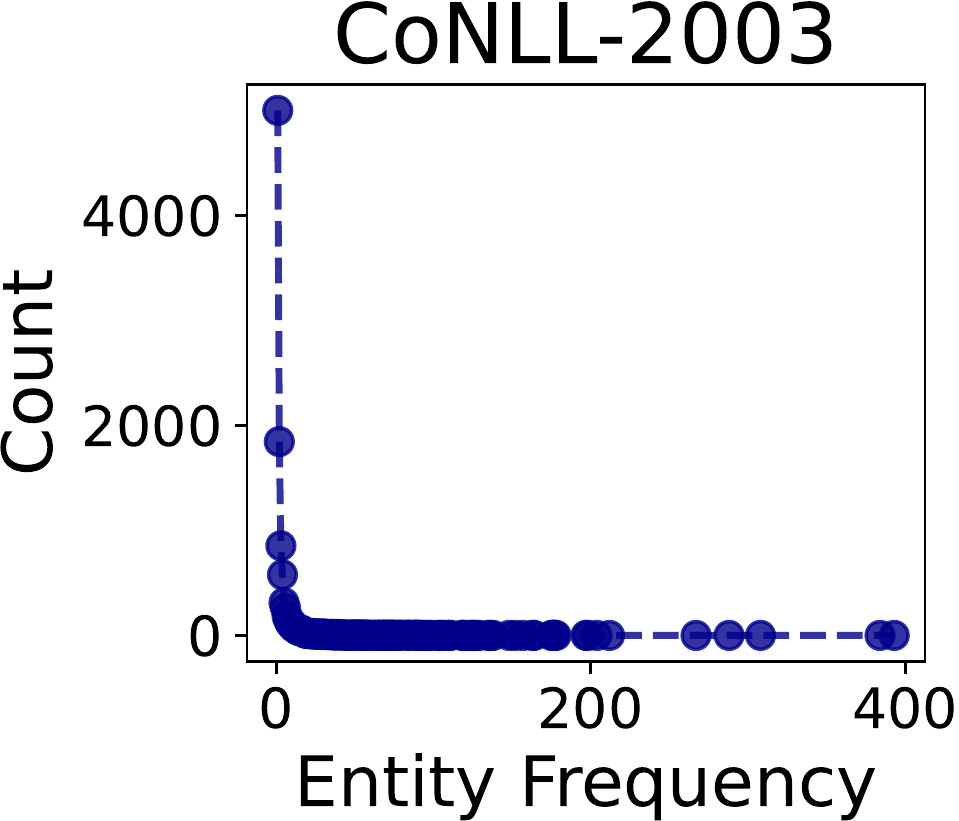}
    \end{minipage}
    \begin{minipage}{0.325\linewidth}
        \centering
        \includegraphics[width=1\linewidth]{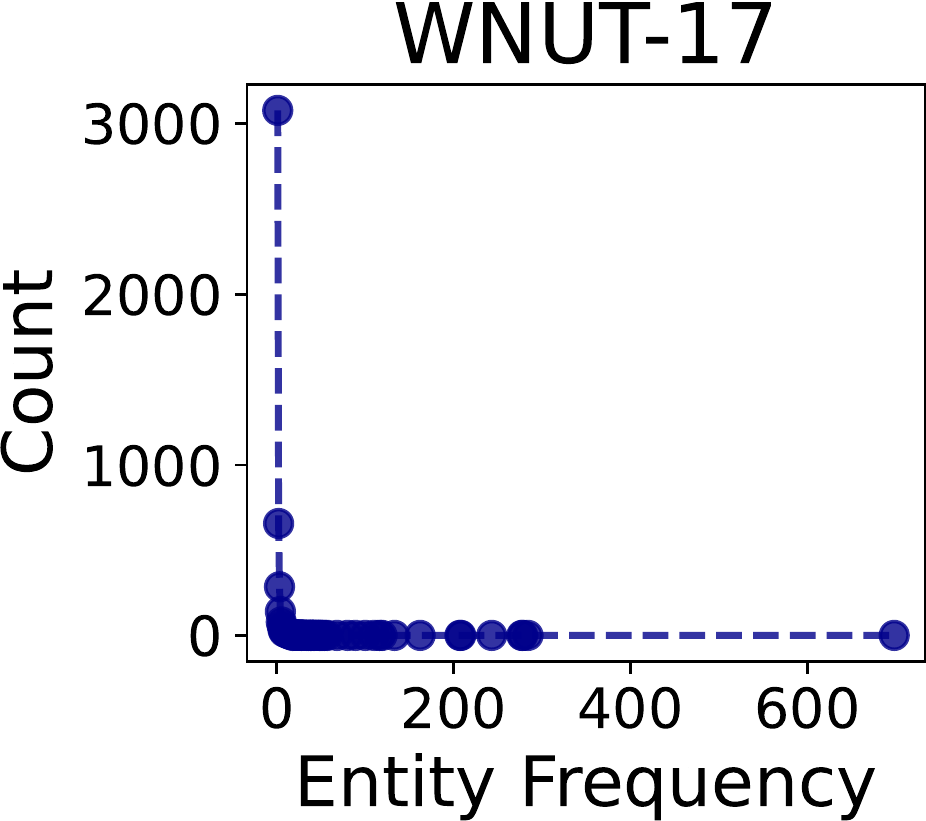}
    \end{minipage}
    \begin{minipage}{0.325\linewidth}
        \centering
        \includegraphics[width=1\linewidth]{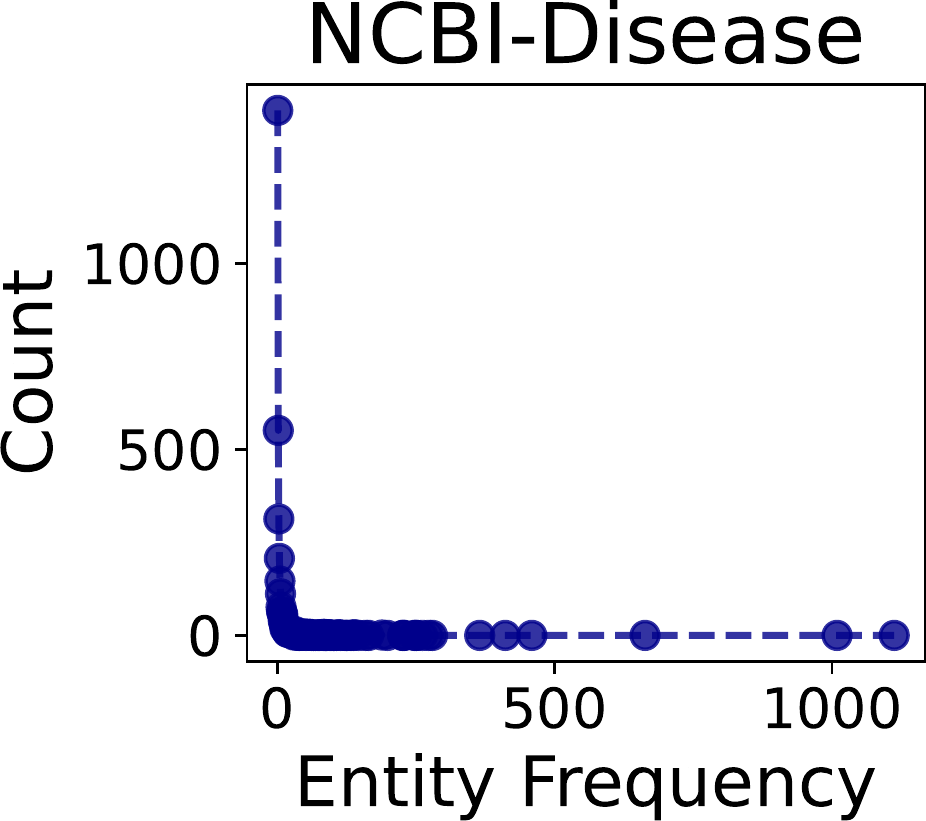}
    \end{minipage}
    \vspace{-0.1in}
    \caption{\small Distribution of frequency of entities on NER datasets: CoNLL-2003, WNUT-17, and NCBI-Disease, where almost all entities appear less than 10 times, while an extremely few numbers of entities appear very frequently.}
    \label{fig:longtail/ner}
    \vspace{-0.05in}
\end{figure}

We visualize the frequency of entities in Figure~\ref{fig:longtail/qa} and~\ref{fig:longtail/ner}. The entity frequency denotes the number of mentions of their associated entities within the entire text corpus of the training dataset. As shown in Figure~\ref{fig:longtail/qa} and~\ref{fig:longtail/ner} of QA and NER datasets, the entity frequency follows the long-tail distribution, where most entities appear a few times. For instance, in the NewsQA dataset, more than 20k entities among entire 60k entities appear only once in the training dataset, whereas one entity (\textit{CNN}\footnote{Almost every context in NewsQA includes the text `CNN' since they are originated from the CNN News.}) appears approximately 20k times. This observation suggests that most of the elements in the entity memory are not utilized frequently. In other words, only few entities are accurately trained with many training instances, whereas there exists the \textit{stale} embeddings which are rarely updated. This observation raises an interesting research question on the efficient usage of the entity memory, as we can see in Figure~\ref{fig:memory} that the small size of entity memory could result in the better performance (See Appendix~\ref{appendix:memorysize}). We leave the more in-depth analysis on the entity memory as the future work.

\begin{figure*}[t]
    \centering
    \includegraphics[width=1.0\linewidth]{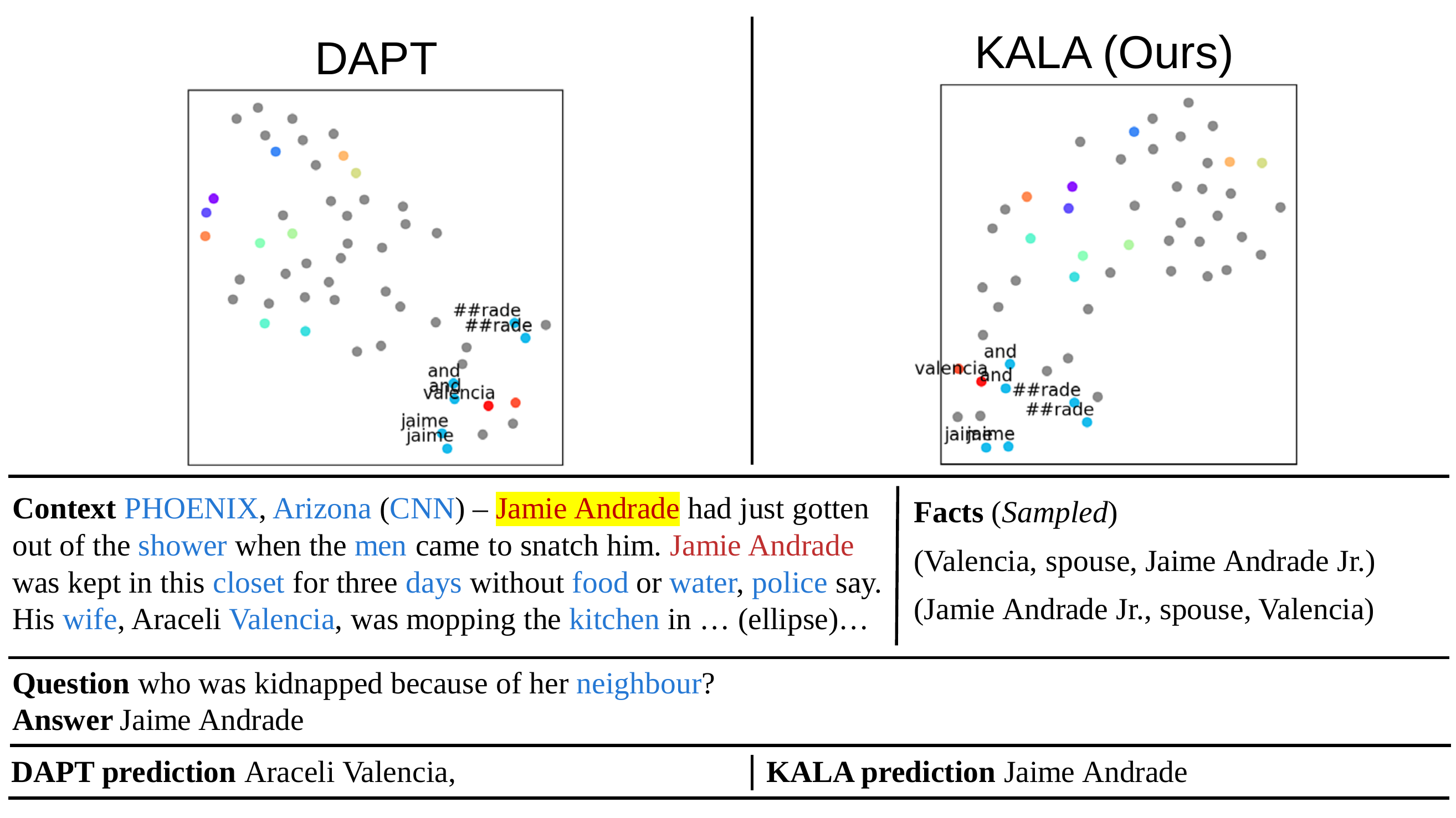}
    \vskip -0.1in
    \caption{\small A textual example from NewsQA with predictions from each method (DAPT and KALA), and also the T-SNE plot of contextualized representations from the last layer of BERT obtained by each method. Grey dots indicate tokens without any mentions, and dots in other colors indicate tokens with mentions to the entity. We also represent sampled facts in Knowledge Graph we used. Blue text indicates the mention of seen entities and red text indicates the mention of unseen entities. The fact is represented as the format of (head, relation, tail). Text with yellow background indicates the ground truth answer span.}
    \label{fig:casestudy_qa}
\end{figure*}

\subsection{Additional Case Study}
In addition to the case study in Figure~\ref{fig:casestudy}, we further show the case on the question answering task in Figure~\ref{fig:casestudy_qa}, like in Section~\ref{sec:case_study},
With this example, we explain how the factual knowledge in KGs could be utilized to solve the task via our KALA.
The question in the example is ``who was kidnapped because of \textbf{her} neighbor''.
We observe that DAPT answers this question as \textit{Araceli Valencia}. This prediction may come from matching the word `her' in the question to the feminine name `Araceli Valencia' in the context.
In contrast, our KALA predicts the \textit{Jaime Andrade} as an answer, which is the ground truth. We suspect that this might be because of the fact ``(Jaime Andrade, spouse, Valencia)'' in the knowledge graph, which relates the `Valencia' to the `Jaime Andrade'. Although it is not clear how it directly affects the model's performance, we can reason that KALA can successfully answer the question by utilizing the existing facts.

\begin{figure*}[!b]
    \centering
    \includegraphics[width=1.0\linewidth]{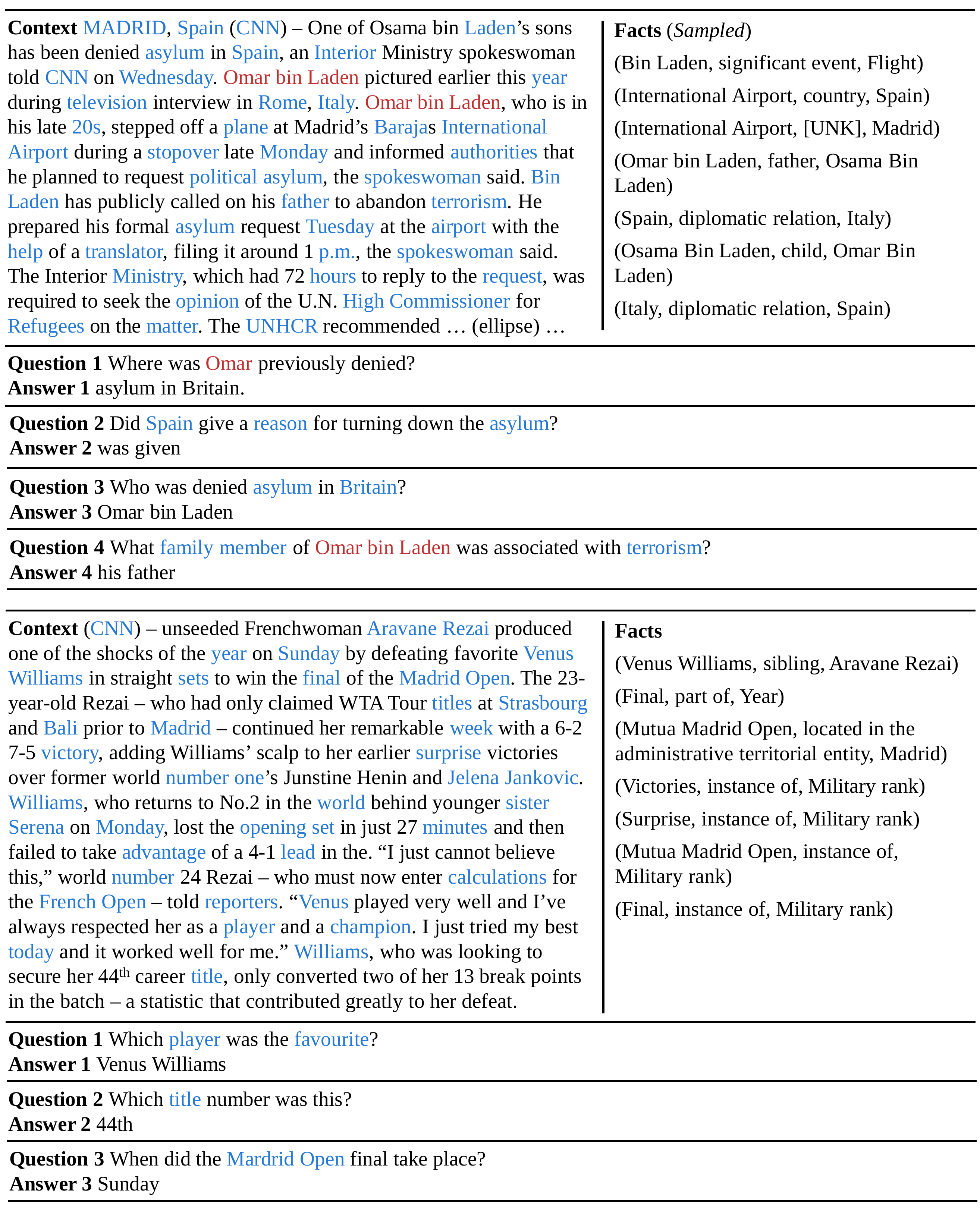}
    \vskip -0.1in
    \caption{\small NewsQA examples with facts in Knowledge Graph we used in this work. Blue text indicates the mention of seen entities and red text indicates the mention of unseen entities. The fact is represented as the format of (head, relation, tail).}
    \vskip -0.2in
    \label{fig:example_qa}
\end{figure*}

\subsection{Additional Data Visualization}
\begin{figure*}[t]
    \centering
    \includegraphics[width=1.0\linewidth]{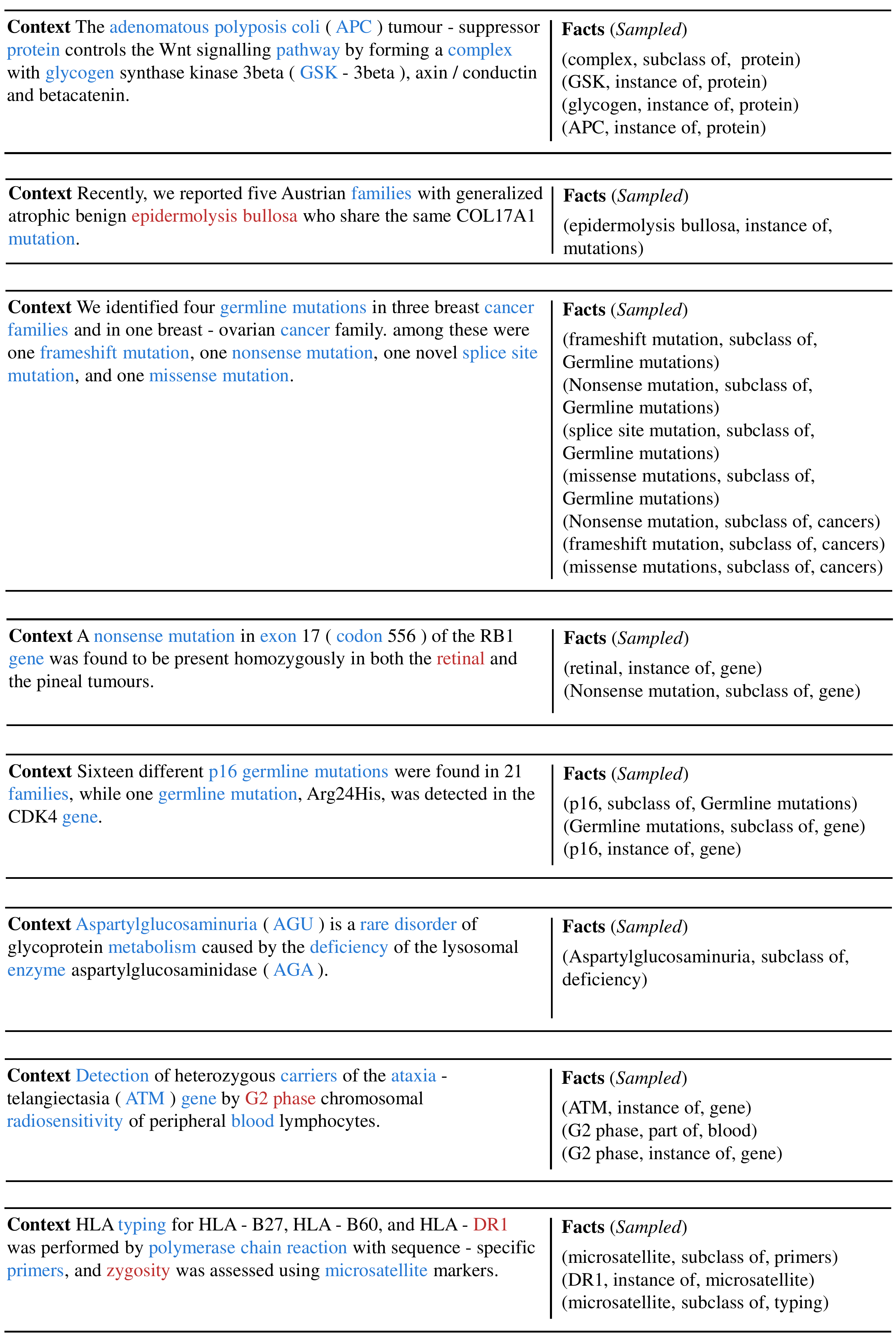}
    \vskip -0.1in
    \caption{\small NCBI-Disease examples with facts in Knowledge Graph we used in this work. Blue text indicates the mention of seen entities and red text indicates the mention of unseen entities. The fact is represented as the format of (head, relation, tail).}
    \vskip -0.2in
    \label{fig:example_ner}
\end{figure*}

In Figure~\ref{fig:example_qa} and~\ref{fig:example_ner}, we visualize the examples of the context with its seen and unseen entities and its relational facts. We first confirm that the quality of facts is moderate to use.
For instance, in the first example of Figure~\ref{fig:example_qa}, the fact in the context that \textit{Omar\_bin\_Laden} is son of \textit{Osama\_bin\_Laden}, is also appeared in the knowledge graph. In addition, we observe that there are facts that link unseen entities to the seen entities in both Figure~\ref{fig:example_qa} and~\ref{fig:example_ner}. Thus, while some of the facts in the knowledge graph are not accurate, we can represent the unseen entities with their relation to the seen entities. We expect that there is a still room to improve in terms of the quality of KGs, allowing our KALA to modulate the entity representation more accurately. We leave the study on this as the future work.

\end{document}